\documentclass[twocolumn, DIV26, 8pt]{scrartcl}
\usepackage{tgschola}
\usepackage{helvet}

\usepackage{graphicx} 
\usepackage{color}
\usepackage{subcaption}
\usepackage{url}

\usepackage{xcolor}

\definecolor{blue}{rgb}{0.31,0.44,0.66}

\usepackage[pdftex,
colorlinks=true,
linkcolor=blue,
citecolor=black,
urlcolor=blue,
hyperfootnotes=false]{hyperref}

\usepackage[normalem]{ulem}
\usepackage{todonotes}
\usepackage{amsmath}
\usepackage{amsmath,bm}
\usepackage{amssymb}
\usepackage{multicol}
\usepackage{booktabs}
\usepackage{scicite}

\usepackage[labelfont={bf,small,sf}]{caption}
\DeclareCaptionLabelSeparator{pp}{.}
\setcapindent{0cm}
\DeclareCaptionType{si-figure}[Fig.]
\DeclareCaptionLabelFormat{bf-parens}{#1 S#2.}
\usepackage{fancyhdr}
\newcommand{\raisedrule}[2][0em]{\leaders\hbox{\rule[#1]{15pt}{#2}}\hfill}
\makeatletter
\newcommand{\removelatexerror}{\let\@latex@error\@gobble}
\makeatother
\let\oldnl\nl
\newcommand{\nonl}{\renewcommand{\nl}{\let\nl\oldnl}}

\def\bfp{{\mbox{\boldmath $p$}}}

\date{}

\begin{document}
\pagestyle{fancy}
\makeatletter
\twocolumn[
   \begin{@twocolumnfalse} 
    {\fontsize{25}{12}\selectfont
    \textsf{\textbf{Prescient whole-body teleoperation of humanoid robots}}}
    
    \bigskip
    
    \textsf{\textbf{\large{Luigi Penco$^{1}$, Jean-Baptiste Mouret$^{1}$, Serena Ivaldi$^{1}$}}}

     \bigskip   
     \textsf{$^1$~Inria Nancy -- Grand Est, CNRS, Universit\'e de Lorraine, France}
     
     \bigskip
     
     \textsf{serena.ivaldi@inria.fr}
     \bigskip
     
     \bfseries \sffamily
\noindent{}

Humanoid robots have the potential to be versatile and intuitive human avatars that operate remotely in inaccessible places:
the robot could reproduce in the remote location the movements of an operator equipped with a wearable motion capture device while sending visual feedback to the operator.
While substantial progress has been made on transferring (``retargeting'') human motions to humanoid robots, a major problem preventing the deployment of such systems in real applications is the presence of communication delays between the human input and the feedback from the robot: even a few hundred milliseconds of delay can irremediably disturb the operator, let alone a few seconds.
To overcome these delays, we introduce a system in which a humanoid robot \emph{executes commands before it actually receives them}, so that the visual feedback appears to be synchronized to the operator, whereas the robot executed the commands in the past. To do so, the robot continuously predicts future commands by querying a machine learning model that is trained on past trajectories and conditioned on the last received commands.
In our experiments, an operator was able to successfully control a humanoid robot (32 degrees of freedom) with stochastic delays up to 2 seconds in several whole-body manipulation tasks, including reaching different targets, picking up, and placing a box at distinct locations.

\bigskip

    \end{@twocolumnfalse}
]
\makeatother


\section*{Introduction} 
\label{sec:intro}

Humanoid robots are some of the most versatile machines ever created \cite{Yamane2016}. Thanks to their anthropomorphism, they can perform bimanual manipulation like grasping large boxes, they can bend to pick up objects from the ground, or reach high places by the ceiling; they can also walk on rough terrain, crawl, climb stairs or push objects. Differently from other robotic platforms, they have a kinematic structure that allows them to enter narrow or cluttered spaces, and operate in environments that were primarily designed for humans. In short, humanoid robots have the potential of being as versatile as humans, being currently limited only by their mechatronics, which need to be further developed to match the human flexibility. 

This versatility is especially important when robots are sent to truly unexpected situations in which they have to perform a task for which they were not designed \cite{drc}. While robots in the manufacturing industry perform well-defined tasks in known environments, many situations in the outside world involve some creative adaptation of the tasks. For instance, working in remote places like in space or on an oil platform very often requires performing tasks in unexpected ways. Similarly, disasters --- natural or industrial --- are events that are, by definition, out of the ordinary routine, and during which robots could be critical to save lives.

However, autonomous robots are far from being the creative problem solvers that humans are; this is why humanoid robots are most useful when they are teleoperated \cite{drc,stilman2008humanoid}: the capabilities of the robot and of the operator are combined to solve complex problems remotely. While humanoid robots can be operated from high-level commands and classic joysticks \cite{stilman2008humanoid}, they can exploit their human-like morphology to directly replicate the human posture and gestures, captured by a motion tracking system. This technique coupled with the intelligence of the human operator makes them realize tasks fluently as human avatars. With such a \emph{whole-body teleoperation}, the entire body of the robot can be controlled precisely while the human operator, equipped with a virtual reality headset, can feel as if they have been ``projected'' to the remote situation. Additionally, there is a good match between teleoperation and humanoid robots because they are typically considered for high-risk missions (life-or-death situations, nuclear accidents, ...) for which the wage of an operator is negligible, but the stakes are too high to let a robot decide everything by itself.

Two main challenges are raised by humanoid teleoperation: (1) how to map a whole-body motion of the human to a robot that has close, but different physical constraints and dynamics \cite{humanoids2018,pencoral}? and (2) how to cope with the unavoidable communication delays between the movement of the operator and the feedback? The present paper addresses both of these challenges but focuses on delay compensation because substantial work has already been achieved for whole-body retargeting \cite{humanoids2018,penco2019}.

The first studies with robot manipulators \cite{ferrell1965} suggested that the typical user behavior in a teleoperation system with time delays is to adopt a ``move-and-wait'' strategy to avoid overcompensations for the delayed perceived errors. However, follow-up experiments \cite {ferrell1967} showed that this strategy is not effective, even with time delays around $0.3$s. This is in sharp contrast to the typical delays that are considered in real applications, for instance $1$ second on average ($30$s maximum) during the DARPA Robotics Challenge \cite{drc}, $10$ seconds ($20$s maximum) for the NASA Space Challenge \cite{NRSC}, or $0.8$s ($3$s maximum) in the METERON project~\cite{kontur-meteron}, during which a robot was teleoperated on Earth from the International Space Station.

\begin{figure*}[!t]
    \centering
    \includegraphics[width=\linewidth]{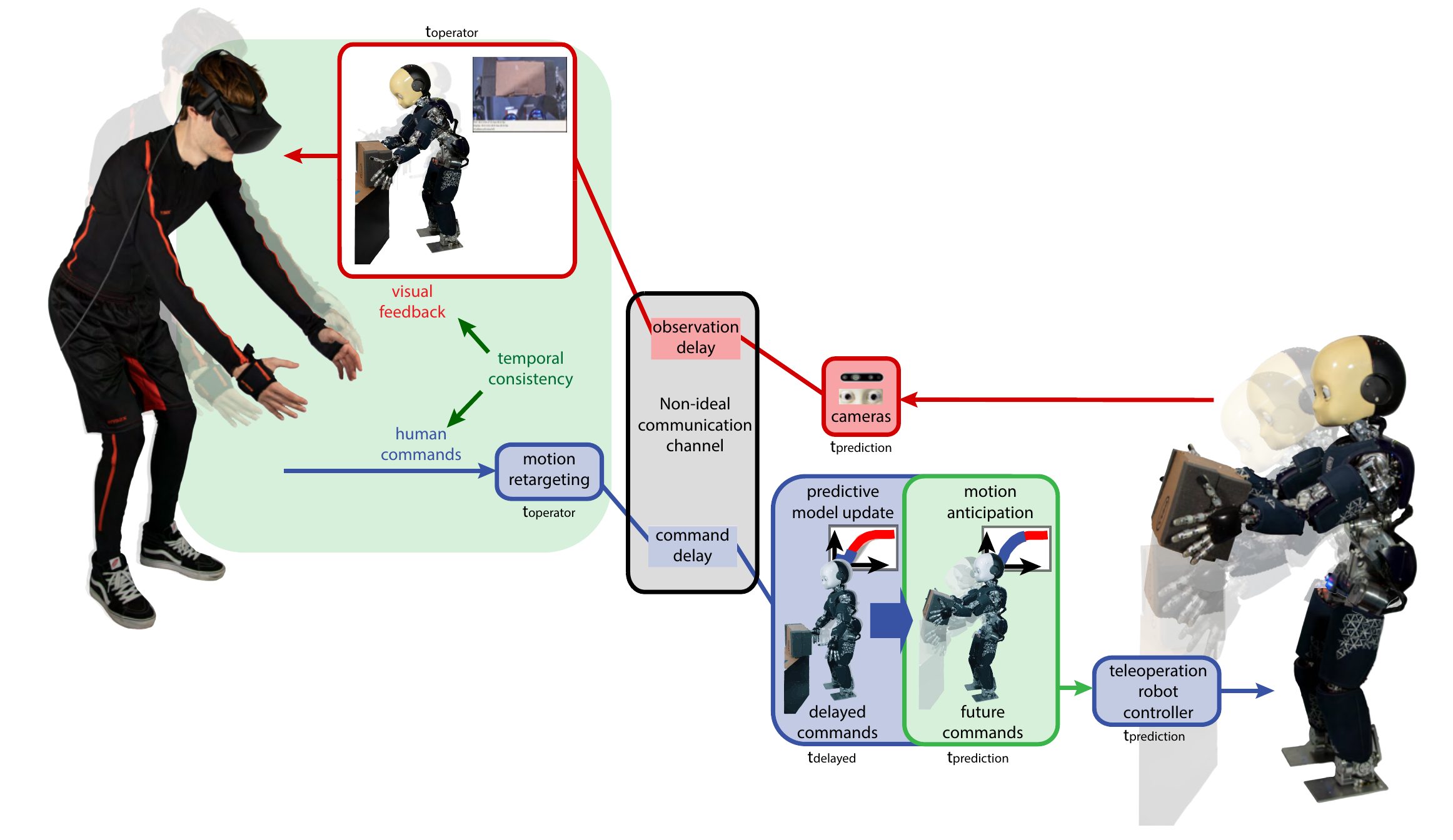}
    \caption{\textbf{Concept of prescient teleoperation.} The operator wears a motion capture suit and a virtual reality headset. They teleoperate a humanoid robot over a network with communication delays (up to 2 seconds).
    To send a synchronized visual feedback  to the operator, the robot \emph{anticipates} the commands thanks to data-driven predictors that are trained from a few examples beforehand: the robot executes commands that the operator has not given yet by predicting the most likely commands in the next few seconds.
    }
    \label{fig:intro}
\end{figure*}

Time-delayed teleoperation almost exclusively uses \textit{predictive displays} \cite{penin2002,hernando2007}, which are virtual displays representing a model of the robot and its environment, to compensate for the delayed visual feedback. In these systems, the operator performs the task on the display, controlling the simulated robot without any time delay, while sending the same commands -- which will be delayed -- to the real robot.
More complex predictive displays overlay the predicted graphics on the delayed video \cite{Bejczy1990,mitra2008}, which can be achieved by identifying a model of the environment and transmitting it back to the operator side \cite{mitra2008}.
However, in most cases, the predicted graphics and the delayed video signals are kept in different displays, due to the difficulty of mixing video and graphics with enough quality and robustness.
In all these cases, since perfect modeling is impossible, there is always going to be a difference between the real and modeled environment, which has to be coped with by some local robot autonomy. Hence, these techniques just offer approximate cues until the actual feedback information is available, and should be considered as support tools rather than standalone solutions.

\begin{figure*}
    \centering
    \includegraphics[width=\linewidth]{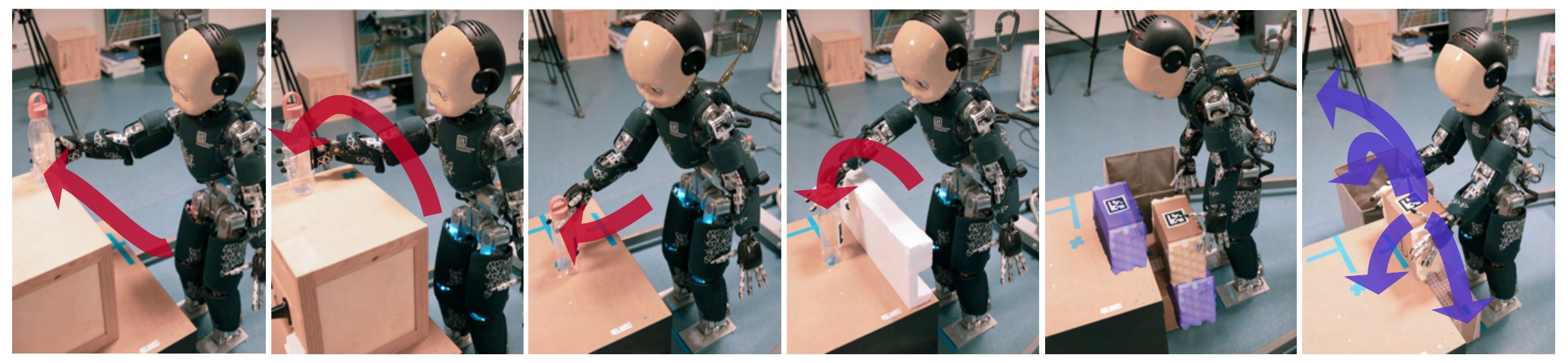}
    \caption{\textbf{Tasks performed by the robot during the experiments (dataset ``Multiple tasks'').} In the first scenario (4 left images), the robot is teleoperated to reach a bottle at different locations and in different ways; in the second scenario (2 right images), the robot has to pick up a box from different locations then placing it in another location.}
    \label{fig:tasks}
\end{figure*}

\begin{figure*}
    \centering
    \includegraphics[width=\linewidth]{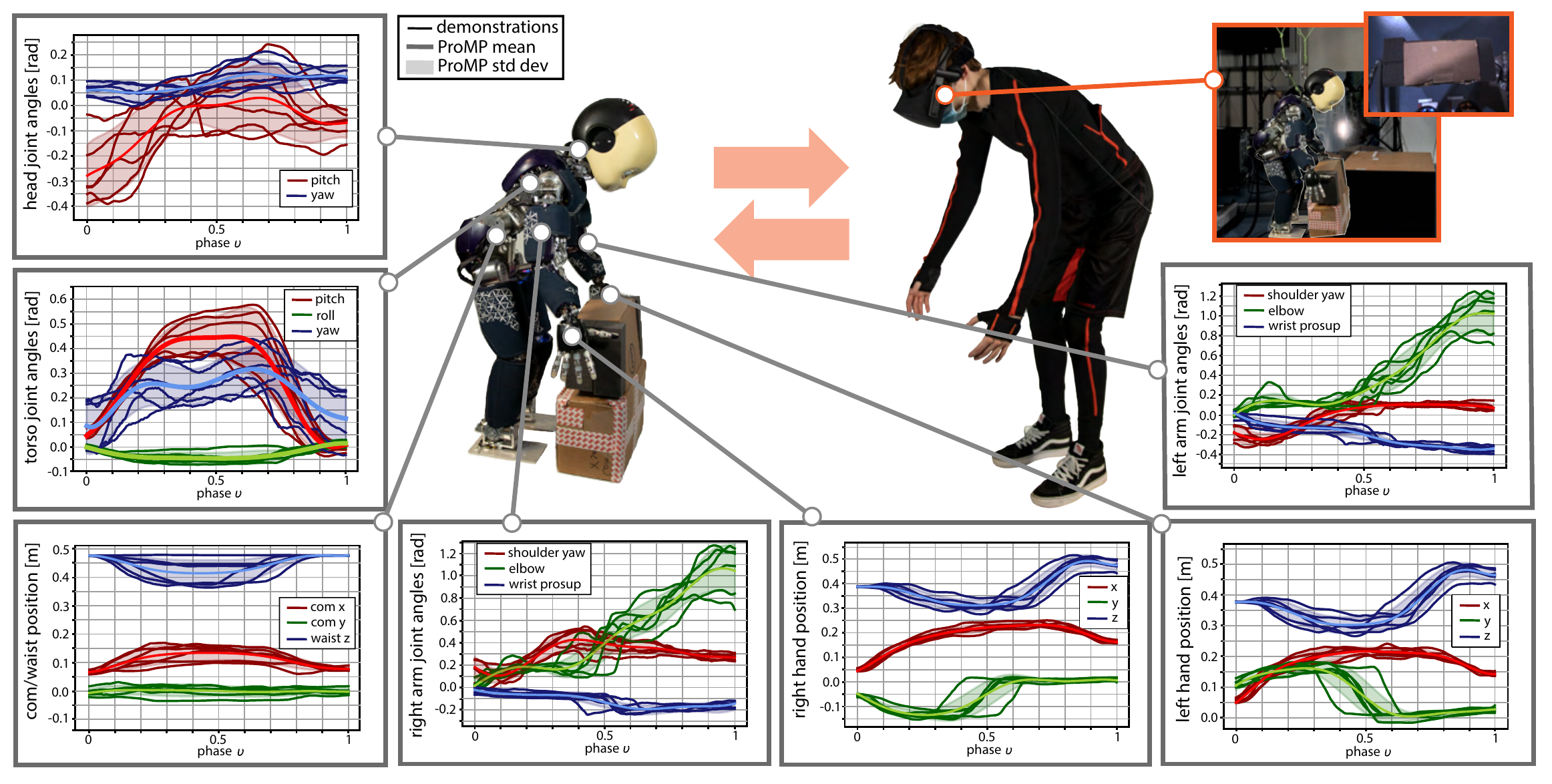}
    \caption{\textbf{Learned predictors for the task of picking up a box at a low position.} For each of the 6 demonstrations, the motion of the operator is first ``retargeted'' to the robot using the whole-body controller (ignoring delays). The trajectories of each body/joint of the robot is then recorded. From this set of demonstrations (thin lines), a ProMP is fitted for each trajectory; this ProMP is represented here as a thick line (the mean) and a light zone (the standard deviation). The computed mean is a smooth trajectory that averages all the demonstrations and the standard deviation captures the variability of the demonstrations.}
    \label{fig:learned}
\end{figure*}


Here, we introduce a novel teleoperation system in which the operator gets a synchronized video feed of \emph{real images}, even when the communication channel imposes a $1$ to $2$ seconds delay. 

Our key idea is that if the robot executes the desired movement \emph{before} the operator performs it, then the operator will watch a delayed video feed that will be almost indistinguishable from a real-time feed (Fig. \ref{fig:intro}). At each time-step, the robot analyzes the data that it has received so far, measures the communication time, estimates the communication time to send the feedback, and predicts what the operator is most likely to do in the next seconds. This prediction makes it possible to execute the command with enough anticipation so that the user receives a video feed that correspond to the past of the robot, but that matches the present time for the operator. We call this feedback scheme based on a prediction of the inputs \emph{prescient teleoperation}.

\section*{Results}
\label{sec:results}
\subsection*{Overview of the prescient teleoperation system}

The whole-body motion of the operator is captured with a motion capture suit that computes both the joint angles and the Cartesian positions of the operator (Methods). This motion cannot be directly used as reference for the robot because of the difference in kinematics
(e.g., joint limits, body dimensions) and dynamics (e.g., mass distribution). The system therefore needs to ``retarget'' the motion \cite{humanoids2018, pencoral, penco2019}, that is, to compute references that make sense for the robot. To do so, Cartesian references are scaled using a fixed factor that accounts for the size difference between the human and the operator (Methods). This means that when the operator moves the hands by $10$cm, the robot might move them by only $5$cm. The angular references encourage the robot to take the same posture as the operator when it is possible; they are retargeted by mapping each joint of the operator to the robot (Fig. S\ref{fig:retarget}) and computing angular positions relative to the initial joint positions. Last, the reference of the center of mass is computed from the human reference to fit the robot kinematics (Methods).

We use the humanoid robot iCub \cite{natale2017icub, icub}, which has 32 degrees of freedom (ignoring the hands and the eyes) and is position-controlled (Methods). The whole-body motion of the robot is defined by the trajectories of the center of mass ground projection, the waist height, the hands positions, and the posture of the arms, of the neck and of the torso. Each of these trajectories has a different priority which determines how the robot's controller executes the entire movement (Methods): the top priority is given to the center of mass (to avoid falling) and the feet poses (which should not move in our experiments, since we only target double support motions), and the postural trajectories have the lowest priority.
The exact hard and soft priorities were found in previous work with a multi-objective stochastic optimizer so that the robot is unlikely to fall but tracks the trajectories as precisely as possible \cite{pencoral}. 

At each time-step (100 Hz), the robot searches for the joint positions by solving a hierarchical constrained quadratic problem with inequality and equality constraints \cite{escande2014hierarchical,rocchi2015opensot} (Methods) that minimizes the tracking error, i.e., the distance to the references, while taking into account the priorities and the constraints (kinematic model, joint velocities and zero moment point bounds). In absence of delays, when prediction is not necessary, the references are the trajectories retargeted from the human operator \cite{penco2019}.

To operate with delays, 
the system predicts at each time-step the most likely future trajectory of the most likely task given the commands received so far from the operator 
(for instance, the 3D trajectories of the hands, etc.). The system is trained beforehand on a few example trajectories that encode the different ways of executing a task, using the principle of ``motion primitives'':
for instance, the human is likely to reach for a bottle on the table with a similar straight hand trajectory at all times, but the trajectory may be more or less curved in presence of an obstacle.
Though human gestures are generally stereotyped, the intrinsic motor noise, the human preference of movement and the small differences in the task executed in the real world (e.g., a displacement of a target object) induce variability in the human motion trajectories to realize a specific task. 
Even if the trajectory asked by the operator is different from the training set but included in the distributions of what has been previously demonstrated, accurate predictions can still be generated on a short time-scale.

Numerous generic machine learning techniques have been proposed to predict the future of time-series, especially with neural networks \cite{parmezan2019evaluation,lim2021time}. Nevertheless, the robotics community has been working for a long time on regression techniques that are well-suited for robot trajectories. In particular, Movement Primitives (MPs) are a well-established approach for representing and learning motion trajectories in robotics.
They provide a data-driven representation of movements and support generalization to novel situations, temporal modulation and  sequencing of primitives.
Several kinds of MPs have been proposed in the literature.
Dynamic Movement Primitives (DMPs) \cite{Schaal2006} are a formulation of movement primitives with autonomous nonlinear differential equations.
The linear parameterization of DMPs makes them suitable for supervised learning from demonstration. Moreover, the temporal, scale, and translation invariance of the differential equations with respect to these parameters provides a useful means for movement recognition.

Our system is based on Probabilistic Movement Primitives (ProMPs) \cite{paraschos2017}, which extend the concept of dynamic movement primitives to model the variance of the demonstrations (Methods). Hence, a ProMP is a distribution over trajectories, which makes it ideal tool to model the variability of human demonstrations in representing motion primitives \cite{paraschos2017}.
This represents an advantage over deterministic  approaches, that can only represent the mean solution. 

Another advantage of working with distributions is that the properties of motion primitives can be translated into operations from probability theory. For example, modulation of a movement to a novel target can be realized by \emph{conditioning} on the desired target’s positions or partial trajectories \cite{maeda2017probabilistic}.
Specifically, our system uses the ProMP’s conditioning operator to adapt the predictions according to the incoming observations, hence obtaining at each timestep an updated prediction (posterior) for the prosecution of the current movement (Methods) given the learned model of its associated primitive (prior). In other words, a ProMP predicts the mean trajectory of prior demonstrations when there is no conditioning data, but when some data is available, such as observations of the current movement, it adapts its prediction to resemble the most likely trajectories from the training set.

\subsection*{Training and test sets}
To train our system, an operator teleoperated the iCub humanoid robot in a local network, which we consider as an approximation of an ideal network without any delay, and performed several tasks (Figs. \ref{fig:tasks}, S\ref{fig:dataObst}, S\ref{fig:traindiff}). They were wearing an inertial motion capture suit to get motion references and a virtual reality headset to get the visual feedback (Methods). For each motion, we recorded the trajectories of the center of mass ground projection, the waist height, the hands positions, the arms posture (shoulder rotation, elbow flexion, forearm rotation), the neck posture (flexion and rotation) and the torso posture (flexion, rotation and abduction). For each of these trajectories, a ProMP was learned (Fig. \ref{fig:learned}). Hence, for each task, we have a set of ProMPs associated to the whole-body motion.

We considered several scenarios, which resulted in the acquisition of three different datasets, each consisting of a training and a testing set (Methods). In the first dataset (Multiple Tasks, Fig. \ref{fig:tasks}), the tasks correspond to reaching a bottle (scenario 1) or picking a box with the two hands (scenario 2). The test trajectories correspond to different repetitions of the tasks used for training, and encode the intrinsic movement variability of the operator. In the second dataset (Obstacles, Fig. S\ref{fig:dataObst}-\ref{fig:testObst}), the test trajectories correspond to different ways of reaching a bottle while avoiding obstacles that were not considered during training, that is, performing the same task but in different ways. In the last dataset (Goals, Fig. S\ref{fig:traindiff}-\ref{fig:testdiff}), each motion correspond to reaching a bottle at different positions, and the test trajectories are for positions that were not used for training.

\begin{table*}
\center
\caption{\textbf{Prediction error after observing different portions of the commanded trajectories (dataset Multiple Tasks).} 
We evaluated the difference between the actual trajectory (commands retargeted from the operator) and the predicted trajectory for the 20 testing motions from the bottle reaching scenario (including those from Fig.~S\ref{fig:tests1}) and the 21 testing motions from the box handling scenario (including those from Fig.~S\ref{fig:test2})}. To understand the influence of the conditioning of the ProMPs, we computed the mean error by following the mean of the ProMP selected by hand (`no obs'), after the initial recognition ('recognition') that takes around 1s, after a fourth of the motion (1/4 motion) and after half of the motion (1/2 motion). Thanks to the conditioning, when more data is used, the prediction is more accurate, which means that the prediction is adjusted to suit the particular motions of the operator (that is, the robot does not simply follow the mean trajectory once it has recognized it). Examples of predicted trajectories are displayed in Fig.~\ref{fig:pred}.
    \begin{scriptsize}
\begin{tabular}{ccccccccc}
\toprule
& \multicolumn{4}{c}{Box handling} & \multicolumn{4}{c}{Bottle reaching}\\
{\textbf{Trajectory}} &  \multicolumn{4}{c}{\textbf{RMS error [rad]}} &  \multicolumn{4}{c}{\textbf{RMS error [rad]}}\\
 & no obs & recognition & 1/4 motion & 1/2 motion & no obs & recognition & 1/4 motion & 1/2 motion \\
\cmidrule(lr){2-5}
\cmidrule(lr){6-9}

head yaw  & 0.112$\pm$0.049 & 0.080$\pm$0.028 & 0.045$\pm$0.012 & 0.012$\pm$0.009 & 0.083$\pm$0.038 & 0.040$\pm$0.015 & 0.023$\pm$0.011 & 0.010$\pm$0.008\\
torso pitch  & 0.155$\pm$0.064 & 0.120$\pm$0.046 & 0.054$\pm$0.030 & 0.018$\pm$0.008 & 0.103$\pm$0.056 & 0.061$\pm$0.022 & 0.044$\pm$0.015 & 0.019$\pm$0.008\\
torso roll  & 0.119$\pm$0.049 & 0.103$\pm$0.038 & 0.055$\pm$0.028 & 0.017$\pm$0.008 & 0.082$\pm$0.042 & 0.054$\pm$0.016 & 0.032$\pm$0.010 & 0.016$\pm$0.008\\
torso yaw  & 0.168$\pm$0.053 & 0.136$\pm$0.049 & 0.079$\pm$0.032 & 0.049$\pm$0.019 & 0.088$\pm$0.046 & 0.065$\pm$0.039 & 0.049$\pm$0.023 & 0.033$\pm$0.018\\
r. should. yaw  & 0.164$\pm$0.059 & 0.109$\pm$0.050 & 0.053$\pm$0.019 & 0.040$\pm$0.011 & 0.172$\pm$0.059 & 0.114$\pm$0.056 & 0.055$\pm$0.024 & 0.027$\pm$0.011\\
r. elbow  & 0.169$\pm$0.072 & 0.146$\pm$0.038 & 0.097$\pm$0.032 & 0.033$\pm$0.010 & 0.220$\pm$0.083 & 0.097$\pm$0.032 & 0.065$\pm$0.017 & 0.034$\pm$0.011\\
r. wrist pros.  & 0.113$\pm$0.049 & 0.092$\pm$0.022 & 0.043$\pm$0.012 & 0.026$\pm$0.008 & 0.124$\pm$0.052 & 0.071$\pm$0.022 & 0.048$\pm$0.016 & 0.021$\pm$0.008\\
 &  \multicolumn{4}{c}{\textbf{RMS error [cm]}} &  \multicolumn{4}{c}{\textbf{RMS error [cm]}}\\
 & no obs & recognition & 1/4 motion & 1/2 motion & no obs & recognition & 1/4 motion & 1/2 motion \\
\cmidrule(lr){2-5}
\cmidrule(lr){6-9}
r. hand x  & 3.76$\pm$0.91 & 2.10$\pm$0.48 & 1.57$\pm$0.49 & 0.78$\pm$0.12 &  3.54.$\pm$0.88 & 1.62$\pm$0.65 & 0.61$\pm$0.33 & 0.48$\pm$0.14\\
r. hand y  & 3.55$\pm$0.93 & 1.91$\pm$0.40 & 0.97$\pm$0.32 & 0.52$\pm$0.10 & 3.71$\pm$0.89 & 2.02$\pm$0.48 & 0.89$\pm$0.19 & 0.35$\pm$0.11\\
r. hand z  & 2.98$\pm$0.91 & 2.34$\pm$0.83 & 1.22$\pm$0.27 & 0.66$\pm$0.19 & 2.71$\pm$0.80 & 1.99$\pm$0.79 & 0.77$\pm$0.29 & 0.50$\pm$0.13\\
com x  & 2.40$\pm$0.76 & 0.98$\pm$0.26 & 0.57$\pm$0.21 & 0.23$\pm$0.14 & 0.78$\pm$0.53 & 0.39$\pm$0.22 & 0.29$\pm$0.12 & 0.12$\pm$0.10\\
com y  & 1.12$\pm$0.55 & 0.58$\pm$0.26 & 0.29$\pm$0.18 & 0.20$\pm$0.13 & 0.80$\pm$0.44 & 0.52$\pm$0.29 & 0.29$\pm$0.12 & 0.11$\pm$0.10\\
waist z  & 3.53$\pm$1.04 & 2.74$\pm$0.93 & 1.68$\pm$0.57 & 0.65$\pm$0.22 & 0.95$\pm$0.36 & 0.65$\pm$0.22 & 0.39$\pm$0.19 & 0.28$\pm$0.14\\
\bottomrule
\end{tabular}
\end{scriptsize}
\label{tab:pr}
\end{table*}

\begin{figure*}
    \centering
    \includegraphics[width=\linewidth]{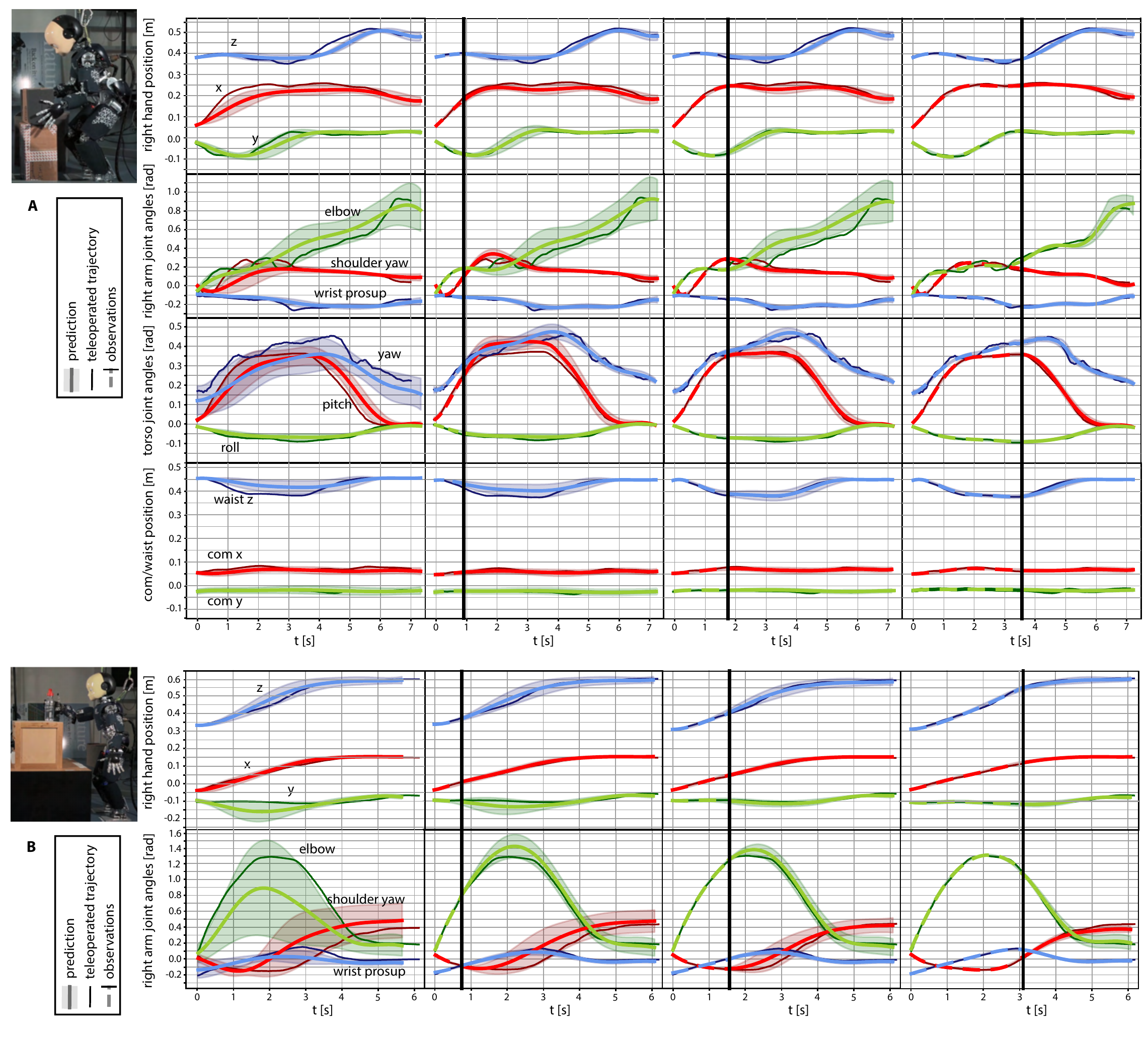}
    \caption{\textbf{Prediction update according to observations.} (A) The robot is picking up a box in front of it at a mid height. (B) The robot is reaching a bottle located on top of a box. The most relevant predicted trajectories (light colored lines) are compared to the non-delayed trajectories at the operator's side (dark colored lines), after observing different portions of the motion; a perfect prediction would mean that the light line (green/blue/red) line matches the dark line (green/blue/red). The non-delayed trajectories are from the testing scenario of dataset Multiple Tasks and the experiment corresponds to a particular case of those reported in Table~\ref{tab:pr}. From left to right, the figure shows the prediction given by the ProMPs learned from the demonstrations, the prediction updated after observing the first portion of motion used to infer the task and its duration, the prediction updated after observing a fourth of the motion, and after observing half of the motion. After less than 1s of observation, the light trajectory is similar to the dark trajectory, whereas it might have been far from the ProMP mean (see, for instance, the elbow in experiment (b)). In most cases, 2 seconds is enough to obtain very accurate predictions for the next 4-5 seconds.}
    \label{fig:pred}
\end{figure*}

\subsection*{Evaluation of the predicted trajectories}
Using testing data from the dataset Multiple Tasks, we evaluated the ability of conditioned ProMPs to predict the future motion of the operator, independently of any consideration for teleoperation. At each time-step, the robot identifies which ProMP best describes the current motion by selecting the ProMP that minimizes the distance between the observations so far and the mean of each ProMP (Methods), and it continuously updates the posterior distribution with the observation. 
Fig. \ref{fig:pred} illustrates the quality of the prediction for the two scenarios of the dataset Multiple Tasks in different situations due to the observed data: after observing a sufficient portion of the trajectory (around $1$s) that identifies the ProMP that best describes the current motion, after a fourth of the trajectory, and after half the trajectory. The robot uses the data received so far to predict the trajectory up to the end of the experiment. Visually, the predictions match the actual trajectory very well, although it is smoother, even after less than $1$s of observations. In addition, the real trajectory is almost always contained in the distribution of possible trajectories of the prediction (visualized with the variance of the ProMP). 
To quantitatively compare the prediction to the ground truth, we considered the whole-body motions from the testing set of the dataset Multiple Tasks and reported in Table \ref{tab:pr} the corresponding prediction error between the two. The error decreases over time as more observations become available to update the prediction, reaching an error around half centimeter for the Cartesian trajectories after observing half of the motion. After observing only half of the motion, the quality of the prediction improves by one and a half centimeters in the Cartesian trajectories of the hands with respect to the first prediction available after the recognition, which means that the ProMPs not only recognized the motion but also identified the specific execution of movement by the operator.
A significant improvement can also be observed in the postural trajectories. These results show that continuously updating the prediction allows the operator to influence the predicted motion so that it matches better their intentions. Put differently, the system is not simply recognizing the motions then executing the mean of the learned demonstrations: it accurately predicts the future trajectories given the data received so far.

\begin{figure*}
    \centering
    \includegraphics[width=\linewidth]{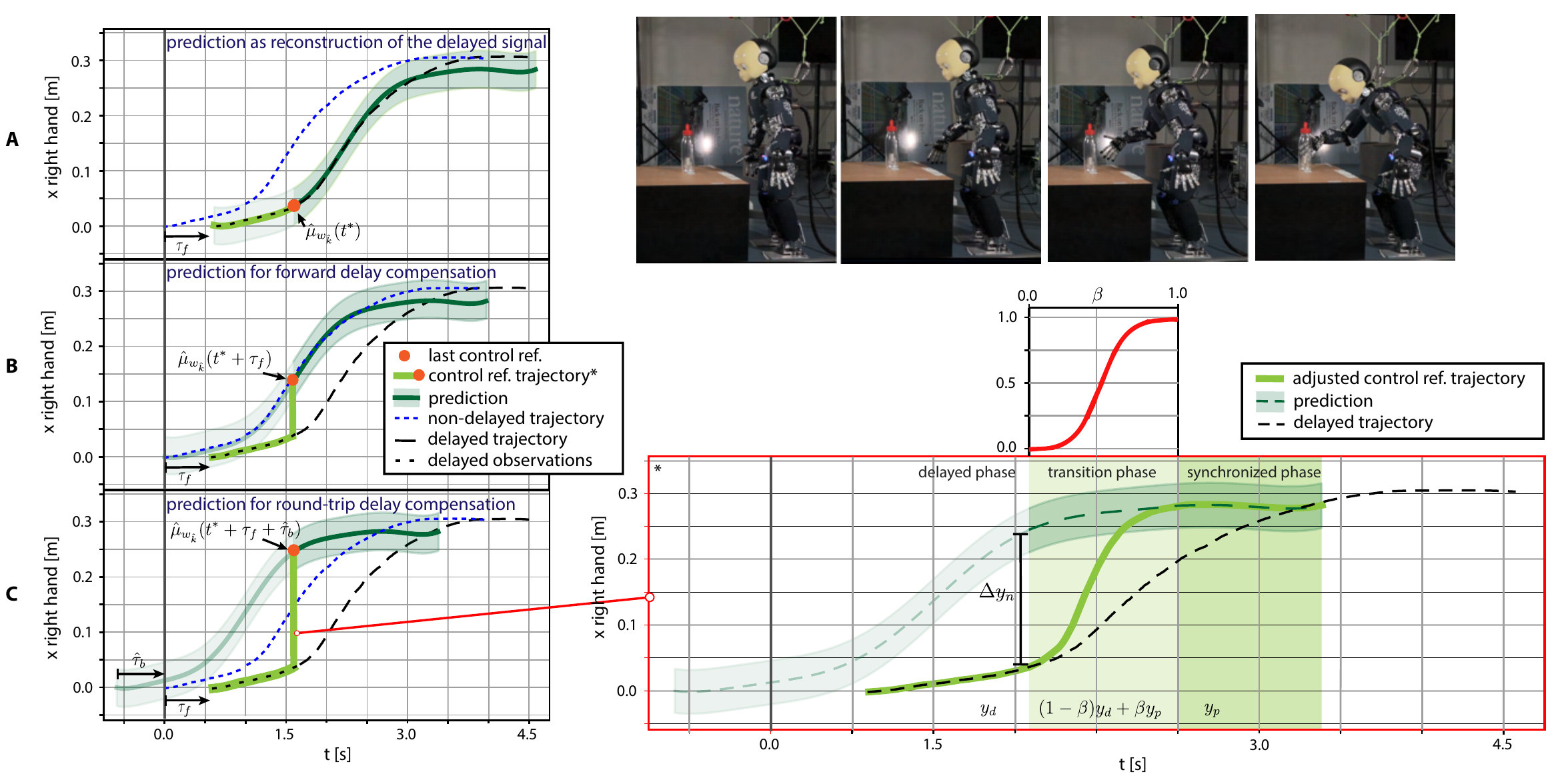}
    \caption{\textbf{Round-trip delay compensation.} Given the past delayed observations, the robot produces at each time a prediction ($\hat{\mu}_{w_k}$) of the current command. To compensate for the delays, the right sample (orange dot) from the prediction has to be selected as reference for the robot controller at each time. (A) The sample corresponding to the last received observation is an estimate of the delayed command. (B) By knowing the forward delay $\tau_f(t)$, a sample from the prediction can be selected so as to achieve a synchronization between the operator's movement and the robot movements.
    (C) By knowing the forward $\tau_f(t)$ and backward delay $\tau_b(t)$, the robot can select the right sample from its prediction so as to achieve a synchronization between the operator's movement and the feedback from the robot side. A policy blending arbitrates the delayed observations with the samples selected from the prediction, which guarantees a smooth transition from delayed to compensated teleoperation.}
    \label{fig:alpha}
\end{figure*}

\subsection*{Delay generation and compensation}
The time-varying round-trip delay over the network is made of a forward delay $\tau_f(t)$ in the communication from the operator to the robot, and a backward delay $\tau_b(t)$ between the robot and the operator. Each one-way delay is made of two components, one deterministic, mainly caused by the transmission and propagation time, and one stochastic \cite{Gurewitz2006}, often called the ``jitter''. In our experiments, we generate forward delays with a deterministic component between $100$ and $1000$ ms (depending on the experiment) and a stochastic component that follows a normal distribution (Methods). We generate similar deterministic backward delays but no stochastic backward delay, because we assume that the video streaming system implements a ``jitter buffer'' (Methods) that in effect transforms the jitter into an additional constant delay (Methods).
This is a reasonable assumption, as this is well implemented in existing video streaming systems (Methods).

The robot needs to know both the forward and backward delay. It computes the forward delay exactly (both in its deterministic and stochastic components) thanks to time-stamps attached to the packets sent by the operator and synchronized clocks (Methods). The robot can easily ask the operator's computer for the average backward delay because timestamps are included in most video streaming protocols (Methods), but it cannot know the stochastic part of the backward delay before sending the packet. This is why the robot relies on the jitter buffer on the operator side to transform this stochastic delay into a deterministic delay (Methods). In our implementation, we assume that the robot knows both the deterministic backward delay (average delay) and the length of the jitter buffer (and additional constant delay). In a deployed system, these data would be gathered by the operator's computer and sent periodically to the robot.

\begin{figure*}
\includegraphics[width=\linewidth]{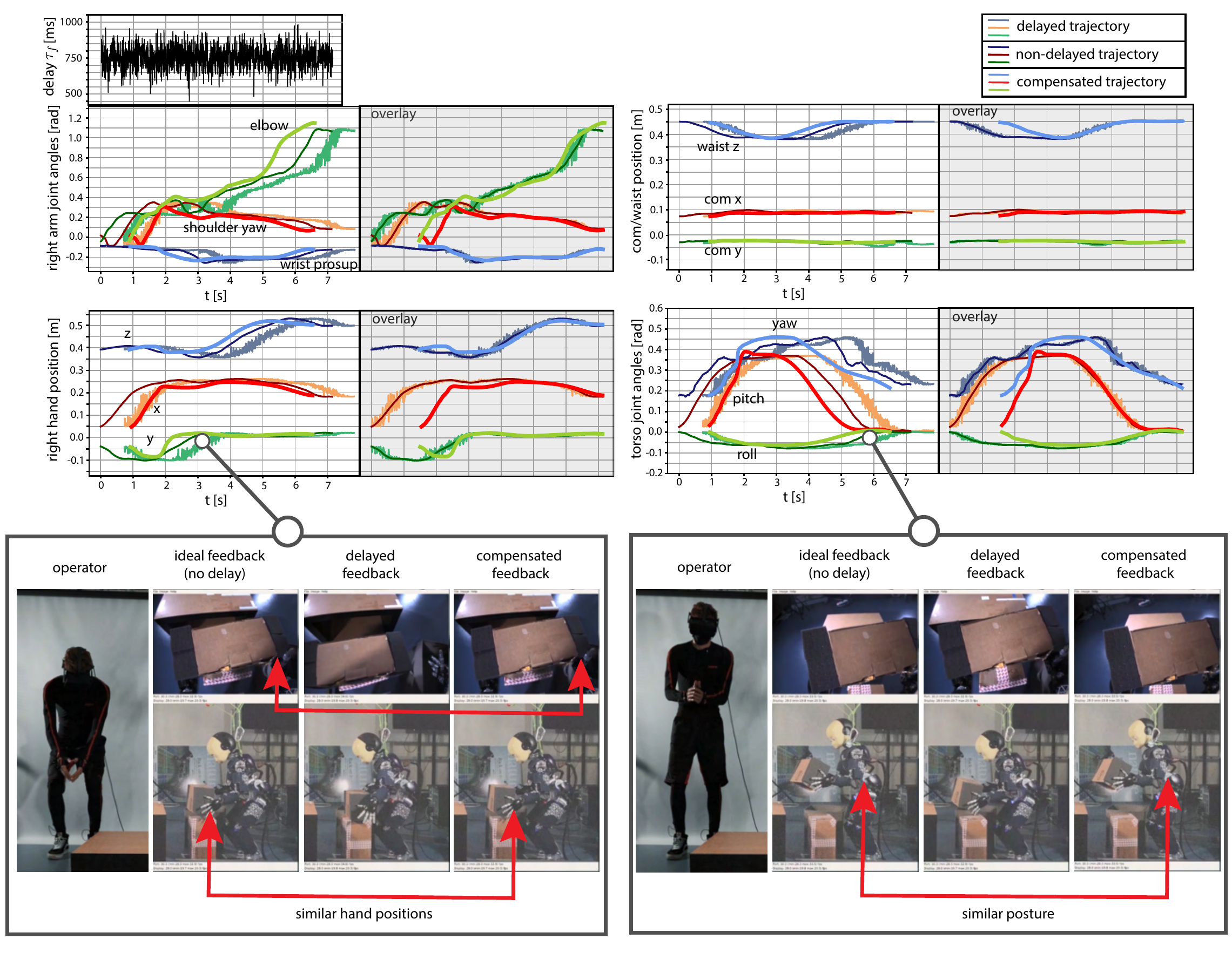}
\caption{\textbf{Teleoperation  with  compensation of a round-trip delay around 1.5s.} The robot is picking up a box in front of it at mid-height. The forward delay follows a normal distribution with 750ms as mean and 100ms as standard deviation, while the backward delay is 750ms. 
The trajectories retargeted from the human to the robot without any delay are the dark-colored red-green-blue lines. The orange-teal-gray lines are the corresponding delayed trajectories and the light-colored red-green-blue lines are the compensated trajectories.
The compensated trajectories (light red-green-blue lines) are the final robot reference trajectories. These, first follow the delayed teleoperated signals (orange-teal-grey lines). Then, when the prediction is available, they anticipate the teleoperated motion (dark-colored  red-green-blue lines) so to get a visual feedback at the user side coherent with what the operator is doing}.
\label{fig:tele15}
\end{figure*}

To compensate for the delay, once the robot has identified the best ProMP that matches the observations, it has to select the predictions that correspond to the right time-step from the mean of the ProMP.
By computing the round-trip delay, the robot chooses the right samples from the current movement prediction, i.e. posterior ProMPs' distributions (Fig. \ref{fig:alpha}). At the beginning of the motion, before any ProMP is recognized, the robot uses the delayed commands; however, once a ProMP is recognized, the robot can start compensating for the delays, but first the delayed trajectory needs to catch up with the ProMP. To keep the trajectory smooth, we take inspiration from the shared control literature \cite{LoseyAMR2018} and use blending between the current trajectory and the mean of the selected ProMP (Methods).

\subsection*{Prescient teleoperation experiments}

Overall, our \textit{prescient} whole-body teleoperation system relies on the following components:
\begin{itemize}
  \setlength\itemsep{0em}
    \item a whole-body controller based on quadratic optimization;
    \item a dataset of whole-body trajectories retargeted from human motions;
    \item a set of ProMPs that can predict future trajectories given observations;
    \item a computation of the round-trip delay to select the appropriate commands from the prediction, so that the robot anticipates both the operator-to-robot and the robot-to-operator delays;
    \item a blending to keep the trajectory smooth, at the start of a trajectory or in case of changes of delays;
    \item a video streaming system that uses a jitter buffer to cope with the stochastic part of the backward delay.
\end{itemize}

We evaluated the system on the iCub robot with a time-varying round-trip delay around 1.5s, given by a stochastic forward delay following a normal distribution with 750ms as mean and 100ms as standard deviation, and a constant backward delay of 750ms (Fig. \ref{fig:tele15} and Fig. S\ref{fig:tele15s}). We used a constant backward delay in these experiments because we relied on an existing video streaming system (Methods) that cannot artificially delay images randomly. Nevertheless, when a jitter buffer is used, the resulting video feedback is delayed by a constant delay (Methods), hence our implementation has not affected the final outcome of the experiments.
Additional experiments are presented in Supplementary
Materials, in which different tasks are performed under stochastic round-trip delays ranging from 200ms to 2s (Fig. S\ref{fig:tele200}-\ref{fig:tele2}). In all our experiments, the operator was able to successfully complete the tasks in spite of these large delays (Fig.~\ref{fig:tele15}) thanks to the compensation.  All the experiments can be seen in Movie M1.

If the operator decides to stop or to perform a different movement from the one that has begun, 
then the predicted trajectories are blended into the delayed trajectories in a way similar to that adopted to switch from delayed to predicted references (as shown in Fig. S\ref{fig:mistake}), hence avoiding any undesired prolonged mismatch (Methods). Movie M2 illustrates two examples of this behavior with the simulated robot.

To quantify the quality of the compensation, we compared the compensated trajectory to the non-delayed trajectory in the $20$ testing motions from the bottle reaching scenario of the dataset Multiple Tasks and for the $21$ testing motions from the box handling scenario of the dataset Multiple Tasks, with time-varying forward delay following a normal distribution with $750$ms as mean and $100$ms as standard deviation and backward delay being equal to $750$ms (Table \ref{tab:err}). The comparison is performed once the prediction is available and after the transition from delayed signals to prediction. In the box handling scenario, the results show that the error is around or less than $1$cm for all the considered references (in particular for the hands) whereas without compensation the error is about three times higher (about $3$cm for the hands). Similarly, in the bottle reaching task, the error of the compensated trajectory is about $1$ to $1.4$cm for the hands versus about $4$cm for the hands and $1$cm for the center of mass when there is no compensation. The angular errors show a similar pattern. While an error of about $1$cm is often enough to achieve a task, for instance grasping an object, an error of $3$ to $4$cm makes it very likely to miss the object, in addition to frustrating and disorienting the operator. 

We then evaluated the performance of the compensation when the communication delay increases (Fig. \ref{fig:delaycomparison}A). To do so, we compared the compensated trajectory to the non-delayed trajectory, for the right hand, in the task of reaching the bottle on the table of the dataset Multiple Tasks (Fig. S\ref{fig:tests1}). During the synchronization, the error is roughly proportional to the delay (Fig. \ref{fig:alpha}), which adds up to the prediction errors. In that case, we observe a mean error of about $2.5$cm for $1$s delay,  but more than $10$cm for a $3$s delay, because the transition time takes a significant amount of time on a short trajectory ($30\%$ for a delay of $3$s and a trajectory of $10$s). In longer movements or in an actual teleoperation session where the operator commands longer sequences of movements, this synchronization time should become negligible. When we exclude the synchronization time, the results show that the compensated tracking error is less than $2$cm for delays around $0.5$s, about $2.5$cm for delays around $1.5$s, and increases to about $5$cm for delays of $3$s and $4$s. Qualitatively, the operator found the system difficult to use with a delay of more than $2$s, for which the error is about $3$cm after the transition and around $7$cm including the transition.

\subsection*{Conforming the teleoperation to the intended motion}
Fig. S\ref{fig:cuvsmean} shows that the user, and therefore the robot, can pick up a box by bending the back without using the legs, by bending the legs and limiting back movements, or any other way in between. This means that the robot is not following the mean trajectories learned from the training demonstrations, but that it 
adapts the teleoperation trajectories to conform with the received data.
Such a feature can be obtained if the delay represents a reasonable portion of the whole motion duration, that in our experiments was a delay around 1.5s for motions with a  duration around 6s/7s.
When the delay increases to a significant portion of the complete motion duration, the early delayed observations may not be sufficient to update the posterior distribution of the ProMPs following the actual operator's commands (especially when the demonstrations exhibit a late variance). In this case (Fig. S\ref{fig:cuvsmean2}), we obtain a robot behavior that tends to follow the mean of the demonstrations for a significant part of the task. This is visible in Movie M1 at 1min53s) where, for a round-trip delay around 2s, the robot picked up the box by bending the knees while the human was only using the back without moving the legs.

To experiment explicitly with this adaptation of motion, we teleoperated the robot in simulation to reach a bottle located on the table in 3 different ways during training (Fig. S\ref{fig:dataObst}, dataset ``Obstacles''). During testing, we evaluated our compensation approach on the same task but avoiding different obstacles (Fig. S\ref{fig:testObst}). As with the previous dataset, the prediction continuously improves as more observations are available, trying to fit the observed operator commands (Fig. S\ref{fig:predcomp}-\ref{fig:predex}): the quality of the prediction improves over time in a similar fashion to the predicted trajectories from the bottle reaching scenario in Fig. S\ref{fig:tests1}, reaching about the same accuracy after observing half of the motion. To evaluate how the resulting compensation performs with different delays, we compared the compensated trajectories for the right hand position to the corresponding non-delayed ones in the 9 testing motions (Fig. \ref{fig:delaycomparison}B). The results show that the error in position is comparable (if higher, only by about half a centimeter) to the error from the bottle reaching scenario, where the same obstacles were used for training and testing.

Some examples of the compensated trajectories are illustrated in Fig. S\ref{fig:custom15}, where the robot is reaching the bottle while trying to avoid new obstacles. After an initial recognition phase, the compensated trajectories follow the specific way the human is commanding the robot in order to avoid the different obstacles, which could have been hit if the mean trajectories from the demonstrations were used. For instance, if we look at the right shoulder yaw angle (top row, center, red line), we see that both the prediction and the command are far from the mean of the ProMP: when the operator approached the object by the side (see the red arrow in the picture of the robot), the system adapted to perform the motion in the same way as the human intends. Similar observations can be made for the other degrees of freedom (e.g., the $y$ position of the right hand: green line on the top left panel; the yaw of the torso: blue line on the top right panel). For reference, Fig. S\ref{fig:custom0} shows how the same trajectories fit almost exactly the actual trajectories retargeted from the human in the case of no delay, which confirms that when there is no delay the robot almost perfectly follows the orders from the operator, even if they do not correspond to any training trajectory.

\subsection*{Conforming the teleoperation to new goals}
Similarly to how we tested the adaptability of the approach to the specific way the operator is performing a given task, we also investigated its ability to adjust to new object locations. To do that, we recorded a third dataset (data ``Goals'', Methods) by teleoperating the robot in simulation to reach a bottle located onto the table in several positions (Fig. S\ref{fig:traindiff}), then we tested the approach while reaching the same bottle but at different locations (Fig. S\ref{fig:testdiff}).

\begin{figure*}
    \centering
    \includegraphics[width=\linewidth]{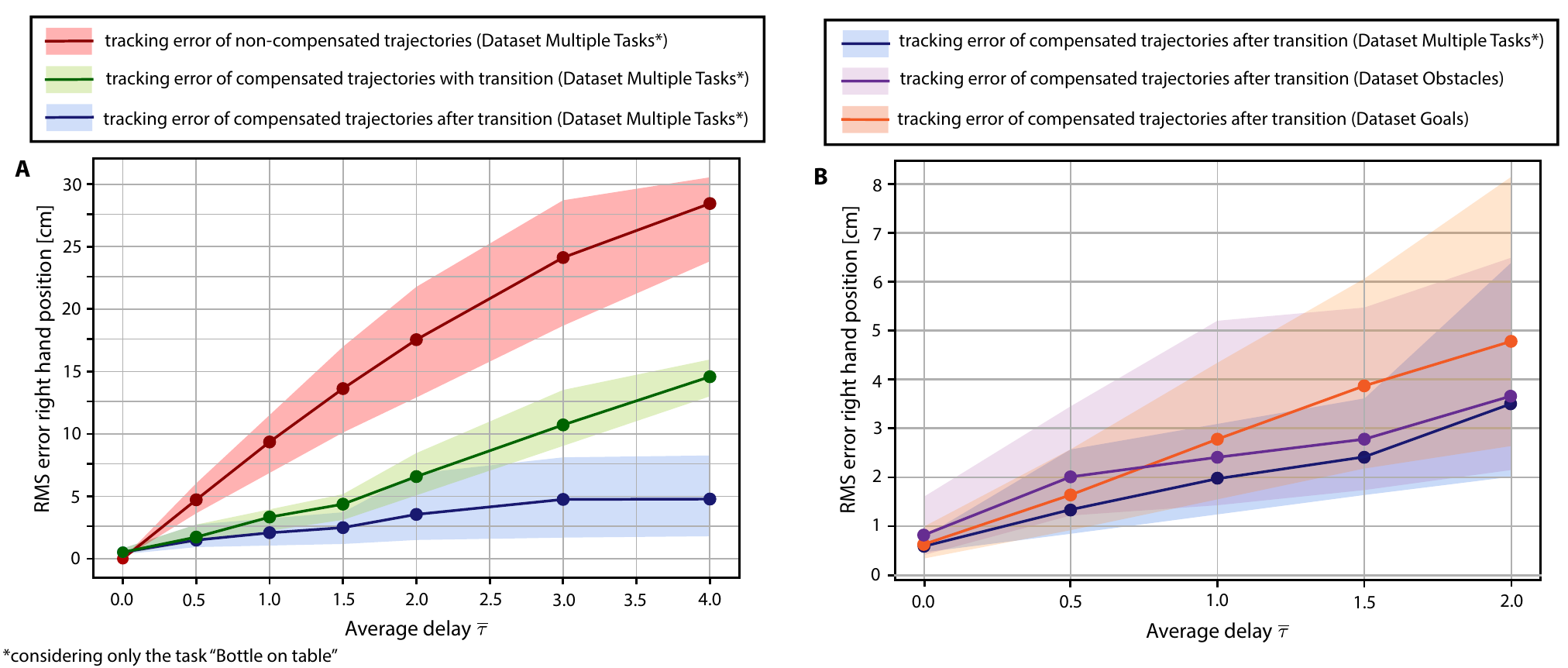}
    \caption{\textbf{Scalability of the delay compensation with respect to increasing time delays.} (A) Tracking error of the compensated trajectories for the right hand position with respect to the non-delayed ones compared to the tracking error of the corresponding non-compensated (delayed) trajectories with respect to the same non-delayed ones. The tracking error of the compensated trajectories is considered both including the transition from the delayed phase to the synchronization phase (Fig. \ref{fig:alpha}), which adds a non-compensable error, and without transition. The RMS of the error is computed from the $10$ testing motions of the task of reaching the bottle on the table from dataset Multiple Tasks (Fig. S\ref{fig:tests1}). (B) The tracking error of the compensated trajectories for the right hand position with respect to the non-delayed ones (after the transition phase) is evaluated also on the testing trajectories from the dataset ``Obstacles'' with different obstacles (Fig. S\ref{fig:testObst}) and the dataset Goals with different reaching goals (Fig. S\ref{fig:testdiff}). The tracking error is computed as the Euclidean distance between the evaluated trajectory and the reference trajectory. The compensated trajectories are temporally realigned with the non-delayed trajectories for computing the error, which is evaluated with different round-trip delays $\tau(t)$: $0$s, around $0.5$s, $1$s, $1.5$s, $2$s, $3$s and $4$s. The time-varying forward delay follows a normal distribution with mean $\overline{\tau_f}=\overline{\tau}/2$ and standard deviation equal to $\frac{2}{15}\overline{\tau}_f$. The backward delay is set equal to $\overline{\tau}_f$.}
    \label{fig:delaycomparison}
\end{figure*}

As one could expect, the prediction is less accurate than with the previous datasets where the goal position is always the same (S\ref{fig:predcomp}-\ref{fig:predexdiff}).  During the initial recognition phase the prediction error is around $2.5$cm higher compared to the trajectories from Fig. S\ref{fig:tests1}, but it continuously improves as more observations are available, decreasing the error gap to $1$cm after observing half of the motion. For delays around $1.5$s, the average tracking error on the hand position is $4$cm (Fig. \ref{fig:delaycomparison}B), which is about a centimeter higher than the error with a delay of $2$s on the datasets in which the bottle is in the same position for both training and testing. 
The operator found it difficult to teleoperate the robot for delays greater than $1.5$s, but the approach can still be used with novel goals for lower delays, typically between $0.5$ and $1$s, when a similar accuracy to the other datasets is often obtained (Fig. \ref{fig:delaycomparison}B).

Some successful examples of compensated trajectories are illustrated in Fig. S\ref{fig:customdiff}, where the robot is reaching the bottle located at unexpected positions. After an initial recognition phase, the compensated trajectories try to follow the operator's commands, which are guiding the robot hand toward the new target. The figure shows that both the prediction and the commands are far from the mean of the ProMP, which means that using the mean trajectories from the training demonstrations instead of our approach, would have certainly made the robot miss the bottle.

\begin{table*}
\center
\caption{\textbf{Difference (root mean square error) with the non-delayed trajectories, for both the compensated and the non-compensated (delayed) trajectories (average delay: 1.5 s).} The error is computed for the 20 testing motions from the bottle reaching scenario of the dataset Multiple Tasks (including those from Fig.~S\ref{fig:tests1}), and for the 21 testing motions from the box handling scenario of the dataset Multiple Tasks (including those from Fig.~S\ref{fig:test2}). The compensated trajectories are temporally realigned with the non-delayed trajectories for computing the error, which is considered only once the prediction starts, and once the blended transition from delay to compensation is over (Fig.~\ref{fig:alpha}).  The time-varying forward follows a normal distribution with $750$ms as mean and $100$ms as standard deviation. The backward delay is set equal to $750$ms. Examples of compensated trajectories are displayed in Fig.~S\ref{fig:cuvsmean}.} 
\begin{tabular}{ccccc}
\toprule
& \multicolumn{2}{c}{Box handling} & \multicolumn{2}{c}{Bottle reaching}\\
 &  \multicolumn{2}{c}{\textbf{RMS error [rad]}} &  \multicolumn{2}{c}{\textbf{RMS error [rad]}}\\
 & compensation &  no compensation & compensation &  no compensation \\
\cmidrule(lr){2-3}
\cmidrule(lr){4-5}
head yaw  & 0.024$\pm$0.011 & 0.035$\pm$0.012 
          & 0.013$\pm$0.007 & 0.021$\pm$0.011 \\
torso pitch  & 0.045$\pm$0.020 & 0.136$\pm$0.064 
             & 0.027$\pm$0.012 & 0.041$\pm$0.019   \\
torso roll  & 0.022$\pm$0.011 & 0.089$\pm$0.038 
            & 0.015$\pm$0.008 & 0.020$\pm$0.010  \\
torso yaw  & 0.069$\pm$0.028 & 0.129$\pm$0.055 
           & 0.019$\pm$0.009 & 0.032$\pm$0.011  \\
r. shoulder yaw  & 0.071$\pm$0.025 & 0.145$\pm$0.051 
                 & 0.065$\pm$0.018 & 0.221$\pm$0.092  \\
r. elbow  & 0.062$\pm$0.020 & 0.171$\pm$0.067 
          & 0.096$\pm$0.030 & 0.194$\pm$0.071  \\
r. wrist prosup.  & 0.025$\pm$0.007 & 0.077$\pm$0.033 
                  & 0.054$\pm$0.012 & 0.091$\pm$0.041\\
\cmidrule(lr){2-3}
\cmidrule(lr){4-5}
 &  \multicolumn{2}{c}{\textbf{RMS error [cm]}} &  \multicolumn{2}{c}{\textbf{RMS error [cm]}}\\
  & compensation &  no compensation & compensation &  no compensation \\
\cmidrule(lr){2-3}
\cmidrule(lr){4-5}
r. hand x  & 1.02$\pm$0.31 & 2.95$\pm$1.12 
           & 1.29$\pm$0.33 & 4.97$\pm$1.46   \\
r. hand y  & 0.90$\pm$0.26 & 3.36$\pm$1.21 
           & 1.21$\pm$0.31 & 4.33$\pm$1.17  \\
r. hand z  & 0.96$\pm$0.30 & 2.96$\pm$0.75 
           & 1.11$\pm$0.25 & 4.15$\pm$1.13   \\
com x  & 0.90$\pm$0.13 & 1.24$\pm$0.36 
       & 0.33$\pm$0.07 & 1.01$\pm$0.20   \\
com y  & 0.79$\pm$0.11 & 1.06$\pm$0.39 
       & 0.24$\pm$0.06 & 1.01$\pm$0.32  \\
waist z  & 0.88$\pm$0.14 & 2.02$\pm$1.22 
         & 0.22$\pm$0.07 & 0.61$\pm$0.09  \\
\bottomrule
\end{tabular}
\label{tab:err}
\end{table*}

\section*{Discussion} 
\label{sec:discussion}
Humanoids can only be deployed at their full potential if they can exploit their whole-body to perform non-trivial, high-value tasks. Whole-body teleoperation is the ideal framework to achieve this goal because it provides an intuitive and flexible approach to operate the robot, provided that the operator can rely on a synchronized feedback. By leveraging machine learning to anticipate the commands of the operator, we showed that it is possible to compensate for delays of 1-2 seconds, which typically correspond to the round-trip communication time between Earth and space \cite{lii2015} and between continents on the Internet \cite{hoiland2016measuring}.

To achieve the synchronization between the human motion and the visual feedback as soon as possible, our approach strongly relies on fast motion recognition. In fact, if this step were to take too long, the motion would be over by the time any compensation algorithm could be applied. 

In our experiments we leveraged the ProMPs to represent human gesture primitives and also to predict the operator's intention of movement. In the past, we had already used ProMPs with success for real-time prediction of whole-body movements \cite{dermy2017} and robot's gestures \cite{dermy2018prediction}. They only require a few robot/operator demonstrations to be trained, and the posterior update that enables to predict the future trajectory also requires a relatively small set of observations. For the above elements, they are a suitable and valid technique for predicting the operator's intention in our teleoperation problem. 
Other data-driven methods for predicting gestures as time-series could be used in principle, for example LSTM-RNN \cite{zhao2018,RNNestimation2020,corona2020context}, at the expense of more training data and potentially less smooth movements. Put differently, the concept of \emph{prescient teleoperation} can be implemented with any other predictor, including neural networks.

We showed that the robot follows the specific execution of motion, i.e., the particular way the human is performing the task, by continuously updating the prediction of the current motion (thanks to the conditioning operator of the ProMPs). In the experiments, we were able reach a bottle in ways that had not been demonstrated before (but included in the distribution of the training demonstrations), avoiding new obstacles and reaching new object positions that were not included during the training phase. This feature is extremely valuable because it allows the operator to adapt the commands on the fly to situations different from those in the training. For instance, there might be an obstacle that forces the operator to approach an object with the hand from the right, and prevent them from following a straight path. Or there might be a low ceiling that constrains the operator to bend the torso more or flex the legs more. More importantly, there are many different objects and many different environments and the operator will adapt their movement to the local condition, from a different target position to a different body posture. Thanks to a diverse set of ProMPs and the ``conformation'' of the prediction to the observations, our system should be able to handle most of these cases. The main limitation is that our approach assumes that the received data is enough to update the predictions; this is often the case, because humans tend to anticipate their change in a trajectory, that is, the beginning of the trajectory changes when a future goal changes. Nevertheless, the system is not capable of anticipating a last-second change when the beginning of the trajectory is always the same. Future work should evaluate how many demonstrations (and how many ProMPS) are needed to adapt to as many situations as possible.

As shown in Movie M1 and Fig. \ref{fig:tele15}, we were able to identify the current motion with less than a second of data, time during which the delayed signals were used to teleoperate the robots. However, our experimental setup included a limited amount of possible tasks (see Fig. \ref{fig:tasks}). If the dataset of possible tasks was larger, the recognition would take a considerable and prohibitive amount of time. In this case, one could resort to other data-driven prediction methods or exploit the fact that human actions are mostly vision-guided and identify the object that is about to be manipulated. In such a way, the set of possible tasks would be reduced to those related to that object, accelerating significantly the motion recognition. This can be done by using object recognition algorithms~\cite{tan2020} or by attaching ArUco markers~\cite{aruco} on the objects of interest. Then, if the number of tasks related to an object was similar to the number of tasks we considered in our experiments, the motion recognition would take an amount of time close to that of our experiments. Instead, if a considerably larger amount of tasks was related to an object, another strategy to reduce the recognition time could be to examine first the initial position of the hand trajectories in order to limit the recognition computation to those tasks having those initial positions in their distribution.

The proposed compensation algorithm also relies on the fact that the delay is not comparable to the duration of motion. Indeed, it would not be possible to apply any compensation, if the motion was over before computing any prediction. This is an additional reason for executing the teleoperated tasks at a slow pace (the humanoid robot in our experiments is never teleoperated with fast or rough movements anyway, to avoid damaging the platform).
In our experiments we were able to compensate for the delays while following the way the human performs the task with delays around 1.5s. For delays around 2s, we were still able to compensate for the delay but the robot was not following the human precise movements, rather the mean of the learned motions. For longer delays the performance deteriorates critically, producing robot behaviors that cannot adequately compensate for the delay.

In addition, the execution speed of a task has to be similar to those recorded during the demonstrations. Once estimated the duration of the current motion, our algorithm does not take into account any duration variation while carrying out the task.
A real-time adaptation of the speed execution could be addressed by constantly updating the time modulation of the ProMPs based on the last observations and updating the posterior distribution of the original learned ProMPs based on these observations \cite{dermy2017}.


Finally, all the experiments reported here were performed by a highly experienced user, who is very familiar with the teleoperation system and the iCub robot. While novice users 
would not have the same proficiency with the proposed system, they are very unlikely to be trusted to operate a highly expensive humanoid robot in a high-value mission, such as an intervention in a damaged chemical plant or in a remote Moon base. Like drone pilots who are trained extensively before their first mission, one should expect actual humanoid operators to be expert users with a long training period.

The next natural step is to include haptic feedback in addition to visual feedback. First, whole-body feedback would require the operator to wear a full-body exoskeleton \cite{ramos2019} or to use less intuitive distributed vibrotactile feedback \cite{brygo2014b}.
In bilateral systems the force signal is directly coupled between robot and human \cite{ramos2019}, while in master-slave systems the human operator can receive kinesthetic cues not directly related to the contact force being generated by the robot, or as indirect forces in a passive part of her/his body \cite{brygo2014b}. Our approach could certainly be extended to the latter case, where ProMPs could be learned for the force signals. The extension to bilateral techniques represents a more complex challenge, since these systems can be very unstable under time-delays. Several approaches have been proposed to stabilize these systems under time delays \cite{xu2019,singh2020} but the extension to platforms with a high number of degrees of freedom like humanoid robots and their robustness to packet loss and jittering is still an open challenge.
A promising research avenue is to combine our approach with the one proposed in \cite{Valenzuela2019}, in which a haptic feedback is produced using the point cloud data obtained with an RGB-D camera. So far, our robot has been streaming images, but it could very similarly stream a point a cloud. Like with images, this point cloud would correspond to past perceptions of the robot but the operator would perceive it as synchronized because the robot anticipated the commands. The haptic feedback from the point cloud would then not appear to be delayed.

In this sense, both visual and haptic feedback could be improved by using more cameras. In our experiments, we had to use an external camera (in addition to the robot cameras) to provide a better situational awareness to the operator about the robot's status in the environment. In fact with the limited field of view of the cameras in the iCub's eyes, the human can hardly see the environment and the robot's body at the same time, which makes it extremely difficult to grasp objects without making any errors. More situation awareness could also be provided to the operator by integrating our method with already existing predictive-display-based techniques \cite{mitra2008}: during the non-compensated intervals the predictive display could guide the human operator with a gradual shift from the virtual graphics to the real images of the robot cameras once the synchronization with the user commands has been achieved by our approach. 


Overall, this new approach will help deploy more robust, effective teleoperated systems. It is demonstrated here with a motion capture suit, a virtual reality headset and a state-of-the-art humanoid robot; but this is a general framework that could be used in any other robot, from manipulators to cars, as any of these could all anticipate remote commands provided that there is a good enough predictor.

\section*{Materials and methods} 
\label{sec:methods}
\begin{figure*}
    \centering
    \includegraphics[width=\linewidth]{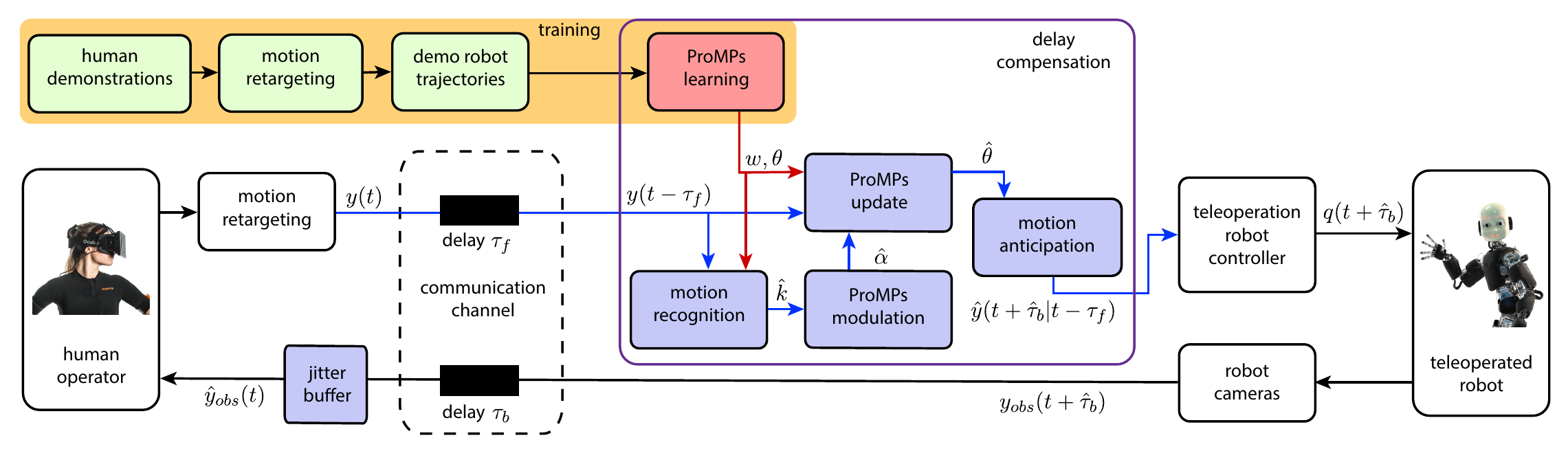}
    \caption{\textbf{Flowchart of the proposed teleoperation system.} During the training phase, the human operator teleoperates the robot without any significant delay (local network), and performs a variety of tasks. The retargeted human motions represent the robot trajectories demonstrations that are later used to train the ProMPs. A ProMP is learned for every task. When teleoperating the robot in a non-ideal network, the ProMPs are used to predict the robot movement: (1) the system recognizes the current task, that is it identifies the most likely ProMP; (2) it estimates the speed of the teleoperated motion and updates the selected ProMPs according to the delayed observed via-points; (3) it compensates for the delay by selecting the trajectory of the posterior ProMPs at the right timestep, so as to synchronize the user movements with what they see from the remote cameras at the robot side. The resulting trajectories are then tracked by the whole-body controller that computes the joint commands for the robot.}
    \label{fig:modules}
\end{figure*}

Our prescient teleoperation system combines six major components (Fig. \ref{fig:modules}): (i) the humanoid robot, (ii) its whole-body controller, (iii) the operator together with the equipment required to control and perceive the robot, (iv) the motion retargeting module, (v) the delayed network and (vi) the delay compensation module.

\subsection*{The iCub humanoid robot}
iCub \cite{natale2017icub} is a research-grade open-source humanoid robot designed by the Italian Institute of Technology (IIT) to experiment with embodied artificial intelligence. It measures 104 cm in height and weighs 22 kg, which roughly correspond to the body dimensions of a five year old child. iCub has 53 actuated degrees of freedom: 7 in each arm, 9 in each hand (3 for the thumb, 2 for the index, 2 for the middle finger, 1 for the coupled ring and little finger, 1 for the adduction/abduction), 6 in the head (3 for the neck and 3 for the cameras), 3 in the torso/waist, 6 in each leg. In this work, we do not use hands and we do not move the eyes, we therefore control 32 degrees of freedom. The head has stereo cameras in a swivel mounting in the corresponding location of the human eye sockets. iCub is also equipped with six 6-axial force/torque (F/T) sensors in the upper arms, legs and feet, an IMU in the head, and a distributed tactile skin.

\subsection*{Whole-body controller}
\label{sec:control}
The whole-body motion of the robot is defined by the following trajectories: center of mass ground projection, waist height, hands positions, arms postures, neck posture, torso posture,
which are either given by the delayed retargeted human motion, or generated by the delay compensation algorithm during the execution of the main task.
Each sample of each of these trajectories represents a control reference $\hat{\bm{y}}$. Given $\hat{\bm{y}}$, the robot commands $\bm{q}$ are generated by solving the redundant inverse kinematics, which can be formulated as a constrained quadratic programming problem with equality and inequality constraints \cite{escande2014hierarchical,pencoral}:
\begin{equation}
\begin{array}{c}
\text{arg}~\underset{\dot{\bm{q}}}{\text{min}} \sum_{i} w_i f_i  + \sum_{j} w_j g_j + \bm{C}\dot{\bm{q}}\\
~f_i=||\bm{J}_{i}\dot{\bm{q}}-\dot{\bm{x}}_i||^{2}\\
g_j=||\dot{\bm{q}_j}-\dot{\bm{q}}^r_j||^{2}\\
\text{subject to}~~\bm{J}\dot{\bm{q}}=\dot{\bm{x}} \\
~~~~~~~~~~~~\bm{A}\dot{\bm{q}}\le\bm{b}\\
\end{array}
\label{eq:control}
\end{equation}
The cost function consists of terms $f_i$ with relative weight $w_i$ concerning the pose of a specific body link $i$, where $\bm{J}_{i}$ is the Jacobian matrix for body link $i$ and $\dot{\bm{x}}_i=\dot{\hat{\bm{y}}}_i$ are the reference velocities for body link $i$. Additionally terms $g_j$ with relative weight $w_j$ concern the posture of certain joints $j$, where $\dot{\bm{q}}^r_j=\dot{\hat{\bm{y}}}_j$ are the reference joint velocities for joints $j$. $\bm{C}\dot{\bm{q}}$ is a regularization term used to get a unique solution and avoid singularities, where $\bm{C}$ is a linear cost vector.

In our implementation, we considered in the terms $f_i$ the hand positions with $w_i=1$ and the waist height with $w_i=0.65$.
Instead, the terms $g_j$ include the head posture with $w_j=1$ and the torso posture with $w_j=0.72$, the elbow and wrist postures with $w_j=0.11$. We computed optimal priorities with a multi-objective stochastic optimization that was run in simulation \cite{pencoral}. More details about the whole-body controller can be found in \cite{pencoral}.

The equality constraints correspond to the highest priority task, which should be solved exactly. 
In our implementation, these include the center of mass x position and the feet poses. The inequality constraints contain the robot joint velocity bounds and zero moment point bounds, which is constrained to stay inside the support polygon. The tracking accuracy of the proposed controller is reported in Fig. S\ref{fig:controller}.

Our controller is based on the OpenSOT framework \cite{rocchi2015opensot} and the qpOASES quadratic programming solver \cite{ferreau2014qpoases}.

This controller is run at $100$ Hz, which is the same frequency at which the motor commands are updated.

\subsection*{Motion retargeting}
\label{sec:mr}
The captured human information cannot be directly used as a reference for the humanoid, due to differences in kinematics (e.g. joint limits, body dimensions) and dynamics (e.g. mass distribution) between the human and the robot. Hence, motion retargeting is employed to map the human information into feasible reference trajectories for the robot. For transferring the translational movements of the end-effectors we used a fixed scaling factor (0.4).
The joint angles of the human joints instead are manually identified and mapped to the corresponding joints of the robot, as shown in Fig. S\ref{fig:retarget}.
The instantaneous reference value of the robot is then computed as:
\begin{equation}
	\Delta q_{i_R} = q_{0_R} + (q_{i_H} - q_{0_H})
\end{equation}
where $q$ is the vector of current joint positions, $\Delta q$ is the vector of joint variations with respect to the initial posture, the indices $0$ and $i$ refer to measurements at initial time and at time $i$, and the subindices $H$ and $R$ indicate measurements on human and robot, respectively. The same applies for the Cartesian positions of the end-effectors.

For the center of mass, the normalized offset-based reconstruction is used \cite{humanoids2018}.  We consider the ground projection of the human center of mass $\bfp_{com}^{g}$. Its position with respect to the left foot is projected onto the line connecting the two feet. The result is then normalized to obtain an offset $o\in[0,1]$
\[
o=\frac{(\bfp_{com}^{g}-\bfp_{lFoot}^{g})\cdot(\bfp_{rFoot}^{g}-\bfp_{lFoot}^{g})}{\mid\mid \bfp_{rFoot}^{g}-\bfp_{lFoot}^{g}\mid\mid^{2}}
\]
where $\bfp_{lFoot}^{g}$ and $\bfp_{rFoot}^{g}$ are the ground projections of the left and right foot respectively.
When the human is in a symmetric pose, the offset $o$ has a value around $0.5$ and when the human stands on a single foot, it is either $0$ (left foot) or $1$ (right foot).
The robot center of mass ground projection is then reconstructed on the line connecting its feet by means of this offset value.
To also retarget changes of the center of mass that are not on the line connecting the feet, we can apply the same concept while considering the maximum backward and forward center of mass displacement in the orthogonal direction of the line connecting the feet as done in \cite{humanoids2018}.




\subsection*{Hardware and communication setup}
The human motion is captured by the Xsens MVN system \cite{xsens}, which considers a human model comprising 66 degrees of freedom (corresponding to 22 spherical joints). The user teleoperating the robot is equipped with the wearable motion capture suit MVN Link, consisting of a Lycra suit with 17 inertial measurement units (IMUs) and a wireless data link transmitting at a frequency of 240Hz.
Our compensation method receives the delayed data from the motion capture system at 100Hz and transmits to the robot controller at 50Hz.

The user teleoperating the robot is also equipped with a VR Oculus headset. Through the headset, the operator can visualize the delayed images from both an external camera at the robot side, as a third-person view of the teleoperated robot, and the robot cameras, for a first-person immersive experience.
The communication protocol employed by the network is UDP with a bandwidth of 3Mbps. The forward delay is artificially generated, using the standard way to delay packets in Linux with the ``netem'' scheduling policy, which is based on the ``iproute2'' set of tools \cite{iproute2}.

The images from the cameras at the robot side are delayed by using the open source application Kinovea \cite{kinovea}, which allows the user to set a constant delay for the streaming of the video. The resulting delayed streaming is projected onto the VR headset through the application Virtual Desktop \cite{vrdesktop}.

\subsection*{Datasets}

\begin{table*}
    \center
    \caption{\textbf{Datasets used to train and test the approach.} Dataset Multiple Tasks has been used to train the approach and to perform the experiments on the real robot. The testing motions represent 9 different tasks and the repetitions of each individual task share the same goal but exhibit some movement variability with respect to those used for training. Dataset Obstacles has been used to train and test the approach in simulation with respect to the presence of unexpected obstacles. The testing scenarios have obstacles in between the robot and the bottle that were not considered during training. Dataset Goals has been used to train and test the approach in simulation with respect to unexpected changes of the goal position. The testing scenarios consider bottle locations different from those included in the training.}
    
    \begin{scriptsize}  
    \begin{tabular}{lcccccccc}
        \toprule
        \textbf{Dataset Multiple Tasks} &       &                                                              &                                   &       &       &       &            &                  \\
                                        &       & \multicolumn{6}{l}{\textbf{Bottle reaching}}                    &                                                                                           \\
                                        & TOTAL & \multicolumn{3}{c}{Bottle on a table}                          & \multicolumn{3}{c}{Bottle on a box}                                                         \\
        \cmidrule(lr){3-5}
        \cmidrule(lr){6-8}
        \#training motions              & 12    & \multicolumn{3}{c}{6}                                        & \multicolumn{3}{c}{6}                                                                     \\
        \#testing motions               & 20    & \multicolumn{3}{c}{10}                                       & \multicolumn{3}{c}{10}                                                                    \\

                                        &       & \multicolumn{7}{l}{\textbf{Box handling}}                                                                                                             \\
                                        &       & \multicolumn{3}{c}{Picking up a box}                           & \multicolumn{4}{c}{Placing a box}                                                           \\
        \cmidrule(lr){3-5}
        \cmidrule(lr){6-9}
                                        & TOTAL & low position                                                 & mid height                        & table & floor & table & inside box & operator's hands \\
        \#training motions              & 42    & 6                                                            & 6                                 & 6     & 6     & 6     & 6          & 6                \\
        \#testing motions               & 21    & 3                                                            & 3                                 & 3     & 3     & 3     & 3          & 3                \\
        \\

        \midrule

        \textbf{Dataset Obstacles}      &       &                                                              &                                   &       &       &       &            &                  \\
                                        & TOTAL & \multicolumn{3}{c}{Bottle on a table with different obstacles} &                                   &       &                                               \\
        \cmidrule(lr){3-5}
        \#training motions              & 6     & \multicolumn{3}{c}{6}                                        &                                   &       &                                               \\
        \#testing motions               & 9     & \multicolumn{3}{c}{9}                                        &                                   &       &                                               \\
        \midrule
        \textbf{Dataset Goals}          &       &                                                              &                                   &       &       &       &            &                  \\
                                        & TOTAL & \multicolumn{3}{c}{Bottle on a table at different positions}   &                                   &       &                                               \\
        \cmidrule(lr){3-5}
        \#training motions              & 7     & \multicolumn{3}{c}{7}                                        &                                   &       &                                               \\
        \#testing motions               & 10    & \multicolumn{3}{c}{10}                                       &                                   &       &                                               \\
        \bottomrule
    \end{tabular}
    
     \end{scriptsize} 
     
    \label{tab:datasets}
\end{table*}

To train our method, we teleoperated the robot in an ideal network without any delay and recorded the corresponding robot motion. Every demonstration contains several Cartesian and postural trajectories that determine the whole-body motion (Fig. \ref{fig:learned}): the center of mass ground projection, the waist height, the hands positions, the arms posture (shoulder rotation, elbow flexion, forearm rotation), the neck posture (flexion and rotation) and the torso posture (flexion, rotation and abduction). We record the retargeted trajectories, that is the reference trajectories for the whole-body controller of the robot.

We used three different datasets (Table \ref{tab:datasets}), each one divided into a training set and a test set.
The first dataset (Dataset Multiple Tasks) is designed to test how well the robot recognizes tasks and deals with the intrinsic variability of the operator's movements. This is the dataset used to perform the experiments on the real robot. The second one (Dataset Obstacles) is designed to evaluate the approach with unexpected obstacles. The third dataset (Dataset Goals) is designed to evaluate the approach with novel goal positions.

\paragraph{Dataset Multiple Tasks.}
This dataset covers two scenarios (Fig. \ref{fig:tasks}): reaching a bottle with the right hand and handling a box. 

The bottle task consists of demonstrations of 2 distinct whole-body reaching primitives: one primitive is for reaching the bottle on the table, the other one is for reaching the bottle on the top of the box (Fig. \ref{fig:tasks}). For each primitive, we recorded 6 repetitions of the task for training, for a total of 12 training whole-body demonstrations with an average duration of $6.1$s. Every demonstration provided by the human operator is different, since it is not possible to exactly reproduce the same whole-body movement twice; to further increase the variance of the demonstrated movements, in 3 repetitions out of the 6, an obstacle was placed in between the robot and the bottle (Fig. \ref{fig:tasks}).  
To assess the quality of the predictions, 10 different testing repetitions were recorded for each of the two primitives; 5 with the obstacle in between the robot and the bottle, and 5 without any obstacle (Fig. \ref{fig:tasks}), for a total of 20 motions. In this dataset, the obstacles are at the same positions in both the training set and the testing set (see the dataset ``Obstacles'' below).

The second scenario consists of demonstrations of 7 distinct whole-body box handling primitives: 3 for picking up the box --- from a low position, from a mid-height and from the table; 4 for placing the box at a specific location --- on the floor, on the table, inside a container, or in a person's hands (Fig. \ref{fig:tasks}). For each primitive, we recorded 6 different repetitions for training, for a total of 42 training whole-body demonstrations, with an average duration of $7.2$ s for the pick-up motions and of $5.8$ s for the box-placing motions.
For the test set, 3 new different repetitions of the 7 motions were recorded (Fig. S\ref{fig:test2}), for a total of 21 testing motions.
Examples of training motions are showed in Fig. S\ref{fig:learneds1} and Fig. \ref{fig:learned}, while the corresponding testing motions are illustrated in Fig. S\ref{fig:tests1} and Fig. S\ref{fig:test2}, respectively.

\paragraph{Dataset Obstacles.}
The training set is composed of 6 demonstrations of bottle reaching motions with 3 different locations of an obstacle (Fig. S\ref{fig:dataObst}): 2 repetitions without obstacles, 2 with an obstacle in between the robot and the bottle, and 2 with a different obstacle. The average duration of the demonstrations is around $6.9$s. The test set consists of motions for the same task but with obstacles at different locations (Fig. S\ref{fig:testObst}): 3 repetitions for each of the 3 distinct scenarios with different obstacles (Figure S\ref{fig:testObst}).

\paragraph{Dataset Goals.} 
The training set is composed of 7 demonstrations of bottle reaching motions (Fig. S\ref{fig:traindiff}), with the goal located in 7 different positions. The average duration of the demonstrations is $6.1$s. The test set consists of motions reaching the same bottle but at 10 different locations (Fig. S\ref{fig:testdiff}).

\subsection*{Probabilistic Movement Primitives (ProMPs)}
A ProMP \cite{paraschos2017} is a probabilistic model for representing a trajectory distribution. The movement primitive representation models the time-varying mean and variance of the trajectories and is based on basis functions.  
A single trajectory is represented by a weight vector $\bm{w}\in \mathbb{R}^m$. The probability of observing a trajectory $\bm{y}$ given the underlying weight vector is given as a linear basis function model
\begin{align}
&\bm{\xi}_t = \bm{\Phi}_t\bm{w}+\bm{\epsilon}_\xi, \\
&p(\bm{y}|\bm{w}) = \prod_{t}\mathcal{N}(\bm{\xi}_t|\bm{\Phi}_t\bm{w},\bm{\Sigma}_\xi),
\label{eq:promp}
\end{align}
where $\bm{\Sigma}_\xi$ is the observation noise variance, $\bm{\epsilon}_\xi\sim\mathcal{N}(0,\bm{\Sigma}_\xi)$ is the trajectory noise. The matrix $\bm{\Phi}_t \in \mathbb{R}^m$ corresponds to the $m$ normalized radial basis functions evaluated at time $t$, with
\begin{equation}
\phi_c(t) = \frac{\text{exp}\Big(-\frac{1}{2}\big(t-\frac{c-1}{m-1}\big)^2\Big)}{\sum_{c=1}^m\text{exp}\Big(-\frac{1}{2}\big(t-\frac{c-1}{m-1}\big)^2\Big)},
\label{eq:rbf}
\end{equation}
where the variable $c \in \{1,2,...,m\}$ specifies the center of each basis function.
The distribution $p(\bm{w};\bm{\theta})$ over the weight vector $\bm{w}$ is Gaussian, with parameters $\bm{\theta}=\{\bm{\mu_w},\bm{\Sigma_w}\}$ specifying the mean and the variance of $\bm{w}$.
The trajectory distribution $p(\bm{y};\bm{\theta})$ is obtained by marginalizing out the weight vector $\bm{w}$, i.e.
\begin{equation}
p(\bm{y},\bm{\theta}) = \int{p(\bm{y}|\bm{w})p(\bm{w};\bm{\theta})d\bm{w}}.
\label{eq:promp2}
\end{equation}

\subsection*{Delay generation}

The round-trip delay $\tau(t)$ at time-step $t$ is divided into a forward $\tau_f(t)$ (operator to robot) and a backward delay $\tau_b(t)$ (robot to operator):
\begin{displaymath}
\tau(t)=\tau_f(t)+\tau_b(t)
\end{displaymath}

Each one-way delay is composed of two parts, one deterministic and one stochastic \cite{Gurewitz2006}. The deterministic component corresponds to the transmission and propagation delays. It does not change when all the transmitted packets have the same format and size and use same physical link \cite{Gurewitz2006}. The stochastic part, often called the ``jitter'' is mainly associated with the queueing delay \cite{garetto2003} and varies from packet to packet, even when the packets have the same size and format.

If we denote by $\tau_{f,D}$ the deterministic part of $\tau_f$ and by $\tau_{f,S}$ the stochastic part:
\begin{eqnarray}
    \tau_f(t) &= \tau_{f,D} + \tau_{f,S}\\
    \tau_b(t) &= \tau_{f,b} + \tau_{b,S}
\end{eqnarray}

In our experiments, we generate a forward delay that follows a normal distribution:
\begin{displaymath}
\tau_f(t) = \tau_{f,D} + \mathcal{N}\big(0, \sigma_{\tau_f}\big)
\end{displaymath}

For both simulations and real experiments, we set the deterministic part of the forward delay $\tau_{f,D}$ between $100$ms to $1$s (depending on the experiment, see the captions of each figure) and the jitter $\sigma_{\tau_f}$ to $\frac{2}{15} \tau_{f,D}$, which is in line with what the jitter usually represents \cite{alharbi2017}.

\begin{eqnarray}
    \tau_{f,D} &\in& [100, 1000] \textrm{ (depending on the experiment)}\\
    \sigma_{\tau_f} &=& \frac{2}{15} \tau_{f,D}
\end{eqnarray}

For the stochastic part of the backward delay, we assume that the robot uses a video streaming software that implements a jitter buffer (sections ``Delay estimation by the robot'' and ``Jitter buffer''), which is the case of all the modern video streaming systems. If we set the jitter buffer length to $d$, then this buffer adds an additional deterministic delay of $d$: all the packets that arrive before $d$ seconds are re-ordered and packets that are not arrived are dropped (dropping a few frames has little consequence for a state-of-the-art video codec). As a consequence, from the operator perspective, the backward delay is constant. This is why, for both simulations and real experiments, we generate a constant backward delay:

\begin{equation}
    \tau_{b}(t)=\tau_{b,D}(t)
\end{equation}

For simplicity, we set $\tau_{b}(t) = \tau_{f,D}$ in all our experiments, but nothing in our system requires these two delays to be equal; in particular, it would be equivalent to set $\tau_{b}(t)=\tau_{b,D}(t) + K$ with $K$ any constant delay caused by the jitter buffer.

\subsection*{Delay estimation by the robot}
We assume that the clocks of the robot and of the computer of the operator are synchronized. In our real experiments, we synchronize the clock using the NTP system \cite{RFC5905}, which is the standard Unix protocol for time synchronization. The two clocks typically differ from less than $1$ms on a local network \cite{veitch2004}. Alternatively, GPS receivers can provide a highly accurate and absolute clock reference with an error of a few nano-seconds \cite{veitch2004}.

We add a time-stamp to each of the packets sent by the operator, which makes it possible for the robot to compute the forward delay $\tau_f(t)$ (this includes both the deterministic and stochastic part) when a packet is received at time $t$ :
\begin{equation}
    \tau_f(t) = \textrm{clock}_{\textrm{robot}}(t) - \textrm{timestamp}_{\textrm{operator}}(t)
\end{equation}
Please note that this does not assume that the delay follows a normal distribution. If necessary, the robot can re-order the packets according to the time-stamps to condition the trajectory predictions.

While the robot needs an estimate of the backward delay, it cannot know in advance the stochastic part before sending it. Our approach is to rely on the jitter buffer (section ``Jitter buffer'') implemented in the video streaming system to make the backward delay deterministic: if we set the jitter buffer length to $d$ s on the operator receiving side, then we know that the delay caused by the jitter will be exactly equal to $d$. 

In our experiment, we therefore assume that the backward delay is known and constant (100 ms, 250ms, etc. depending on the experiments). To keep the implementation simple and easy to reproduce, we assumed that the deterministic backward delay was always equal to the deterministic forward delay (a reasonable assumption given that the same link is used for both directions) and that the stochastic part is negligible (because we chose to not add any jitter on the backward delay in the robot experiment, see the Jitter buffer section below):

\begin{eqnarray}
    \tau_b(t) &= \overline{\tau_f(t)} = \tau_{f,D}(t)\\
    \tau_{b,S}(t) &= 0    
\end{eqnarray}

In a deployed setup, the robot would benefit from a better estimate of the average backward delay (the deterministic part) and the length of the jitter buffer. To do so, most video streaming systems use the RTCP protocol \cite{RFC3550} to get out-of-band statistics and control information for a video streaming session that follows the RTP protocol \cite{Ott2006ExtendedRP}. This data would need to be sent periodically from the operator's computer to the robot so that the robot knows both $d$ and $\tau_{b,D}$ (which are not expected to change at high speed). Alternatively, a custom system can be designed by using time-stamps on the image packets to gather statistics about the delay.

\subsection*{Jitter buffer}
The jitter buffer is the component of a video streaming system \cite{claypool1999effects,oklander2008} that re-orders packets if they are delayed by less than the length of the buffer and drops packets that are too late. Much work has been dedicated to adapt its length automatically \cite{oklander2008}: if it is too small, then the video is jittery, but if it is too large, delays are added, which is detrimental to the user (in particular during video calls). In our system, we assume that the length is fixed and known to the robot, for simplicity.

We did not implement a jitter buffer because we wanted to avoid modifying the video streaming system: reordering or dropping packets would require a considerable expertise in the internals of both the encodings (e.g., MP4) and the video streaming software. Instead, we consider that video streaming with delay and jitter is a problem that is well solved by all the current video streaming systems, as experienced by the million of users who watch videos online on smartphones with non-ideal connections. 

To summarize, from the point of view of our system, the jitter buffer results in an additional but deterministic delay. However, we assume that the robot knows the value of this additional delay as well as the deterministic part of the delay.

\subsection*{Probabilistic Movement Primitives (ProMPs)}
A ProMP \cite{paraschos2017} is a probabilistic model for representing a trajectory distribution. The movement primitive representation models the time-varying mean and variance of the trajectories and is based on basis functions.  
A single trajectory is represented by a weight vector $\bm{w}\in \mathbb{R}^m$. The probability of observing a trajectory $\bm{y}$ given the underlying weight vector is given as a linear basis function model
\begin{align}
&\bm{\xi}_t = \bm{\Phi}_t\bm{w}+\bm{\epsilon}_\xi, \\
&p(\bm{y}|\bm{w}) = \prod_{t}\mathcal{N}(\bm{\xi}_t|\bm{\Phi}_t\bm{w},\bm{\Sigma}_\xi),
\label{eq:promp}
\end{align}
where $\bm{\Sigma}_\xi$ is the observation noise variance, $\bm{\epsilon}_\xi\sim\mathcal{N}(0,\bm{\Sigma}_\xi)$ is the trajectory noise. The matrix $\bm{\Phi}_t \in \mathbb{R}^m$ corresponds to the $m$ normalized radial basis functions evaluated at time $t$, with
\begin{equation}
\phi_c(t) = \frac{\text{exp}\Big(-\frac{1}{2}\big(t-\frac{c-1}{m-1}\big)^2\Big)}{\sum_{c=1}^m\text{exp}\Big(-\frac{1}{2}\big(t-\frac{c-1}{m-1}\big)^2\Big)},
\label{eq:rbf}
\end{equation}
where the variable $c \in \{1,2,...,m\}$ specifies the center of each basis function.
The distribution $p(\bm{w};\bm{\theta})$ over the weight vector $\bm{w}$ is Gaussian, with parameters $\bm{\theta}=\{\bm{\mu_w},\bm{\Sigma_w}\}$ specifying the mean and the variance of $\bm{w}$.
The trajectory distribution $p(\bm{y};\bm{\theta})$ is obtained by marginalizing out the weight vector $\bm{w}$, i.e.
\begin{equation}
p(\bm{y},\bm{\theta}) = \int{p(\bm{y}|\bm{w})p(\bm{w};\bm{\theta})d\bm{w}}.
\label{eq:promp2}
\end{equation}


\subsection*{Learning ProMPs from demonstrations}
\label{sec:learn}
The demonstrations are trajectories retargeted from the human. These are recorded in an ''offline phase'', while the user teleoperates the robot within a local network (approximately without delays) to perform a variety of tasks.
Since the duration of each recorded trajectory may be different, a phase variable $\upsilon \in [0,1]$ is introduced to decouple the movement from the time signal, obtaining a common representation in terms of primitives that is duration independent \cite{dermy2017}. For each task, the modulated trajectories $\bm{\xi}_i(\upsilon)$ are then used to learn a ProMP. 
The parameters $\bm{\theta}=\{\bm{\mu}_w,\bm{\Sigma}_w\}$ of the ProMP are estimated by using a simple maximum likelihood estimation algorithm. For each demonstration $i$, we compute the weights with linear ridge regression, i.e.
\begin{equation}
\bm{w}_i=\big(\bm{\Phi}_\upsilon^\top\bm{\Phi}_\upsilon+\lambda\big)^{-1}\bm{\Phi}_\upsilon^\top\bm{\xi}_i(\upsilon),
\label{eq:promp3}
\end{equation}
where the ridge factor $\lambda$ is generally set to a very small value, typically $\lambda=10^{-12}$ as in our case, as larger values degrade the estimation of the trajectory distribution.
Assuming Normal distributions $p(\bm{w})\sim\mathcal{N}(\bm{\mu}_w,\bm{\Sigma}_w)$, the mean $\bm{\mu}_w$ and covariance $\bm{\Sigma}_w$ can be computed from the samples $\bm{w}_i$:
\begin{equation}
\bm{\mu_w}=\frac{1}{D}\sum_{i=1}^{D} \bm{w}_i, ~~~~\bm{\Sigma_w}=\frac{1}{D}\sum_{i=1}^{D} (\bm{w}_i-\bm{\mu_w})(\bm{w}_i-\bm{\mu_w})^\top,
\label{eq:meanvariance}
\end{equation}
where $D$ is the number of demonstrations. Since a whole-body trajectory is represented by $N$ trajectories ($x$, $y$, $z$ position of the center of mass, of the hands, etc.), we learn a ProMP for each of the $N$ trajectories. These ProMPs all together encode the same task $k$.

\subsection*{Recognizing the category of motion}
\label{sec:recognition}
Each learned $k$-th set of $N$ ProMPs encodes different whole-body trajectories to accomplish a given task like picking up a box or squatting. 
To recognize to which set $k$ the current teleoperated motion belongs to, we can minimize the distance between the first $n_{obs}$ delayed observations and the mean of the $N$ ProMPs of a group $k$, as done in \cite{dermy2017}:
\begin{equation}
\hat{k}= \text{arg}\min_{k \in [1:K]}\bigg[ \sum_{n=1}^{N}\sum_{t\in T_{obs}}|\bm{y}_n(t-\tau_f(t))-\Phi_{n,t-\tau_f(t)}\bm{\mu}_{n,\bm{w}_{k}}|\bigg],
\label{eq:distance}
\end{equation}
where $K$ is the number of tasks in the dataset and $T_{obs}=\{t_1,...,t_{n_{obs}}\}$ is the set of timesteps associated to the $n_{obs}$ early observations. 
While computing $\hat{k}$, the ProMPs are modulated to have a duration equal to the mean duration of the demonstrations.
The recognition (\ref{eq:distance}) starts whenever a motion is detected, i.e. the derivative of the observed end-effector trajectories exceeds a given threshold. 
The distance in (\ref{eq:distance}) is continuously computed after having identified the current motion. In this way, we can verify that the observations do not diverge from the demonstrations (exceed by a given threshold the demonstrated variance), in which case a gradual switch to delayed teleoperation is performed.

\subsection*{Time-modulation of the ProMPs}
\label{sec:modulation}
During motion recognition, we assumed that the duration of the observed trajectories is equal to the mean duration of the demonstrated trajectories, which might not be true.
To match as closely as possible the exact speed at which the movement is being executed by the human operator, we have to estimate the actual trajectory duration (Fig. S\ref{fig:timemod}). More specifically, we want to find the time modulation $\alpha$, that maps the actual duration of a given (observed) trajectory to the mean duration of the associated demonstrated trajectories.

During the learning step, for each $k$-th set of ProMPs we record the set of $\alpha$ parameters associated to the demonstrations: $S_{\alpha k}=\{\alpha_1, ..., \alpha_n\}$. Then, from this set, we can estimate which $\alpha$ better fits the current movement speed.
We considered the best $\hat{\alpha}$ to be the one that minimizes the difference between the observed trajectory and the predicted trajectory for the first $n_{obs}$ datapoints:
\begin{equation}
\hat{\alpha}= \text{arg}\min_{\alpha \in S_{\alpha \hat{k}}}\bigg\{ \sum_{t\in T_{obs}}|\bm{y}(t-\tau_f(t))-\Phi_{\alpha (t-\tau_f(t))}\bm{\mu}_{\bm{w}_{\hat{k}}}|\bigg\}.
\label{eq:distance}
\end{equation}

\subsection*{Updating the posterior distribution of the ProMPs}
\label{sec:prediction}
Once the $\hat{k}$-th most likely set of ProMPs and their duration have been identified, we continuously update their posterior distribution to take into account the observations that arrive at the robot side.
Each ProMP has to be conditioned to reach a certain observed state $\bm{y}^*_{t}$.
The conditioning for a given observation $\bm{x}^*_{t}=\{\bm{y}^*_{t},\bm{\Sigma}^*_{y}\}$ (with $\bm{\Sigma}^*_{y}$ being the accuracy of the desired observation) is performed by applying Bayes theorem
\begin{equation}
p({\bm{w}_{\hat{k}}}|\bm{x}^*_{t})\propto \mathcal{N}(\bm{y}^*_{t}|\bm{\Phi}_{\hat{\alpha}t}{\bm{w}_{\hat{k}}},\bm{\Sigma}^*_y)p({\bm{w}_{\hat{k}}}).
\label{eq:bayes}
\end{equation}
The conditional distribution of $p({\bm{w}_{\hat{k}}}|\bm{x}^*_{t})$ is Gaussian with mean and variance
\begin{align}
&\hat{\bm{\mu}}_{\bm{w}_{\hat{k}}}=\bm{\mu_{\bm{w}_{\hat{k}}}}+\bm{L} \big(\bm{y}^*_{t}-\bm{\Phi}_{\hat{\alpha}t}^\top\bm{\mu_{\bm{w}_{\hat{k}}}} \big), \\
&\hat{\bm{\Sigma}}_{\bm{w}_{\hat{k}}}=\bm{\Sigma_{\bm{w}_{\hat{k}}}}-\bm{L}\bm{\Phi}_{\hat{\alpha}t}^\top\bm{\Sigma_{\bm{w}_{\hat{k}}}},
\label{eq:munew}
\end{align}
where
\begin{equation}
\bm{L}=\bm{\Sigma}_{\bm{w}_{\hat{k}}}\bm{\Phi}_{\hat{\alpha}t} \big(\bm{\Sigma}^*_y +\bm{\Phi}^\top_{\hat{\alpha}t}\bm{\Sigma}_{\bm{w}_{\hat{k}}}\bm{\Phi}_{\hat{\alpha}t} \big)^{-1}.
\label{eq:L}
\end{equation}
Given the delay in the transmitted data $\tau_f(t)$, we can compute the timestep $t^*$ at which the ProMP has to be conditioned to a certain observation $\bm{x}^*_{t}$:
\begin{equation}
t^*=t-\tau_f(t)-t_0.
\label{eq:tstar}
\end{equation}
where $t_0$ is the starting time of the current motion.

\subsection*{Motion anticipation}
\label{sec:compensation}
The references for the robot controller are generated at each time based on the updated ProMPs' mean trajectories $\hat{\bm{\mu}}_{\bm{w}_{\hat{k}}}$. For a given ProMP, the sample $\hat{\bm{\mu}}_{\bm{w}_{\hat{k}}}(t^*)$ corresponding to the last conditioned observation, is a reconstruction of the past retargeted human input
\begin{equation}
\hat{\bm{\mu}}_{\bm{w}_{\hat{k}}}(t^*)=\hat{\bm{y}}(t-\tau_f(t)).
\end{equation}
The sample $\hat{\bm{\mu}}_w(t^*+\tau_f(t))$ is an estimate of the current retargeted human input \begin{equation}
\hat{\bm{\mu}}_{\bm{w}_{\hat{k}}}(t^*+\tau_f(t))=\hat{\bm{y}}(t),
\end{equation}
and can be used to synchronize the human movement with that of the robot, compensating only the forward delay (see Fig. \ref{fig:modules}). In our case, we want to synchronize the motion  of the human operator with what is seen from the robot side, thus compensating for both the forward and backward delays. To do so, we select the sample $\hat{\bm{\mu}}_w(t^*+\tau_f(t)+\hat{\tau}_b(t))$ as a control reference, which corresponds to a future prediction of the retargeted human movements:
\begin{equation}
\hat{\bm{\mu}}_{\bm{w}_{\hat{k}}}(t^*+\tau_f(t)+\hat{\tau}_b(t))=\hat{\bm{y}}(t+\hat{\tau}_b(t)).
\label{eq:yhat}
\end{equation}
The remaining samples $[\hat{\bm{\mu}}_{\bm{w}_{\hat{k}}}(t^*+\tau_f(t)+\hat{\tau}_b(t)+1),\hat{\bm{\mu}}_{\bm{w}_{\hat{k}}}(t^*+\tau_f(t)+\hat{\tau}_b(t)+2),...]$ are also given to the controller. They are used as control references if a new reference cannot be computed in the next control step due to packet losses or jitter.

After generating a first prediction, the transition from delayed to predicted references can be discontinuous (Fig. \ref{fig:alpha}). To smoothen the transition, a policy blending arbitrates the delayed received references $\bm{y}(t-\tau_f(t))$ and the predicted ones $\hat{\bm{y}}(t+\hat{\tau}_b(t)|t-\tau_f(t))$, determining the adjusted reference (Fig. \ref{fig:alpha}):
\begin{equation}
\hat{\bm{y}}'(t+\hat{\tau}_b(t)|t-\tau_f) = (1-\bm{\beta})\bm{y}_d+\bm{\beta}\bm{y}_p,
\label{eq:blending}
\end{equation}
where $\bm{y}_d=\bm{y}(t-\tau_f(t))$, ~$\bm{y}_p=\hat{\bm{y}}(t+\hat{\tau}_b(t)|t-\tau_f(t))$, $\bm{\beta}=\{\beta_0,...,\beta_n,...,\beta_N\}^\top$ with $\beta_n \in~ ]0,1[$
\begin{equation}
\beta_n=\frac{1}{1+e^{-12(\frac{i}{\Delta y_{n}}-\frac{1}{2})}},
\label{eq:beta}
\end{equation}
$i=\{0,1,...,\Delta y_{n}\}$ and $\Delta y_{n}$ is the initial difference between a delayed reference and the corresponding prediction expressed in $mm$ (for Cartesian trajectories) or $deg\times10^{-1}$ (for postural trajectories).

\subsection*{Teleoperation under unexpected circumstances}
\label{sec:unexpected}

If something unexpected happens, or if the operator suddenly changes their mind about what to do and the ongoing motion cannot be completed or is significantly altered,
the prescient teleoperation is transitioned back to the delayed teleoperation. The transition from predicted to delayed references is triggered whenever the distance between the current observation and learned mean exceeds a given threshold $\Delta_\sigma$, which in the experiments was fixed equal to the learned variance plus $5$cm and considered for each of the $x,y,z$ trajectories of the hands. 
Since the transition can be discontinuous, a policy blending arbitrates the last predicted sample $\hat{\bm{y}}(t_{last}+\hat{\tau}_b(t_{last})|t_{last}-\tau_f(t_{last}))$ and the delayed received references $\bm{y}(t-\tau_f(t))$:
\begin{equation}
\hat{\bm{y}}'(t+\hat{\tau}_b(t)|t-\tau_f) = (1-\bm{\beta})\bm{Y}_p+\bm{\beta}\bm{y}_d,
\label{eq:blending}
\end{equation}
where $\bm{y}_d=\bm{y}(t-\tau_f(t))$, ~$\bm{Y}_p=\hat{\bm{y}}(t_{last}+\hat{\tau}_b(t_{last})|t_{last}-\tau_f(t_{last}))$, $\bm{\beta}=\{\beta_0,...,\beta_n,...,\beta_N\}^\top$ with $\beta_n$ defined as in (\ref{eq:beta}).

\section*{Supplementary Materials} 
\label{sec:supp}
\begin{itemize}
    \item Movie M1: Prescient teleoperation experiments.
    \item Movie M2: Teleoperation under unexpected human behaviors.
    \item Fig. S1. Posture retargeting from human (Xsens system) to iCub.
    \item Fig. S2. Learned ProMPs for the task of reaching a bottle on the table.
    \item Fig. S3. Test trajectories for the task of reaching a bottle on the table.
    \item Fig. S4. Test trajectories for the task of picking up a box at a low position.
    \item Fig. S5. Reaching a bottle with compensation of a round-trip delay around 1.5s.
    \item Fig. S6. Teleoperation with compensation of a round-trip delay around 200ms.
    \item Fig. S7. Teleoperation with compensation of a round-trip delay around 500ms.
    \item Fig. S8. Teleoperation with compensation of a round-trip delay around 1s.
    \item Fig. S9. Teleoperation with compensation of a round-trip delay around 2s.
    \item Fig. S10. Teleoperation under unexpected human behaviors.
    \item Fig. S11. Comparison between the compensated trajectory and the ideal (non-delayed) trajectory with a round-trip delay around 1.5 s.
    \item Fig. S12. Comparison between the compensated trajectory and the ideal (non-delayed) trajectory with a round-trip delay around 2 s.
    \item Fig. S13. (Conforming the teleoperation to the intended motion) Learned ProMPs for the task of reaching a bottle on the table with different obstacles.
    \item Fig. S14. (Conforming the teleoperation to the intended motion) Test trajectories for the task of of reaching a bottle on the table with different obstacles.
    \item Fig. S15. Comparison of the prediction error on different datasets regarding the task of reaching a bottle on the table.
    \item Fig. S16. (Conforming the teleoperation to the intended motion) Prediction update according to observations.
    \item Fig. S17. (Conforming the teleoperation to the intended motion) Comparison between the compensated trajectory and the ideal (non-delayed) trajectory with a round-trip delay around 1.5s.
    \item Fig. S18. (Conforming the teleoperation to the intended motion) Prescient teleoperation with no delay.
    \item Fig. S19. (Conforming the teleoperation to new goals) Learned ProMPs for the task of reaching a bottle at different locations on the table.
    \item Fig. S20. (Conforming the teleoperation to new goals) Test trajectories for the task of reaching a bottle at different locations on the table.
    \item Fig. S21. (Conforming the teleoperation to new goals) Prediction update according to observations.
    \item Fig. S22. (Conforming the teleoperation to new goals) Comparison between the compensated trajectory and the ideal (non-delayed) trajectory with a round-trip delay around 1.5s.
    \item Fig. S23. Teleoperation controller tracking error.
    \item Fig. S24. Time modulation estimation.
\end{itemize}
 
\section*{Acknowledgements} 
The authors would like to thank E. Ghini for the precious help and support provided during the experiments and E. D'Elia for constructive criticism of the manuscript.

\paragraph{Funding:}
This work was supported by  the  European  Union  Horizon  2020  Research  and  Innovation  Program  under  Grant Agreement  No.  731540  (project  AnDy), by the  European Research Council (ERC) under Grant Agreement No. 637972 (project ResiBots), an Inria-DGA grant (``humano\"ide r\'esilient''), and the Inria ``ADT'' wbCub/wbTorque. Experiments were performed in the Creativ'Lab facilities, supported by the FEDER Sciarat.

\paragraph{Author contributions:} L.P. implemented the system and performed the experiments. J.-B.M. and S.I. proposed the initial concept and supervised the study. L.P., J.-B.M. and S.I. analyzed the results and wrote the article.

\paragraph{Competing interests:} The authors declare that they have no competing financial interests.

\paragraph{Data and materials availability:} All data needed to evaluate the conclusions in the paper are
present in the paper, the Supplementary Materials and in https://doi.org/10.5281/zenodo.5913573 (dataset of offline training and online testing control reference trajectories).

\begin{small}
    \sffamily
    \bibliographystyle{Science}
    \bibliography{biblio}
\end{small}

\twocolumn[
  \begin{@twocolumnfalse} 

    {\fontsize{18}{12}\selectfont
    \textsf{\textbf{Supplementary material}}}\\
       {\fontsize{25}{12}\selectfont
    \textsf{\textbf{Prescient teleoperation of humanoid robots}}}
    
    \bigskip
    
    \textsf{\textbf{\large{Luigi Penco$^{1}$, Jean-Baptiste Mouret$^{1}$, Serena Ivaldi$^{1}$}}}

     \bigskip   
     \textsf{$^1$~Inria Nancy -- Grand Est, CNRS, Université de Lorraine, France}
     
     \bigskip
     
     \textsf{serena.ivaldi@inria.fr}
     \bigskip
     
    \sffamily
    \noindent{}

\begin{itemize}
    \item Movie M1: Prescient teleoperation experiments. \url{https://youtu.be/N3u4ot3aIyQ}
    \item Movie M2: Teleoperation under unexpected human behaviors.
    \item Fig. S1. Posture retargeting from human (Xsens system) to iCub.
    \item Fig. S2. Learned ProMPs for the task of reaching a bottle on the table.
    \item Fig. S3. Test trajectories for the task of reaching a bottle on the table.
    \item Fig. S4. Test trajectories for the task of picking up a box at a low position.
    \item Fig. S5. Reaching a bottle with compensation of a round-trip delay around 1.5s.
    \item Fig. S6. Teleoperation with compensation of a round-trip delay around 200ms.
    \item Fig. S7. Teleoperation with compensation of a round-trip delay around 500ms.
    \item Fig. S8. Teleoperation with compensation of a round-trip delay around 1s.
    \item Fig. S9. Teleoperation with compensation of a round-trip delay around 2s.
    \item Fig. S10. Teleoperation under unexpected human behaviors.
    \item Fig. S11. Comparison between the compensated trajectory and the ideal (non-delayed) trajectory with a round-trip delay around 1.5 s.
    \item Fig. S12. Comparison between the compensated trajectory and the ideal (non-delayed) trajectory with a round-trip delay around 2 s.
    \item Fig. S13. (Conforming the teleoperation to the intended motion) Learned ProMPs for the task of reaching a bottle on the table with different obstacles.
    \item Fig. S14. (Conforming the teleoperation to the intended motion) Test trajectories for the task of of reaching a bottle on the table with different obstacles.
    \item Fig. S15. Comparison of the prediction error on different datasets regarding the task of reaching a bottle on the table.
    \item Fig. S16. (Conforming the teleoperation to the intended motion) Prediction update according to observations.
    \item Fig. S17. (Conforming the teleoperation to the intended motion) Comparison between the compensated trajectory and the ideal (non-delayed) trajectory with a round-trip delay around 1.5s.
    \item Fig. S18. (Conforming the teleoperation to the intended motion) Prescient teleoperation with no delay.
    \item Fig. S19. (Conforming the teleoperation to new goals) Learned ProMPs for the task of reaching a bottle at different locations on the table.
    \item Fig. S20. (Conforming the teleoperation to new goals) Test trajectories for the task of reaching a bottle at different locations on the table.
    \item Fig. S21. (Conforming the teleoperation to new goals) Prediction update according to observations.
    \item Fig. S22. (Conforming the teleoperation to new goals) Comparison between the compensated trajectory and the ideal (non-delayed) trajectory with a round-trip delay around 1.5s.
    \item Fig. S23. Teleoperation controller tracking error.
    \item Fig. S24. Time modulation estimation.
    
\end{itemize}
 \end{@twocolumnfalse}
]

\newpage

\begin{si-figure*}[!t]
    \centering
    \includegraphics[width=0.6\linewidth]{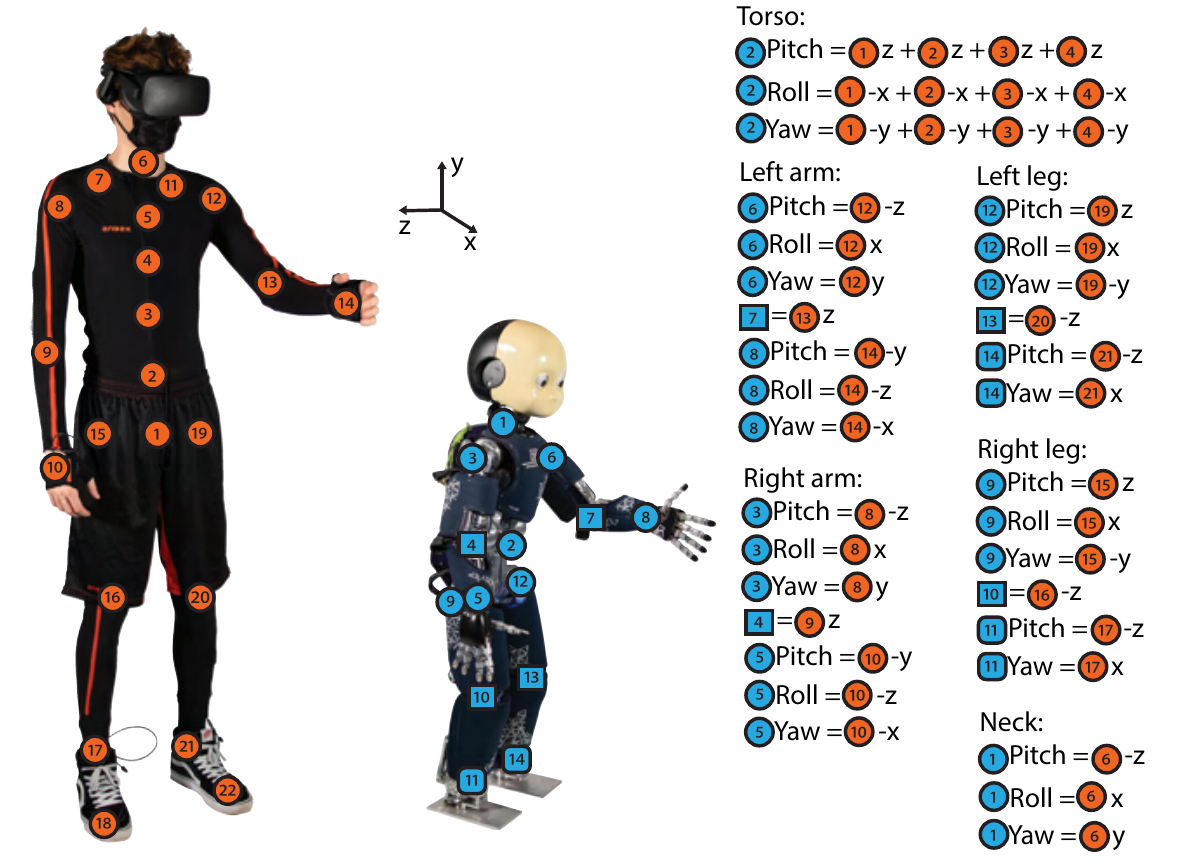}
    \caption{\textbf{Posture retargeting from human (Xsens system) to iCub.} Each joint of the robot is associated to a spherical joint of the Xsens skeleton together with its rotation axis.}
    \label{fig:retarget}
\end{si-figure*}

\begin{si-figure*}[!t]
\centering
\includegraphics[width=\linewidth]{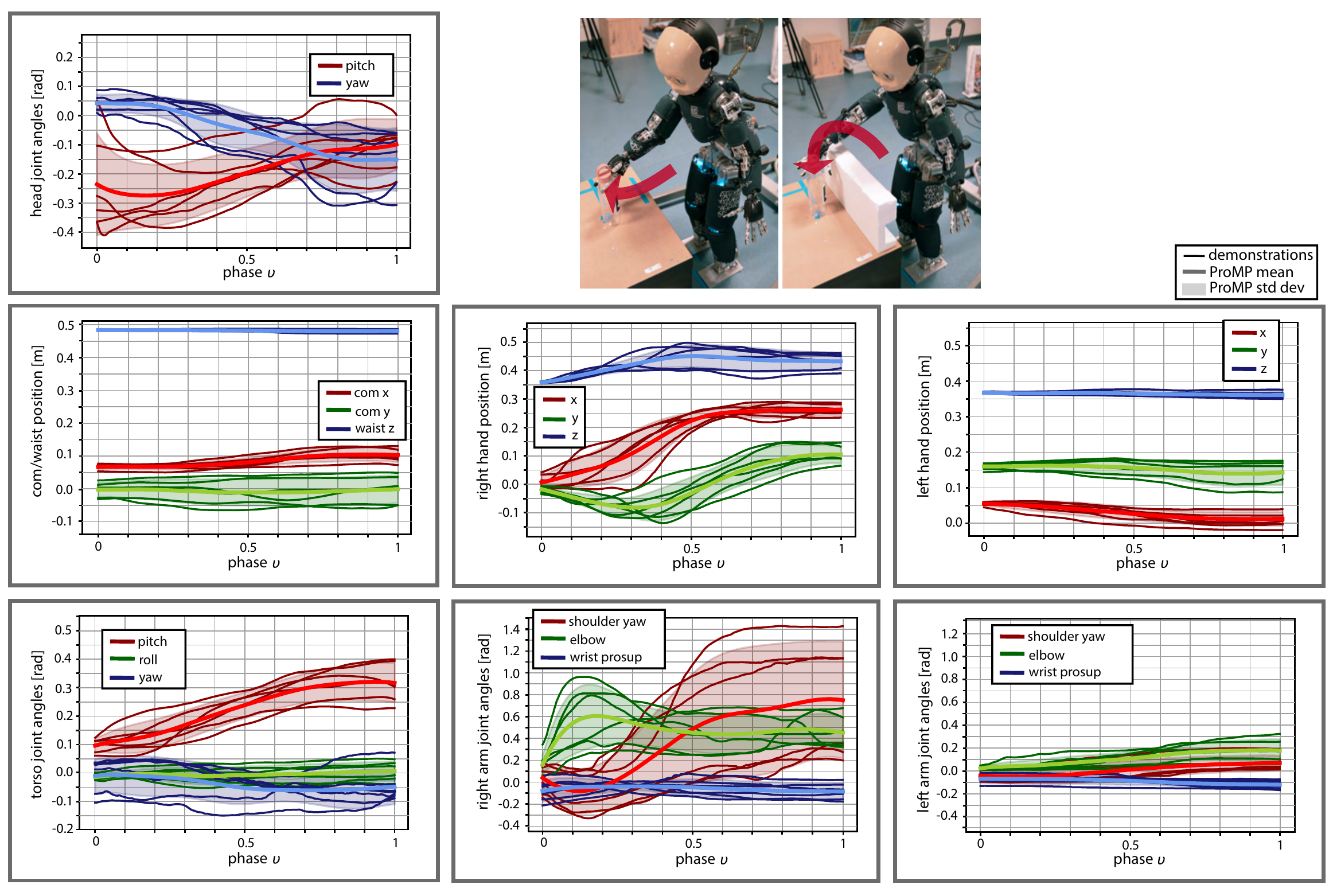}
\caption{\textbf{Learned ProMPs for the task of reaching a bottle on the table.} The whole-body motion of the teleoperated
robot is obtained by following the reference trajectories retargeted from the human. We learned a ProMP for each of these trajectories, given 6 demonstrations in a local network without any delay (3 with an obstacle in between the robot and the bottle, and 3 without)}.
\label{fig:learneds1}
\end{si-figure*}

\begin{si-figure*}[!t]
\centering
\includegraphics[width=\linewidth]{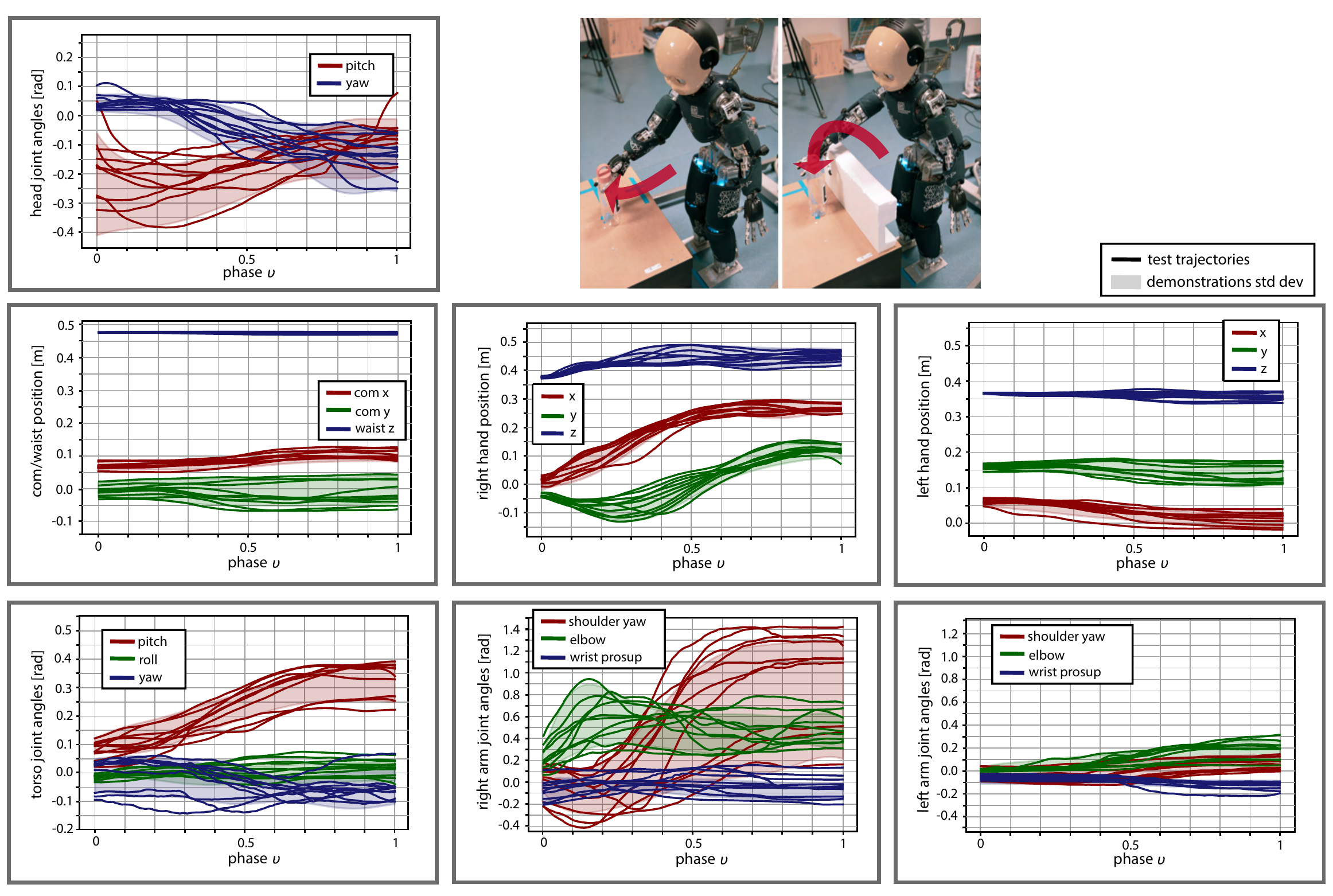}
\caption{\textbf{Test trajectories for the task of reaching a bottle on the table.} The test trajectories are different and additional repetitions of the training motions. For the task of reaching a bottle 10 different repetitions were recorded (5 with an obstacle in between the robot and the bottle, and 5 without).}
\label{fig:tests1}
\end{si-figure*}

\begin{si-figure*}[!t]
\centering
\includegraphics[width=\linewidth]{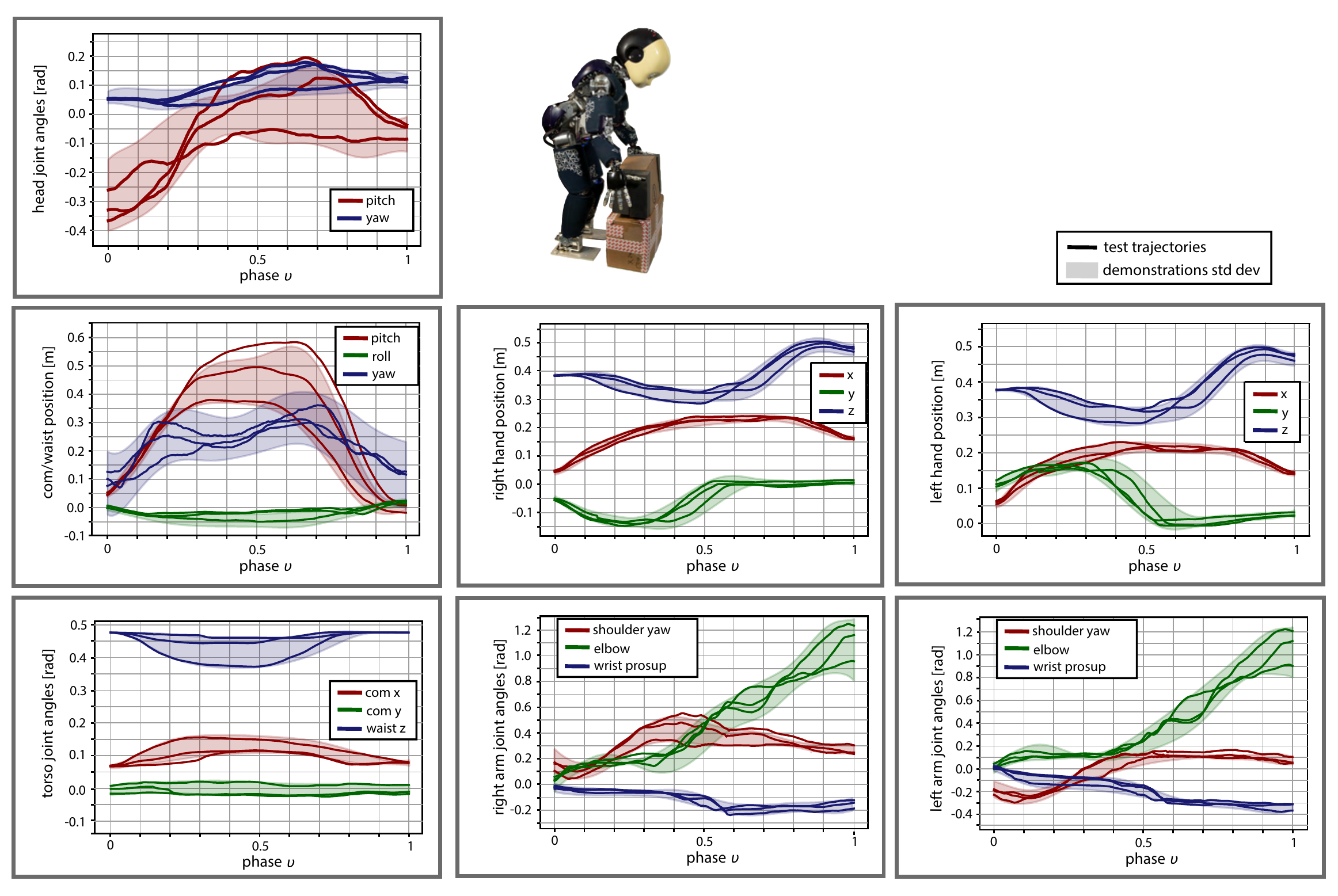}
\caption{\textbf{Test trajectories for the task of picking up a box at a low position.} The test trajectories are different and additional repetitions of the training motions. For the tasks of picking up a box 3 different repetitions were recorded.}
\label{fig:test2}
\end{si-figure*}

\begin{si-figure*}[!t]
\centering
\includegraphics[width=\linewidth]{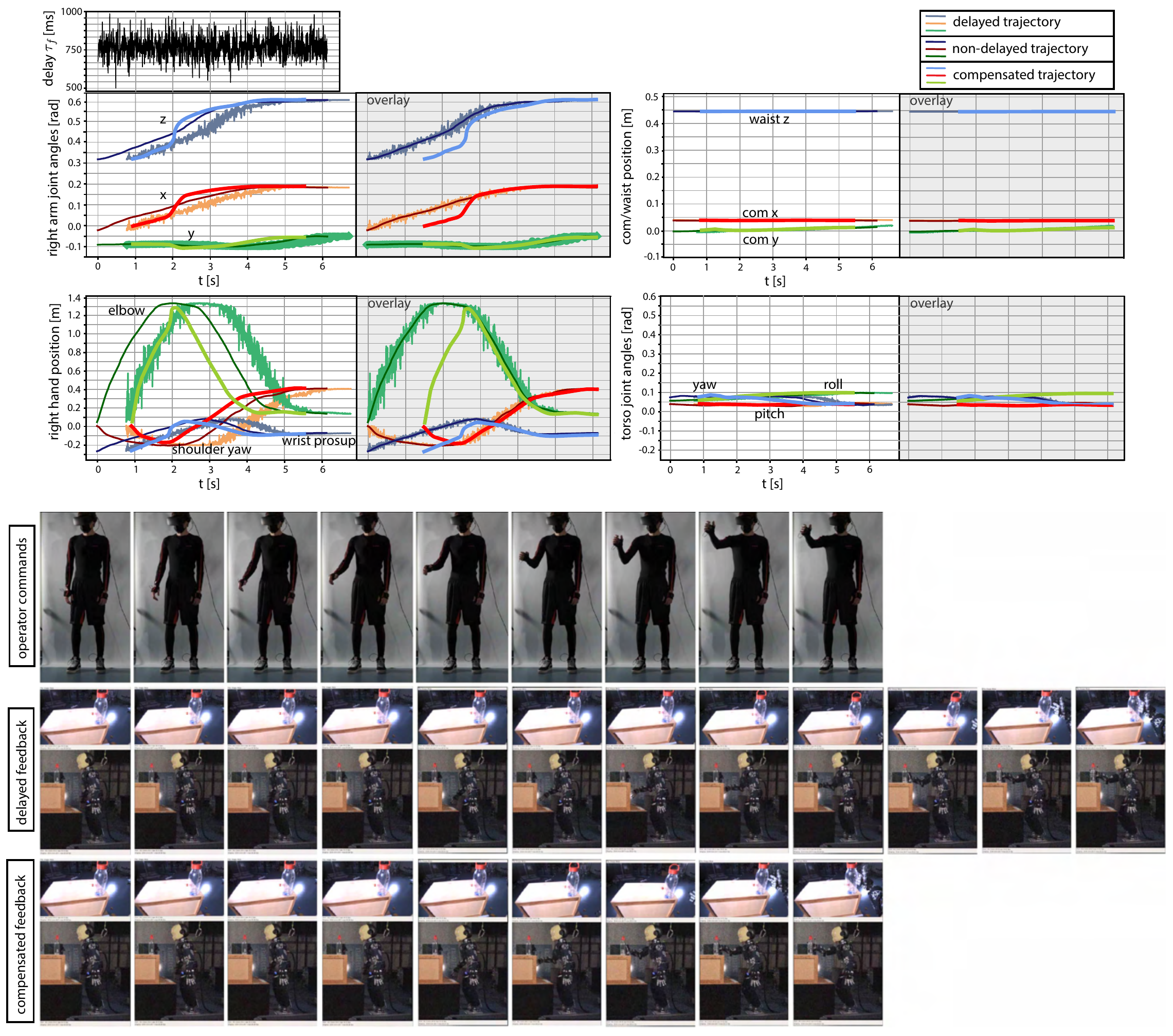}
\caption{\textbf{Reaching a bottle with compensation of a round-trip delay around 1.5s.} The robot is reaching a bottle located on top of a box. The forward delay is around 750ms with a jitter of 300ms (it follows a normal distribution with 750ms as mean and 100ms as standard deviation), while the backward delay is 750ms. 
The compensated trajectories (light red-green-blue lines) are the robot reference trajectories. These, first follow the delayed teleoperated signals (orange-teal-grey lines). Then, when the prediction is available, they anticipate the teleoperated motion (dark-colored lines) so to get a visual feedback at the user side coherent with what the operator is doing}.
\label{fig:tele15s}
\end{si-figure*}

\begin{si-figure*}[!t]
\centering
\includegraphics[width=\linewidth]{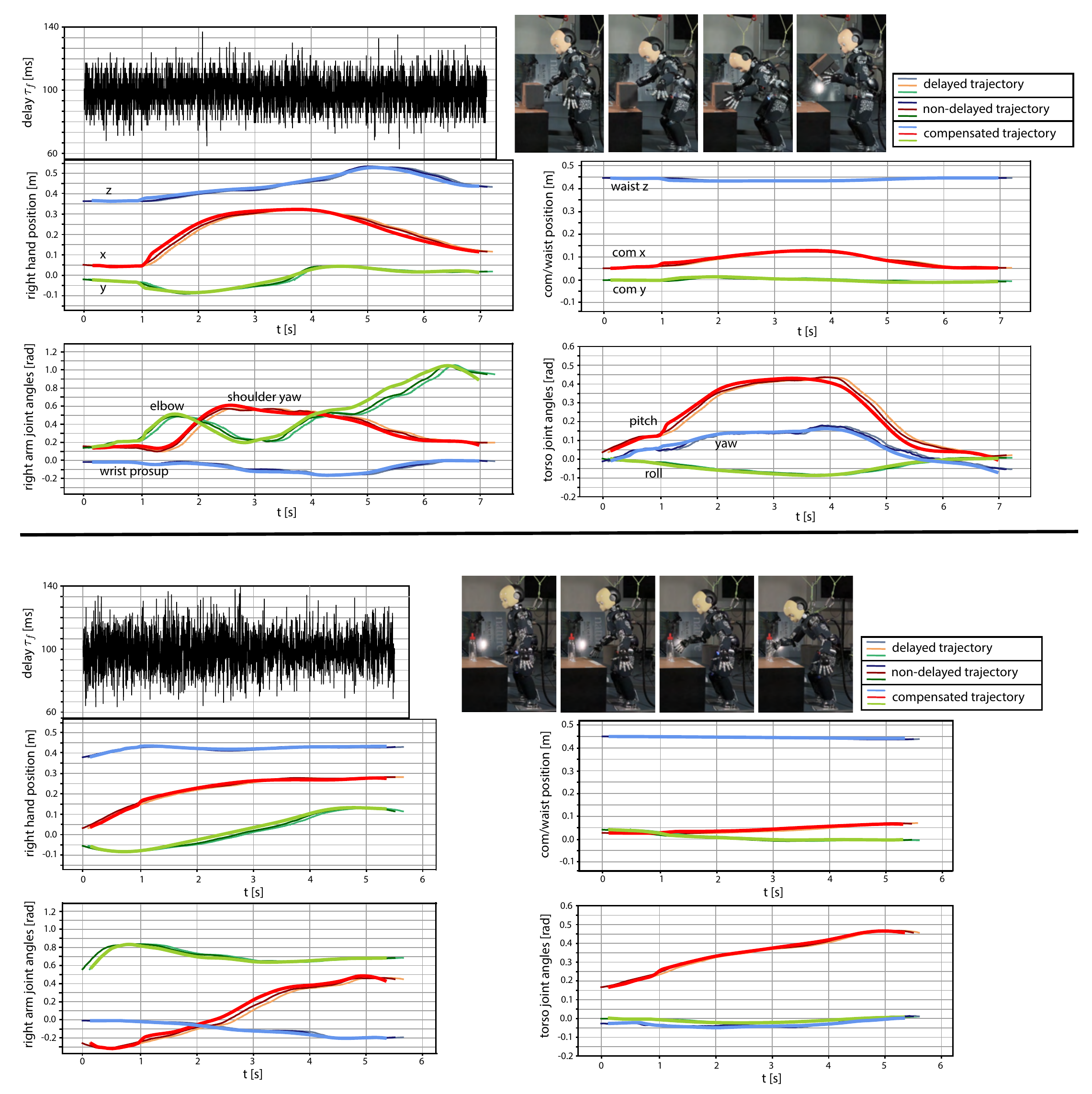}
\caption{\textbf{Teleoperation  with  compensation of a round-trip delay around 200ms.} Top: The robot is picking up a box located on a table. Bottom: The robot is reaching a bottle located on a table. The forward delay follows a normal distribution with 100ms as mean and 13.33ms as standard deviation, while the backward delay is 100ms. 
The compensated trajectories (light red-green-blue lines) are the robot reference trajectories. These, first follow the delayed teleoperated signals (orange-teal-grey lines). Then, when the prediction is available, they anticipate the teleoperated motion (dark-colored lines) so to get a visual feedback at the user side coherent with what the operator is doing}.
\label{fig:tele200}
\end{si-figure*}

\begin{si-figure*}[!t]
\centering
\includegraphics[width=\linewidth]{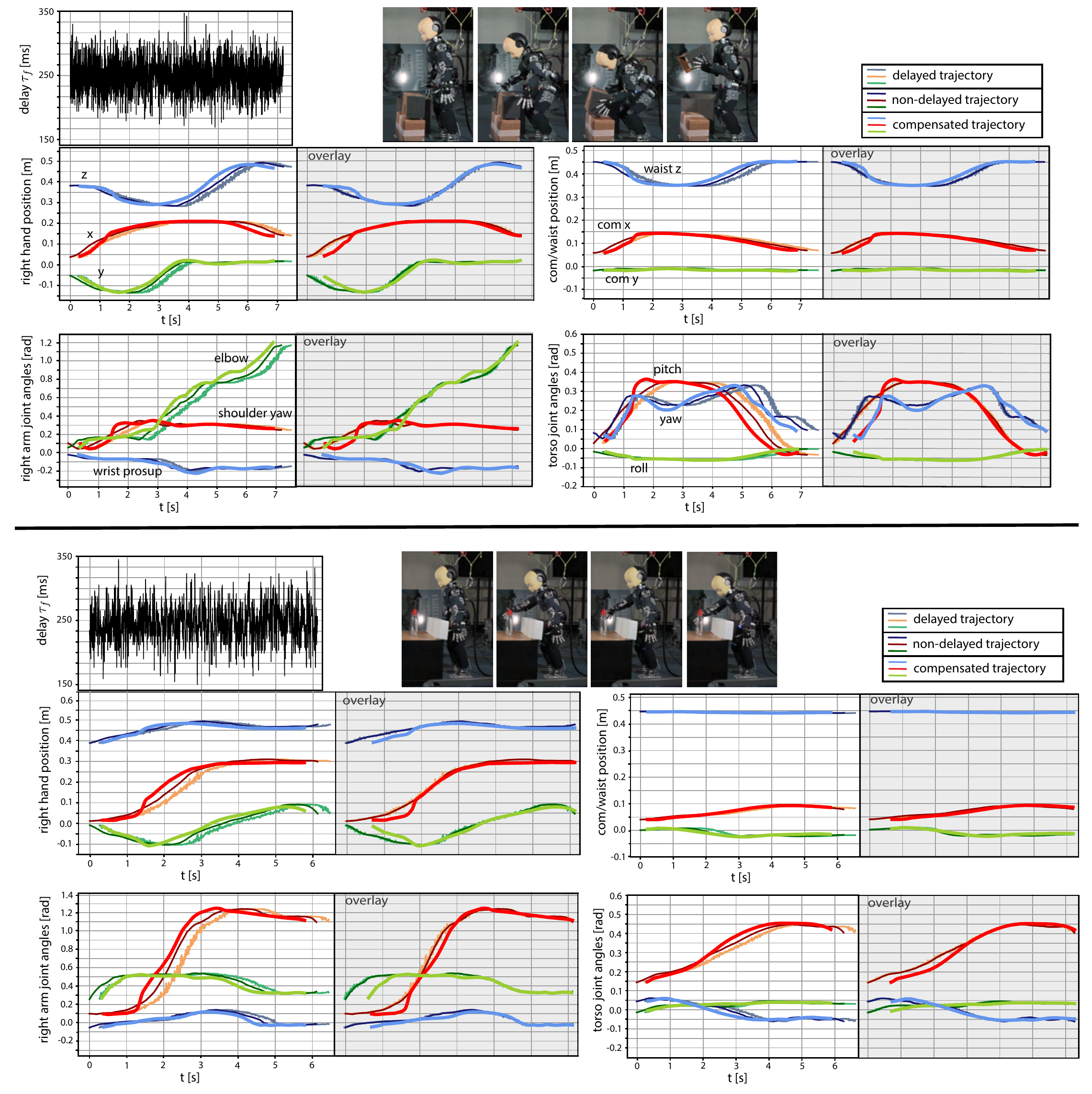}
\caption{\textbf{Teleoperation  with  compensation of a round-trip delay around 500ms.} Top: The robot is picking up a box in front of it at a low height. Bottom: The robot is reaching a bottle located on the table behind an obstacle. 
The forward delay follows a normal distribution with 250ms as mean and 33.33ms as standard deviation, while the backward delay is 250ms. 
The compensated trajectories (light red-green-blue lines) are the robot reference trajectories. These, first follow the delayed teleoperated signals (orange-teal-grey lines). Then, when the prediction is available, they anticipate the teleoperated motion (dark-colored lines) so to get a visual feedback at the user side coherent with what the operator is doing}.
\label{fig:tele500}
\end{si-figure*}

\begin{si-figure*}[!t]
\centering
\includegraphics[width=\linewidth]{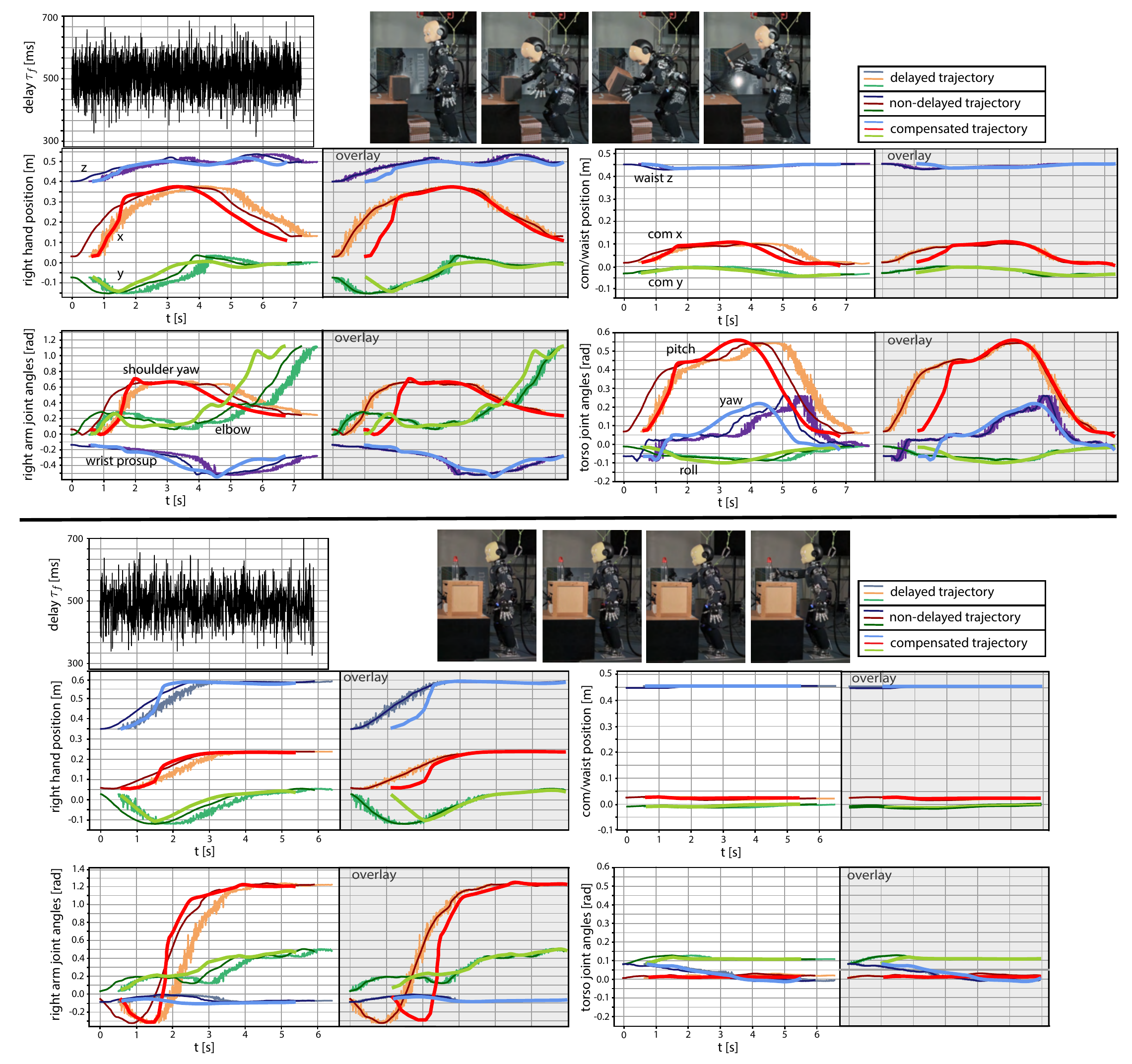}
\caption{\textbf{Teleoperation  with  compensation of a round-trip delay around 1s.} Top: The robot is picking up a box located on a table. Bottom: The robot is reaching a bottle located on a box, on the table. The forward delay follows a normal distribution with 500ms as mean and 66.67ms as standard deviation, while the backward delay is 500ms. 
The compensated trajectories (light red-green-blue lines) are the robot reference trajectories. These, first follow the delayed teleoperated signals (orange-teal-grey lines). Then, when the prediction is available, they anticipate the teleoperated motion (dark-colored lines) so to get a visual feedback at the user side coherent with what the operator is doing}.
\label{fig:tele1}
\end{si-figure*}

\begin{si-figure*}[!t]
\centering
\includegraphics[width=\linewidth]{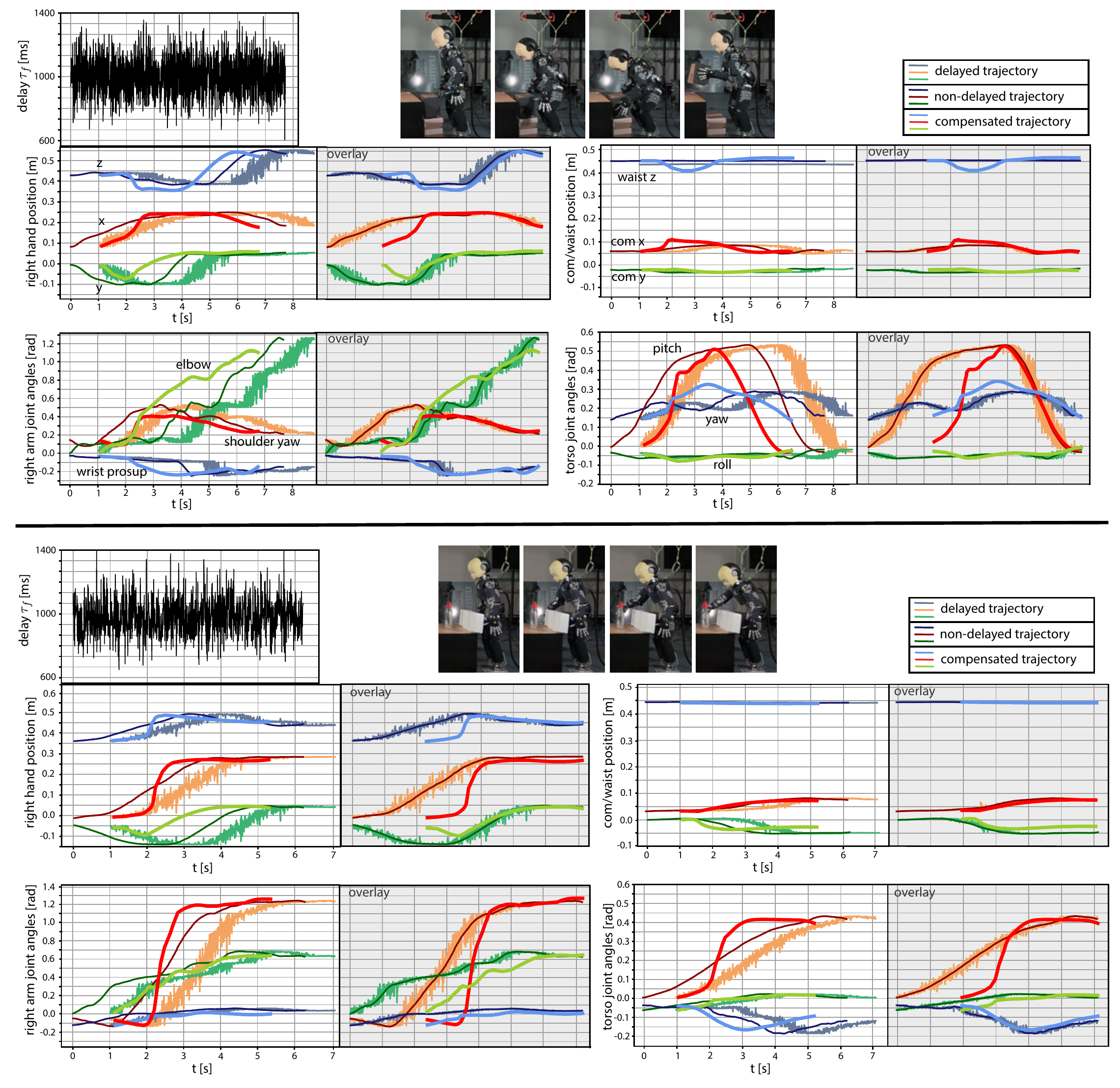}
\caption{\textbf{Teleoperation  with  compensation of a round-trip delay around 2s.} Top: The robot is picking up a box in front of it at a low height. Bottom: The robot is reaching a bottle located on the table behind an obstacle. 
The forward delay follows a normal distribution with 1000ms as mean and 133.33ms as standard deviation), while the backward delay is 1000ms. 
The compensated trajectories (light red-green-blue lines) are the robot reference trajectories. These, first follow the delayed teleoperated signals (orange-teal-grey lines). Then, when the prediction is available, they anticipate the teleoperated motion (dark-colored lines) so to get a visual feedback at the user side coherent with what the operator is doing}.
\label{fig:tele2}
\end{si-figure*}

\begin{si-figure*}[!t]
\centering
\includegraphics[width=\linewidth]{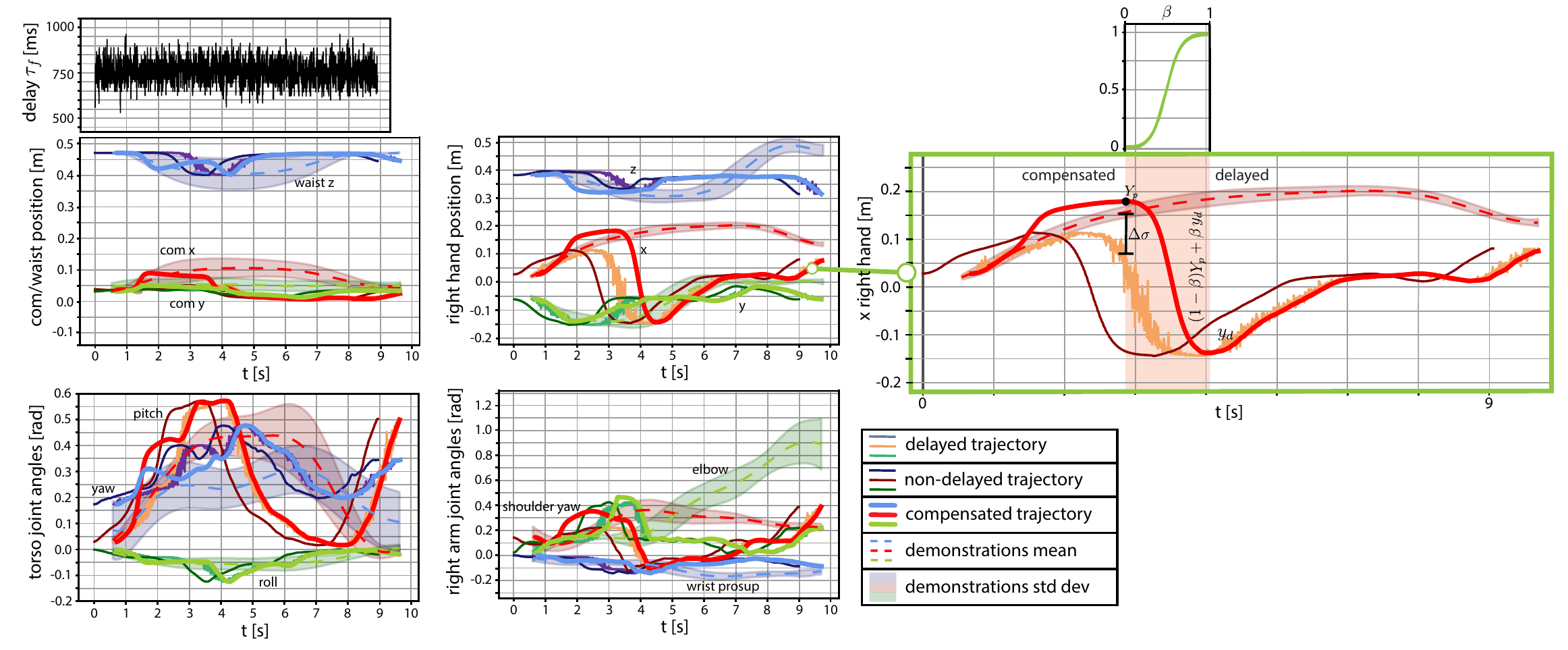}
\caption{\textbf{Teleoperation under unexpected human behaviors.} When the last received delayed signal is too far from the learned distribution (i.e. its distance $\Delta_\sigma$ with the learned mean exceeds a given threshold), the prediction is smoothly blended back into the delayed signals.}
\label{fig:mistake}
\end{si-figure*}

\begin{si-figure*}[!t]
    \centering
    \includegraphics[width=0.75\linewidth]{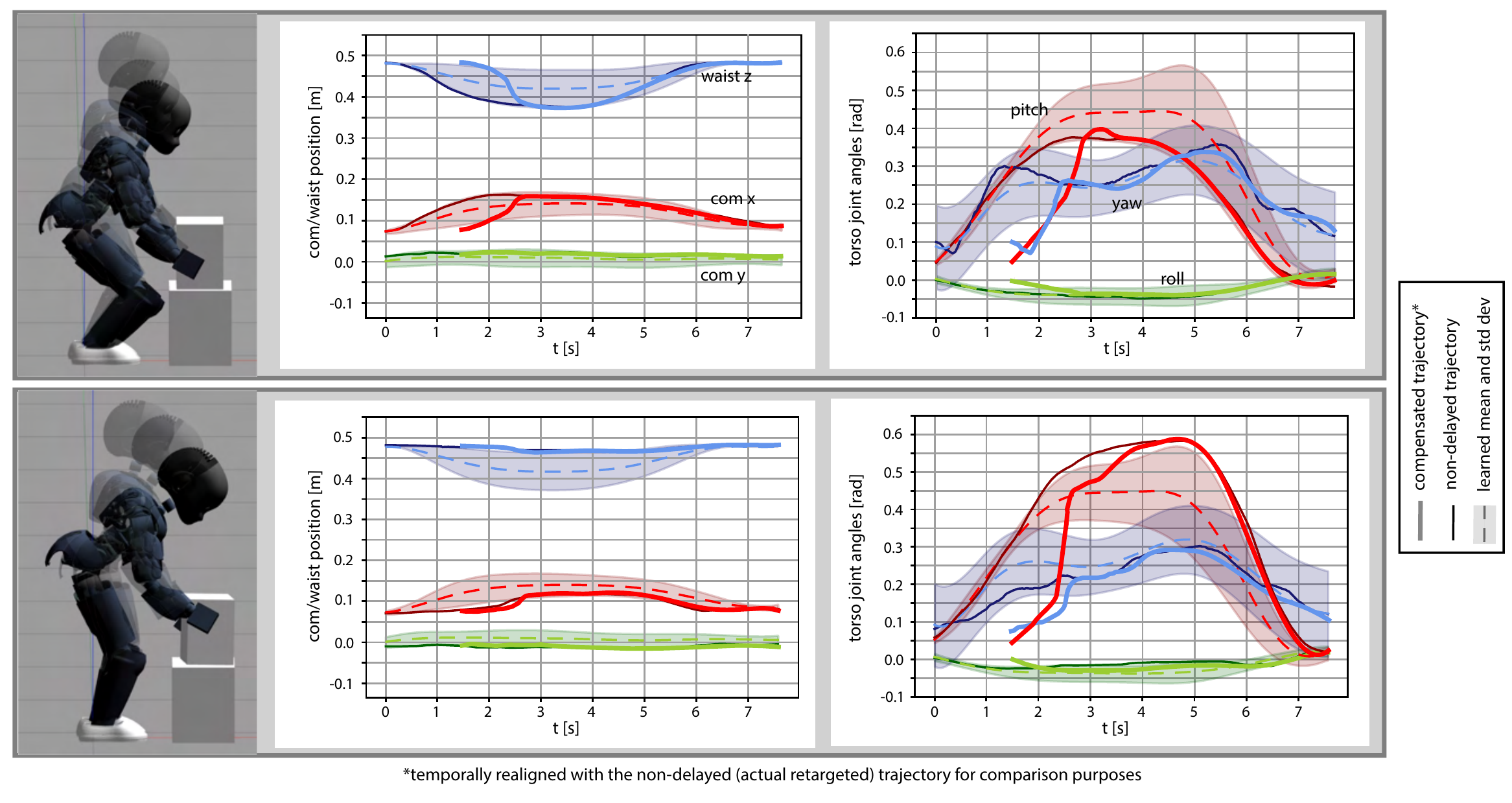}
    \caption{\textbf{Comparison between the compensated trajectory and the ideal (non-delayed) trajectory with a round-trip delay around 1.5 s.} After the initial recognition period, our approach makes the robot follow the specific way the human is performing the task despite the delay. This is not the same as following the mean of previously demonstrated motions (here, the dashed line).}
    \label{fig:cuvsmean}
\end{si-figure*}

\begin{si-figure*}[!t]
    \centering
    \includegraphics[width=0.75\linewidth]{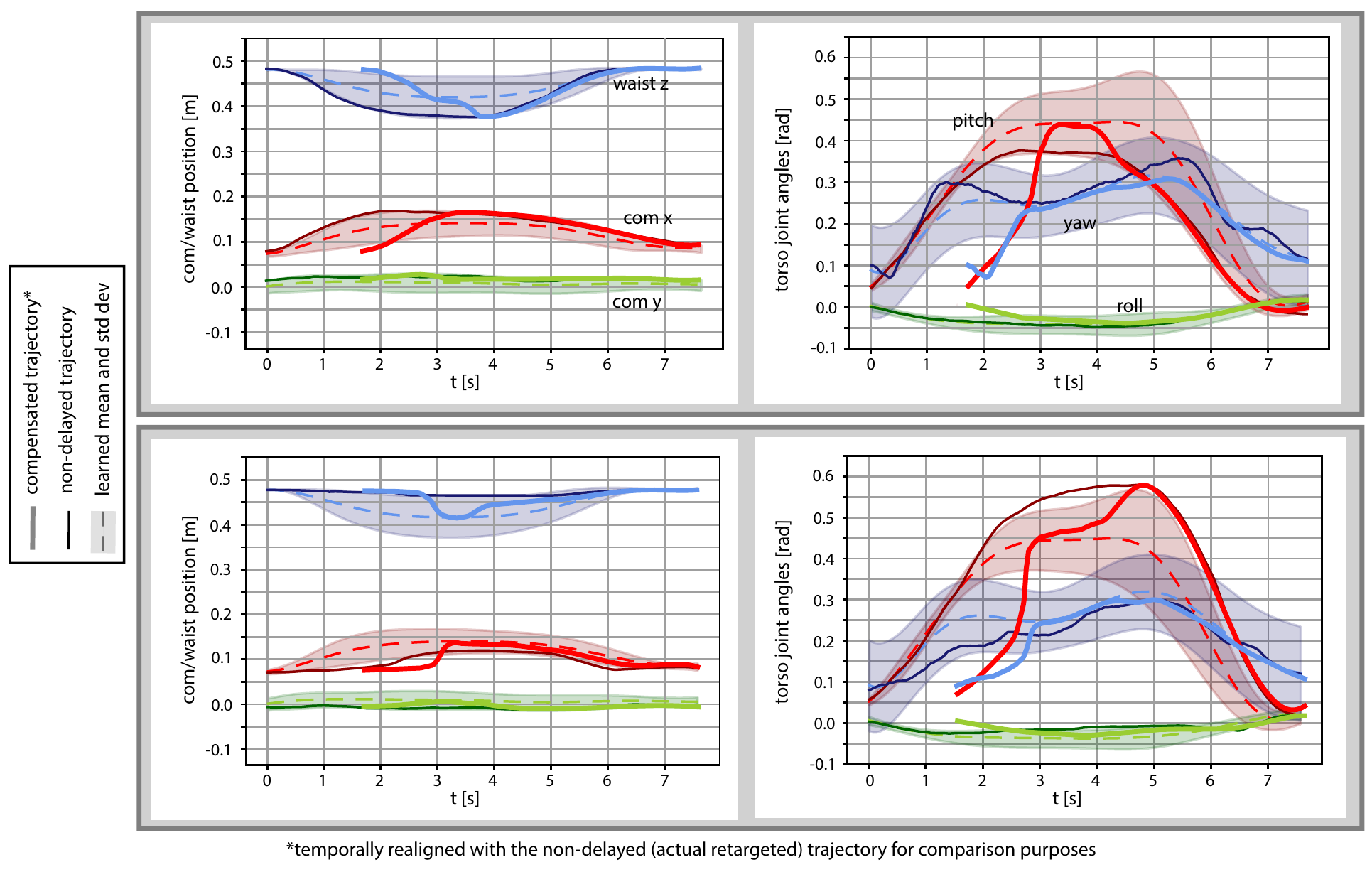}
    \caption{\textbf{Comparison between the compensated trajectory and the ideal (non-delayed) trajectory with a round-trip delay around 2 s.} For very high delays even the continuous update of the prediction might not be sufficient to make the robot follow the specific way the operator is performing the task. Here, first the robot follows the delayed retargeted movements; then tends to follow the learned mean trajectory during the beginning of the compensation; only toward the end is able to compensate for the delay while following the actual retargeted reference.}
    \label{fig:cuvsmean2}
\end{si-figure*}

\begin{si-figure*}[!t]
\centering
\includegraphics[width=\linewidth]{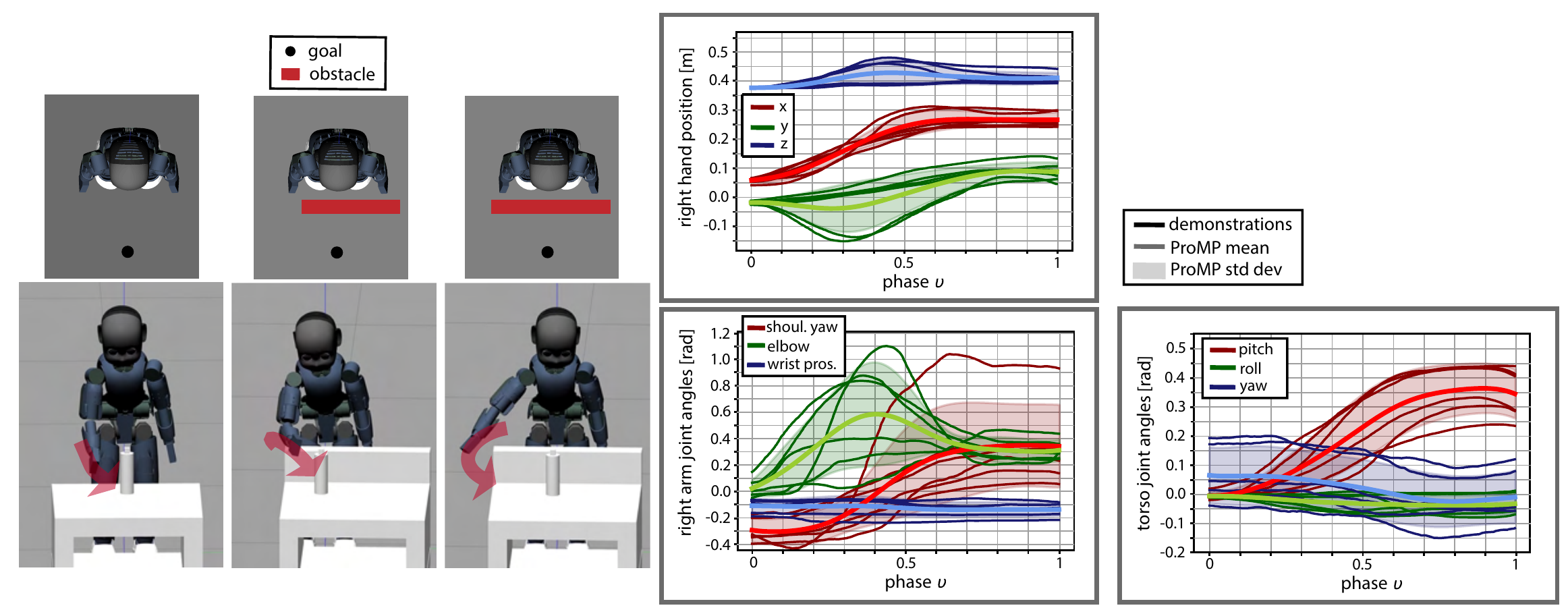}
\caption{\textbf{(Conforming the teleoperation to the intended motion) Learned ProMPs for the task of reaching a bottle on the table with different obstacles.
The most relevant learned ProMPs and the associated demonstrations are reported}. The 6 demonstrations (2 without obstacles, 2 with an obstacle in between the robot and the bottle and other 2 with a different obstacle) have been recorded while teleoperating the robot in simulation, in a local network without any delay.}
\label{fig:dataObst}
\end{si-figure*}

\begin{si-figure*}[!t]
\centering
\includegraphics[width=\linewidth]{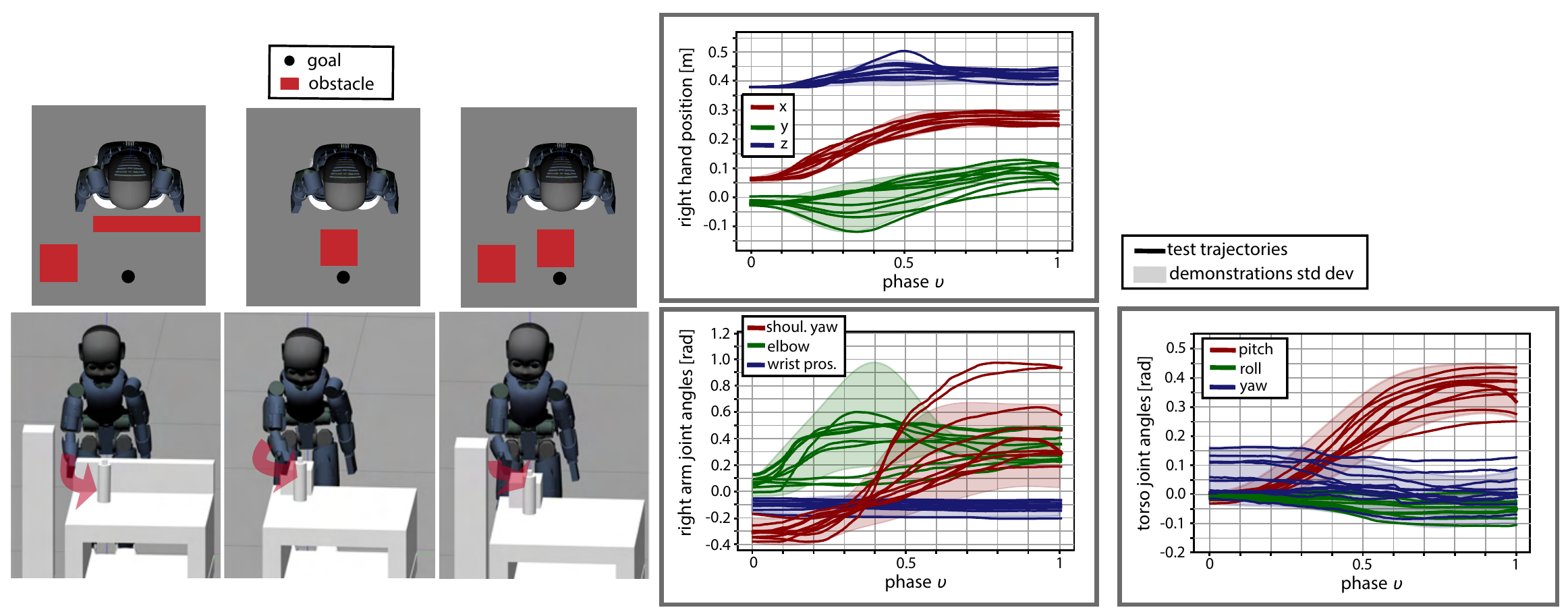}
\caption{\textbf{(Conforming the teleoperation to the intended motion) Test trajectories for the task of of reaching a bottle on the table with different obstacles.} The 9 test trajectories are different from those used for training (Fig. S\ref{fig:dataObst}), and consist of 3 repetitions of the bottle reaching motion for each of the 3 distinct simulated scenarios with different obstacles, illustrated on the left.}
\label{fig:testObst}
\end{si-figure*}

\begin{si-figure*}[!t]
\centering
\includegraphics[width=0.6\linewidth]{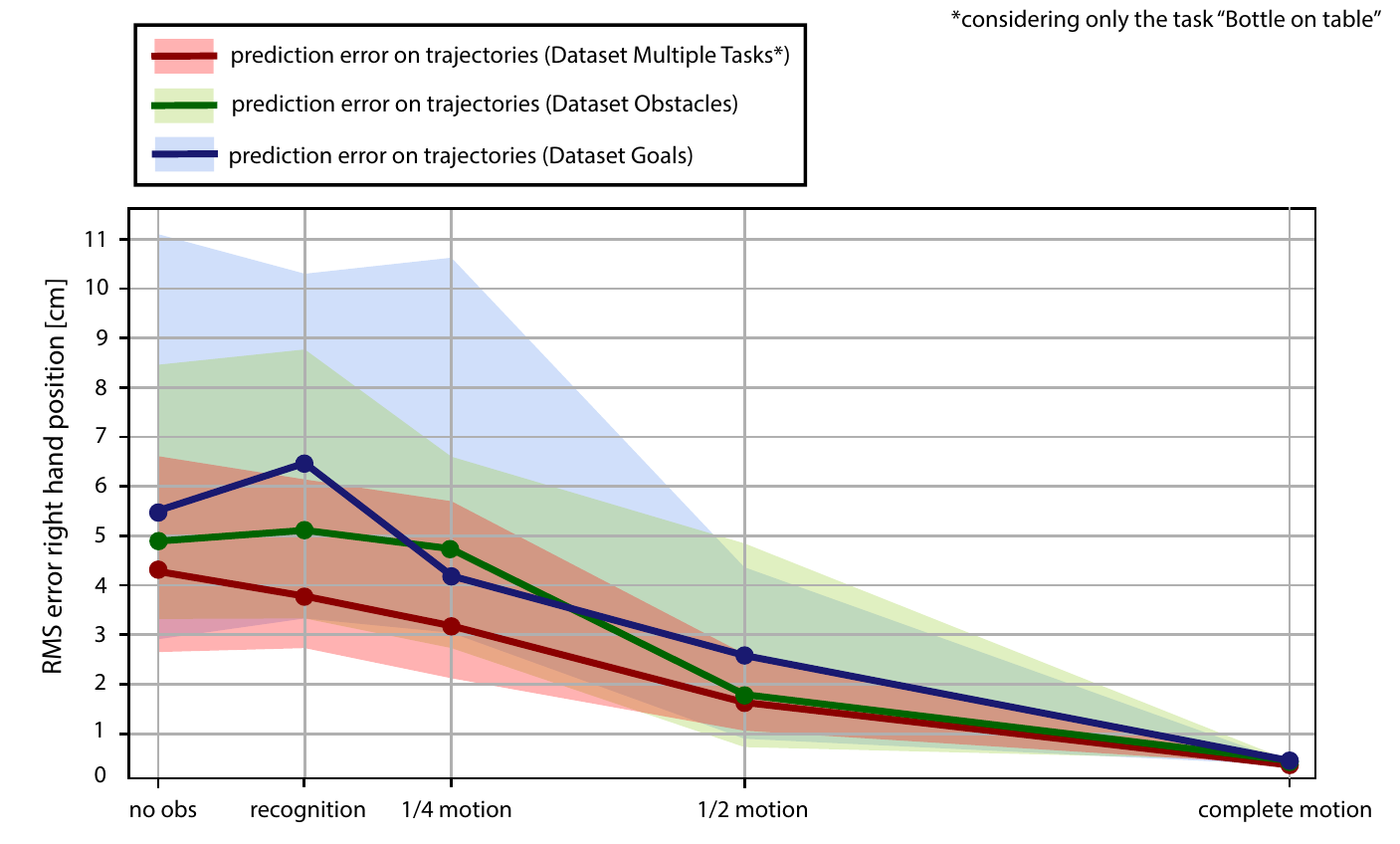}
\caption{\textcolor{blue}{\textbf{Comparison of the prediction error on different datasets regarding the task of reaching a bottle on the table.} The RMS of the error is computed based on the $10$ testing trajectories of the task of reaching the bottle on the table from Fig. S\ref{fig:tests1} (dataset Multiple Tasks), the testing trajectories from the dataset Obstacles (Fig. S\ref{fig:testObst}), and the dataset Goals (Fig. S\ref{fig:testdiff}). The prediction error is computed as the Euclidean distance between the predicted trajectory and the reference trajectory. The plot reports the error given by the mean trajectories of the ProMPs learned from the demonstrations, from the prediction updated after observing the first portion of motion used to infer the task, after observing a fourth of the motion, after observing half of the motion, and after observing the whole motion.}}
\label{fig:predcomp}
\end{si-figure*}

\begin{si-figure*}[!t]
\centering
\includegraphics[width=\linewidth]{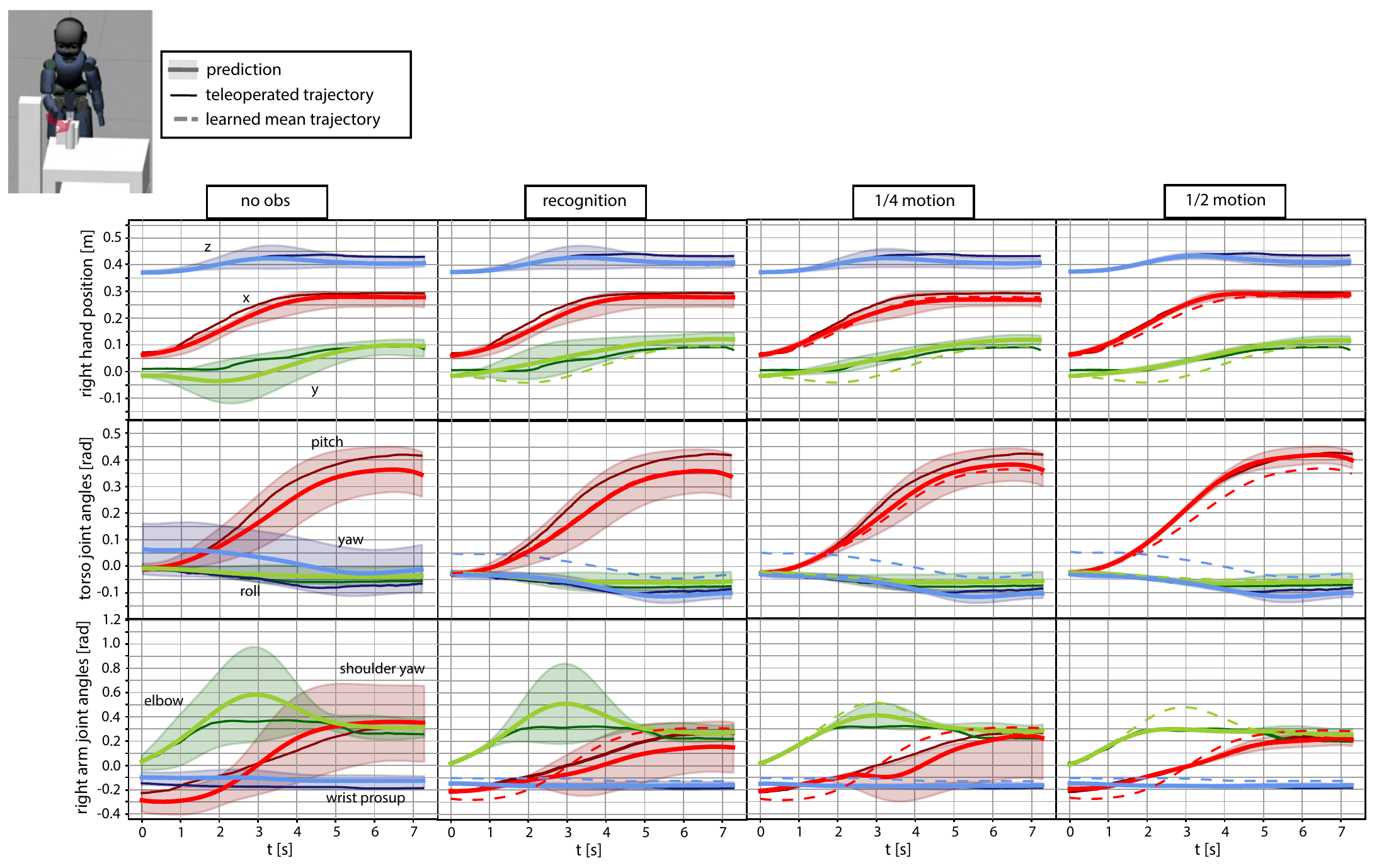}
\caption{\textcolor{blue}{\textbf{(Conforming the teleoperation to the intended motion) Prediction update according to observations.} The most relevant predicted trajectories (light colored lines) are compared to the non-delayed trajectories at the operator's side (dark colored lines), after observing different portions of the motion; a perfect prediction would mean that the light line (green/blue/red) line matches the dark line (green/blue/red). The non-delayed trajectories are from the testing set from Fig.\ref{fig:testObst}. From left to right, the figure shows the prediction given by the ProMPs learned from the demonstrations, the prediction updated after observing the first portion of motion used to infer the task and its duration, the prediction updated after observing a fourth of the motion, and after observing half of the motion.}}
\label{fig:predex}
\end{si-figure*}

\begin{si-figure*}[!t]
    \centering
    \includegraphics[width=\linewidth]{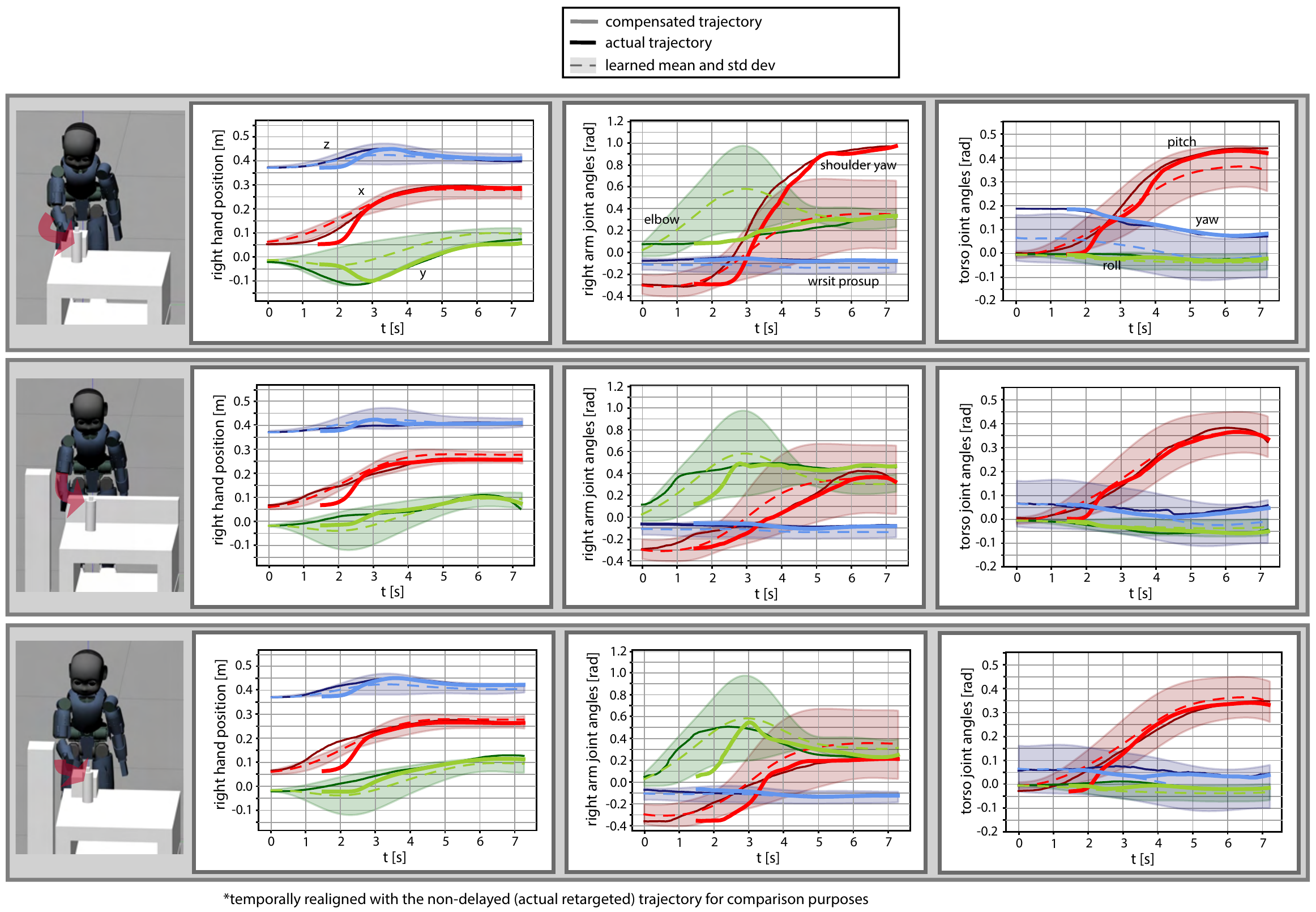}
    \caption{\textbf{(Conforming the teleoperation to the intended motion) Comparison between the compensated trajectory and the ideal (non-delayed) trajectory with a round-trip delay around 1.5s.} On the top row, there is an unexpected obstacle (a small box) to avoid and the operator approaches the object by the right; on the second row, the obstacle in front of the robot is different; on the bottom row, there is the same obstacle from the top row with in addition a large obstacle on the right of the robot, which forces the operator to move the hand in between the two obstacles. These situations were not in the training set. After the initial recognition period, our approach makes the robot follow the specific way the human is performing the task despite the delay, and even if the robot is asked to perform the task in a way that has not been demonstrated before (but included in the distribution of the demonstrations). This is not the same as following the mean of previously demonstrated motions (here, the dashed line) or letting the robot replicate previously demonstrated motions. The non-delayed trajectories are some of the test trajectories from Fig. S\ref{fig:testObst}, where the robot has to reach the bottle on the table in the presence of different obstacles that were not considered during the training.}
    \label{fig:custom15}
\end{si-figure*}

\begin{si-figure*}[!t]
\centering
\includegraphics[width=\linewidth]{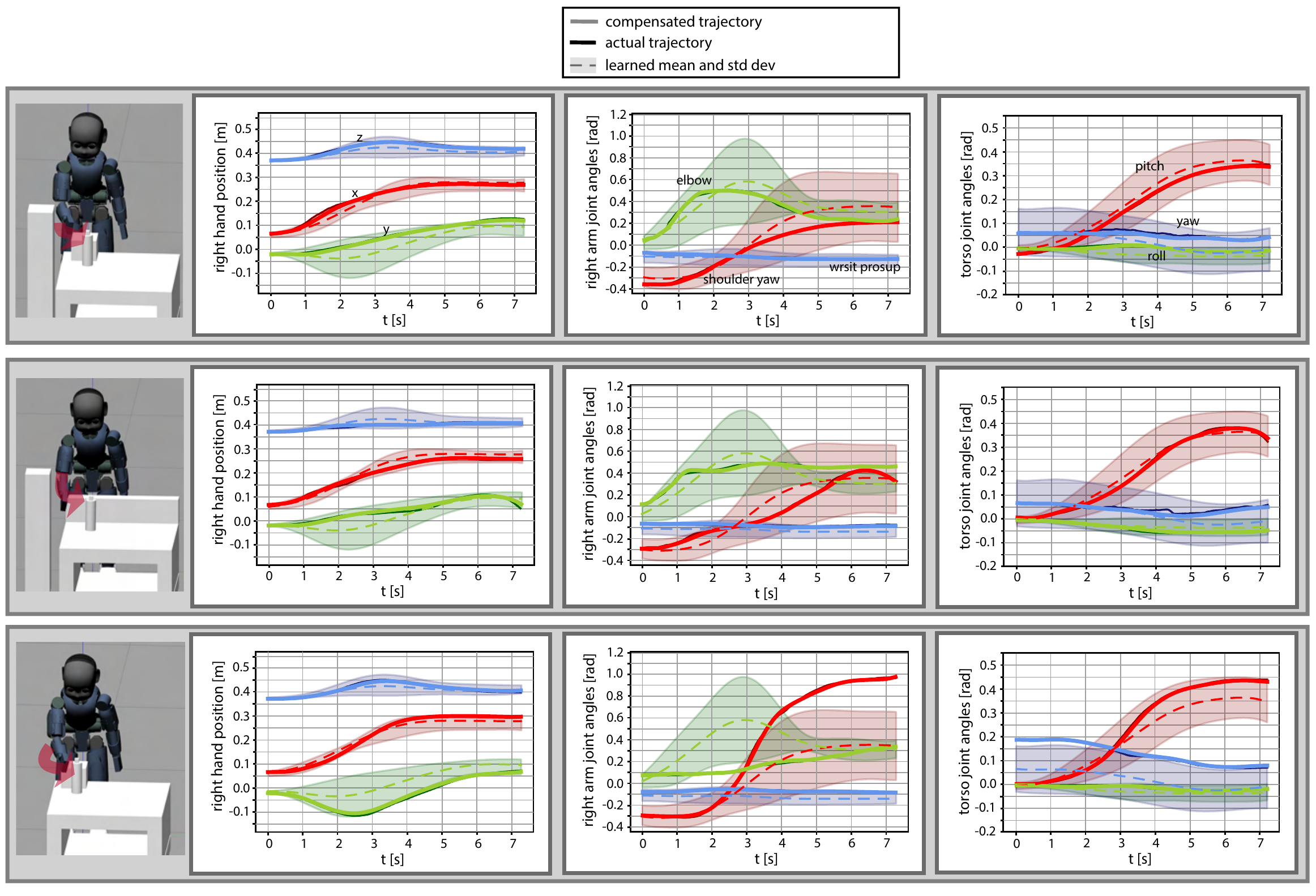}
\caption{\textbf{(Conforming the teleoperation to the intended motion) Prescient teleoperation with no delay.} The compensated trajectories resulting from our approach match almost perfectly the actual retargeted human references, when there is no delay. The actual trajectories are some of the test trajectories from Fig. S\ref{fig:testObst}.}
\label{fig:custom0}
\end{si-figure*}

\begin{si-figure*}[!t]
\centering
\includegraphics[width=\linewidth]{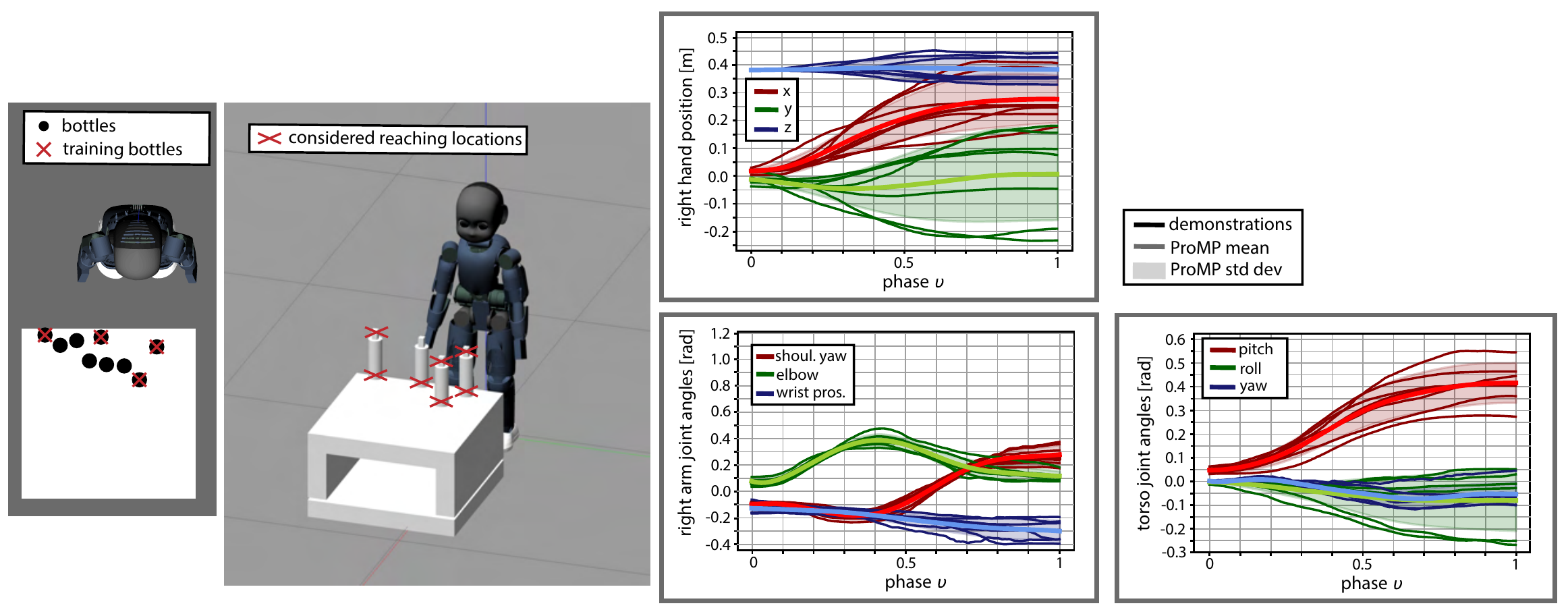}
\caption{\textcolor{blue}{\textbf{(Conforming the teleoperation to new goals) Learned ProMPs for the task of reaching a bottle at different locations on the table.}
The most relevant learned ProMPs and the associated demonstrations are reported. The 7 demonstrations have been recorded while teleoperating the robot in simulation, in a local network without any delay. The different bottle locations are illustrated on the left.}}
\label{fig:traindiff}
\end{si-figure*}

\begin{si-figure*}[!t]
\centering
\includegraphics[width=\linewidth]{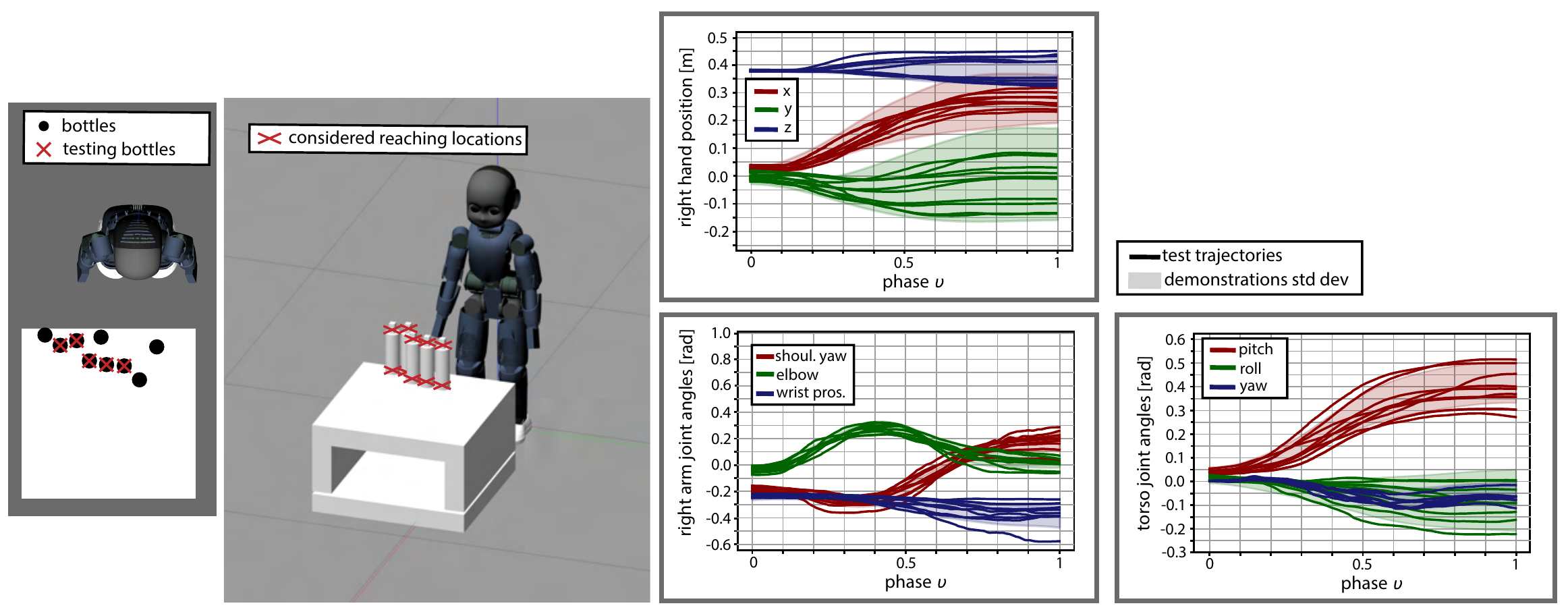}
\caption{\textcolor{blue}{\textbf{(Conforming the teleoperation to new goals) Test trajectories for the task of of reaching a bottle at different locations on the table.} The 10 test trajectories are different from those used for training (Fig. S\ref{fig:traindiff}). The different bottle locations are illustrated on the left.}}
\label{fig:testdiff}
\end{si-figure*}

\begin{si-figure*}[!t]
\centering
\includegraphics[width=\linewidth]{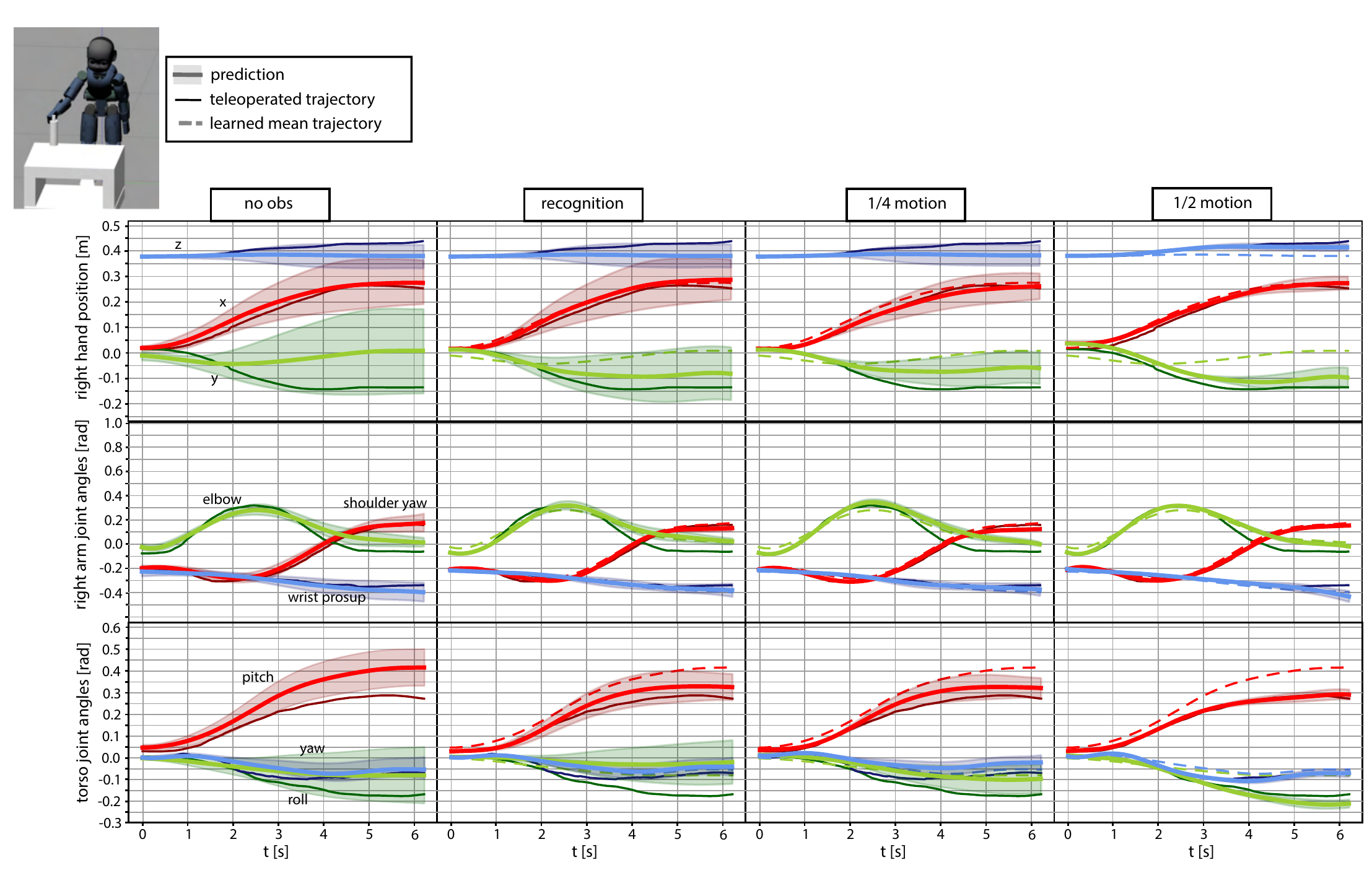}
\caption{\textcolor{blue}{\textbf{(Conforming the teleoperation to new goals) Prediction update according to observations.} The most relevant predicted trajectories (light colored lines) are compared to the non-delayed trajectories at the operator's side (dark colored lines), after observing different portions of the motion; a perfect prediction would mean that the light line (green/blue/red) line matches the dark line (green/blue/red). The non-delayed trajectories are from the testing set from Fig.\ref{fig:testdiff}. From left to right, the figure shows the prediction given by the ProMPs learned from the demonstrations, the prediction updated after observing the first portion of motion used to infer the task and its duration, the prediction updated after observing a fourth of the motion, and after observing half of the motion.}}
\label{fig:predexdiff}
\end{si-figure*}

\begin{si-figure*}[!t]
\centering
\includegraphics[width=\linewidth]{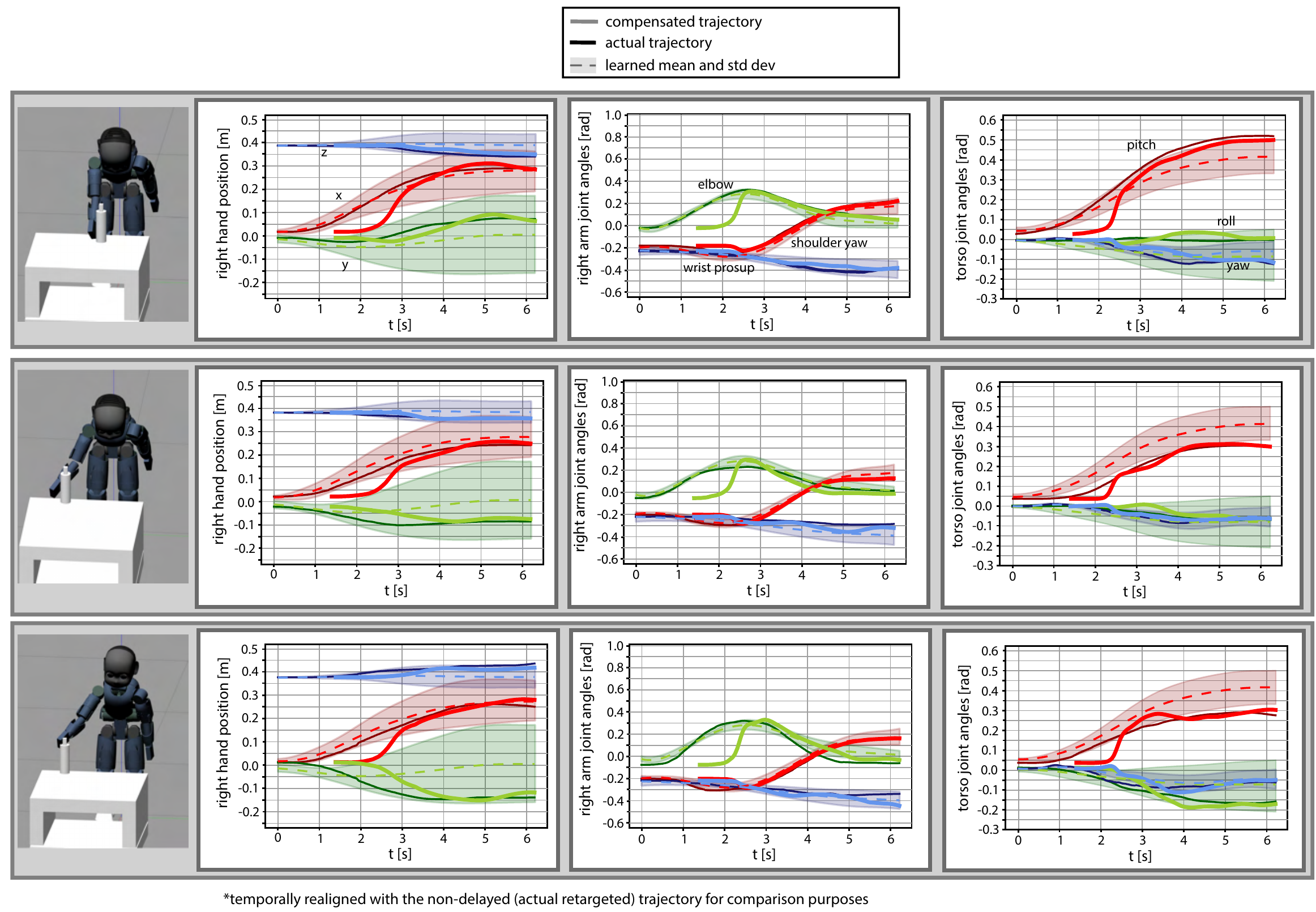}
\caption{\textcolor{blue}{\textbf{(Conforming the teleoperation to new goals) Comparison between the compensated trajectory and the ideal (non-delayed) trajectory with a round-trip delay around 1.5s.} The bottles are located on the table in different positions that were not in the training set (but included in the distribution of the demonstrations). The non-delayed trajectories are some of the test trajectories from Fig. S\ref{fig:testdiff}, where the robot has to reach the bottle on the table in the presence of different obstacles that were not considered during the training.}}
\label{fig:customdiff}
\end{si-figure*}

\begin{si-figure*}[!t]
    \centering
    \includegraphics[width=0.75\linewidth]{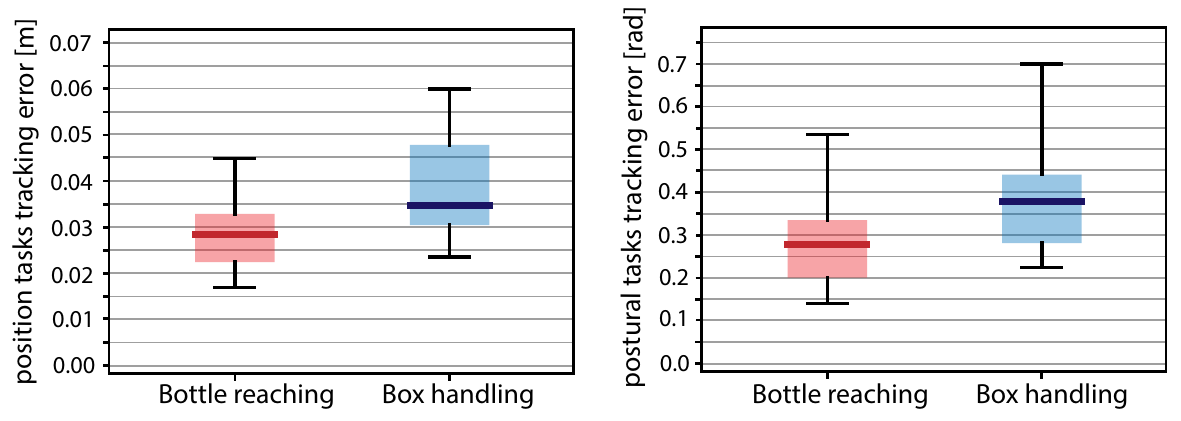}
    \caption{\textbf{Teleoperation controller tracking error.} Boxplots representing the mean cumulative tracking error (during the execution of a teleoperated reference motion) of the position tasks (center of mass x,y position, waist height, left and right hand position) and the postural tasks (head pitch and yaw, torso pitch, roll and yaw, left and right shoulder yaw, elbow and wrist prosupination) of the multi-task controller, in the two main experimental scenarios of reaching a bottle at different locations with the right hand and handling a box. The tracking error was measured while teleoperating the robot during each of the 54 demonstrations from the training set (12 motions from the bottle reaching scenario and 42 motions from the box handling scenario)}.
    \label{fig:controller}
\end{si-figure*}

\begin{si-figure*}[!t]
    \centering
    \includegraphics[width=0.6\linewidth]{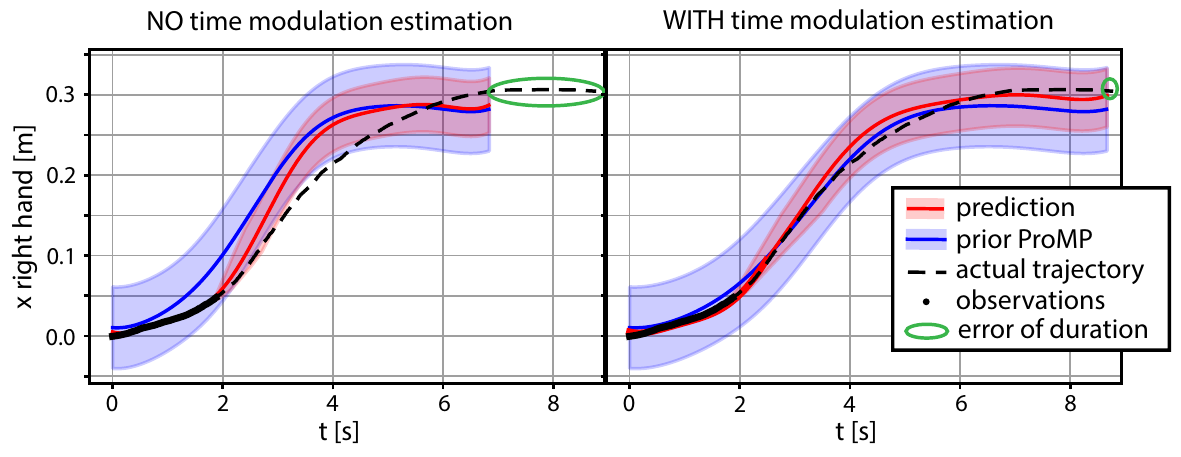}
    \caption{\textbf{Time modulation estimation.} Predicted trajectory given some early observations compared to the actual trajectory, with (right) and without (left) time modulation estimation for the task of reaching a bottle in front of the robot. Without time modulation estimation, the duration of the mean trajectory of a ProMP is set equal to the mean duration of the demonstrations, leading to mismatches between teleoperated (dashed line) and predicted (red line) motion.}
    \label{fig:timemod}
\end{si-figure*}


\end{document}